\documentclass[journal]{new-aiaa}
\usepackage[utf8]{inputenc}
\usepackage{graphicx}
\usepackage[version=4]{mhchem}
\usepackage{siunitx}
\usepackage{longtable,tabularx}
\usepackage{pdfsync} 
\usepackage{subcaption} 	
\usepackage{amsmath} 		
\usepackage{algorithmic}  	
\usepackage{algorithm}    	
\usepackage{mathtools}    	
\usepackage{float}        	
\usepackage{hyperref}     	
\usepackage{enumitem}     	
\usepackage{booktabs}     	
\usepackage{multirow}     	
\usepackage{makecell}     	
\usepackage{bm}           	
\usepackage[table]{xcolor}
\newtheorem{definition}{Definition}
\newtheorem{remark}{Remark}[section]
\newtheorem{theorem}{Theorem}

\makeatletter
\renewenvironment{theorem}[1][]{%
  \refstepcounter{theorem}%
  \ifx\relax#1\relax
    \def\@currentlabel{\thetheorem}%
  \else
    \def\@currentlabel{#1}%
  \fi
  \trivlist
  \item[\hskip\labelsep{\rmfamily\bfseries Theorem \thetheorem.}]%
  \rmfamily
}{%
  \endtrivlist
}

\renewenvironment{definition}[1][]{%
  \refstepcounter{definition}
  \trivlist
  \item[\hskip\labelsep{\rmfamily\bfseries Definition \thedefinition.%
        \ifx\relax#1\relax\else\ #1\fi}]%
  \rmfamily
}{%
  \endtrivlist
}
\makeatother

\title{Recursive Inference for Heterogeneous Multi-Output GP State-Space Models with Arbitrary Moment Matching}

\author{Tengjie Zheng \footnote{Ph.D. Candidate, School of Astronautics, ZhengTengjie@buaa.edu.cn.} 
and
Jilan Mei \footnote{Ph.D. Candidate, School of Astronautics, zb2515202@buaa.edu.cn.}
and
Di Wu \footnote{Associate Professor, School of Astronautics, wudi2025@buaa.edu.cn.}
and
Lin Cheng \footnote{\textbf{Corresponding Author}, Associate Professor, School of Astronautics, chenglin5580@buaa.edu.cn.}
and
Shengping Gong \footnote{Professor, School of Astronautics, gongsp@buaa.edu.cn.}
}

\affil{School of Astronautics, Beihang University, Beijing, 102206, China}
\affil{State Key Laboratory of High-Efficiency Reusable Aerospace Transportation Technology, Beijing, 102206, China}

\begin{document}

\maketitle

\begin{abstract}
    Accurate learning of system dynamics is becoming increasingly crucial for advanced control and decision-making in engineering. However, real-world systems often exhibit multiple channels and highly nonlinear transition dynamics, challenging traditional modeling methods.
    To enable online learning for these systems, this paper formulates the system as Gaussian process state-space models (GPSSMs) and develops a recursive learning method. The main contributions are threefold.
    First, a heterogeneous multi-output kernel is designed, allowing each output dimension to adopt distinct kernel types, hyperparameters, and input variables, improving expressiveness in multi-dimensional dynamics learning. Second, an inducing-point management algorithm enhances computational efficiency through independent selection and pruning for each output dimension. Third, a unified recursive inference framework for GPSSMs is derived, supporting general moment matching approaches, including the extended Kalman filter (EKF), unscented Kalman filter (UKF), and assumed density filtering (ADF), enabling accurate learning under strong nonlinearity and significant noise. 
    Experiments on synthetic and real-world datasets show that the proposed method matches the accuracy of SOTA offline GPSSMs with only $1/100$ of the runtime, and surpasses SOTA online GPSSMs by around $70\%$ in accuracy under heavy noise while using only $1/20$ of the runtime.
\end{abstract}

\section*{Nomenclature}


{\renewcommand\arraystretch{1.0}
\noindent\begin{longtable*}{@{}l @{\quad=\quad} l@{}}
$\mathcal{GP}$			        &	Gaussian process (GP) \\
$\bm m_0(\cdot)$			        &	prior mean function of the GP \\
$\bm K_0(\cdot, \cdot; \bm\theta)$		&	prior covariance function (or kernel function) of the GP \\
$\bm{\theta}$				    & 	kernel hyperparameters, nondimensional unit \\
$\bm K_{ab}$                    &   kernel matrix indexed by function outputs $\bm a$ and $\bm b$, nondimensional unit \\
$\bm{\Sigma}_p$ 		        &   process noise covariance, nondimensional unit \\
$\bm{\Sigma}_m$ 		        &   measurement noise covariance, nondimensional unit \\
$\bm{x}_t$			            &   system state at time $t$, nondimensional unit \\
$\bm{y}_t$			            &   measurement at time $t$, nondimensional unit \\
$\bm c_t$                       &   control input at time $t$, nondimensional unit \\
$\bm\omega_p$                   &   process noise, nondimensional unit \\
$\bm\omega_m$                   &   measurement noise, nondimensional unit \\
$\bm{F}(\cdot)$			        &   known transition function structure \\
$\bm f(\cdot)$			            &   unknown component of the transition function $\bm{F}(\cdot)$ \\
$\bm{g}(\cdot)$			        &   measurement function \\
$\bm z$                         &   $\{\bm z^k\}_{k=1}^{d_f}$, where $\bm z^k$ is the model input for the $k$-th dimension of the latent function $\bm f(\cdot)$, nondimensional unit \\
$\varphi^k(\cdot)$              &   mapping from state $\bm x$ and control input $\bm c$ to the model input $\bm z^k$. \\
$p(\cdot)$			            &   exact probability density function \\
$q(\cdot)$			            &   approximate probability density function \\
$\bm{m}_{a}$			        &   mean of the random variable $\bm a$, nondimensional unit \\
$\bm{S}_{ab}$                   &   covariance between the random variables $\bm a$ and $\bm b$, nondimensional unit \\ 
$\bm{f}$			            &   all function values of $\bm f(\cdot)$, nondimensional unit \\
$\bm{u}$			            &   inducing-point set, nondimensional unit \\
$\bm f_{\ne u}$                 &   function values of $\bm f(\cdot)$ excluding the inducing points $\bm u$, nondimensional unit \\
$\bm{h}_t$                      &   latent function prediction $\bm f(\bm x_t)$, nondimensional unit \\
$\bm f_t$                       &   function value $\bm f(\bm{m}_{x_t})$, nondimensional unit \\
$\bar{\bm{u}}$                  &   augmented inducing-point set, i.e., $\bar{\bm{u}} = [\bm{u}, \bm f_t]$, nondimensional unit \\
$\bm{F}_t$                      &   predicted mean of the state transition function, i.e., $\bm{F}_t = \bm{F}(\bm{m}_{x_t}, \bm m_{f_t})$, nondimensional unit \\
$\bm X_t$                       &   augmented state at time $t$, i.e., $\bm X_t = [\bm{x}_t, {\bm{u}}]$, nondimensional unit \\
$\bm{\xi}_t$                    &   mean of the augmented state $\bm X_t$, nondimensional unit  \\
$\bm{\Sigma}_t$                 &   covariance of the augmented state $\bm X_t$, nondimensional unit  \\
$\bm{\Phi}$                     &   transition Jacobian for the augmented state $\bar{\bm{X}}_t$, nondimensional unit \\
$\bm{\Sigma}_{p,\bm{X}}$        &   process noise covariance for the augmented state $\bar{\bm{X}}_t$, nondimensional unit \\
$M$				                &   budget for inducing-point set size, nondimensional unit \\
$u_{\mathrm{d}}$                &   inducing point to be discarded, nondimensional unit \\
$\bm{u}_{\mathrm{l}}$           &   inducing points left after discarding, nondimensional unit \\
$s_d$                           &   score evaluting the importance of the inducing points, nondimensional unit \\
$\bm Q$                         &   $\bm{K}_{uu}^{-1}$, inversion of the kernel matrix $\bm{K}_{uu}$, nondimensional unit \\
$\bm\Omega$                     &   $\bm{\Sigma}_t^{-1}$, inversion of the joint covariance $\bm{\Sigma}_t$, nondimensional unit \\
$\bm{Q}_{dd}$                   &   the diagonal element of the matrix $\bm Q$ corresponding to the discarded point $u_{\mathrm d}$, nondimensional unit \\
$\bm{Q}_{du}$                   &   the row of the matrix $\bm Q$ corresponding to the discarded point $u_{\mathrm d}$, nondimensional unit \\
$\bm{\Omega}_{dd}$              &   the diagonal element of the matrix $\bm\Omega$ corresponding to the discarded point $u_{\mathrm d}$, nondimensional unit \\
$\gamma$			            &   novelty metric for inducing points, nondimensional unit \\
$\varepsilon_{\text{tol}}$	    &   threshold for adding inducing points, nondimensional unit \\
\multicolumn{2}{@{}l}{Superscripts}\\
$-$ & predicted distribution    \\
$*$ & optimal distribution      \\
$k$ or $l$ & index of the function output dimension \\
\end{longtable*}}

\section{Introduction}



\lettrine{A}{ccurate} modeling of dynamical systems is fundamental for engineering applications, including control system design \cite{islam2025adaptive, scampicchio2025gaussian,patel2025reachable}, target tracking \cite{hua2024space, goel2022injection, sun2021collaborative}, and trajectory prediction \cite{maeder2011trajectory}.
Traditional modeling methods can be broadly classified into two categories: physics-based (or first-principles) methods and data-driven methods.
Physics-based methods use established physical laws to describe system behavior, but their accuracy is often limited due to the inherent complexity of real-world systems, such as nonlinear aerodynamics, friction, and elasticity. As a result, data-driven modeling techniques have attracted increasing attention as a powerful complement, which enables flexible and accurate system identification directly from observed data.
In recent years, various data-driven modeling approaches have been developed, including neural networks \cite{2025Experience}, Gaussian processes (GPs) \cite{zheng2024recursive}, operator learning \cite{iacob2025learning}, and generative models \cite{wang2023extraction}. Among these, GP-based methods are particularly favored by control engineers because of their flexibility, high data efficiency, and ability to quantify uncertainty, which are especially attractive for safety-critical and adaptive control applications \cite{mchutchon2015nonlinear,fisac2018general}.

In many engineering and control problems, system dynamics are represented as state-space models (SSMs), which describe the system evolution and measurement process using the state transition and measurement functions.
SSMs are widely applied because they facilitate system behavior analysis and controller design \cite{rangapuram2018deep,gedon2021deep,frigolaBayesianTimeSeries2015, zheng2025model}.
To leverage the advantages of both GPs and SSMs, Gaussian Process State-Space Models (GPSSMs) \cite{frigolaBayesianTimeSeries2015} have been proposed. In principle, GPSSMs use GPs to model state transition and measurement functions in a probabilistic and data-driven manner.
Over the past decade, a range of inference methods for GPSSMs have been developed, including sample-based \cite{frigola2013bayesian,svensson2017flexible}, variational inference (VI)-based \cite{eleftheriadis2017identification,doerrProbabilisticRecurrentStateSpace2018,Ialongo2019,lindinger2022laplace,Lin2024}, and hybrid approaches \cite{frigola2014variational, fan2023free}. 
Despite these advancements, the majority of existing GPSSM approaches are tailored for offline learning scenarios, wherein the complete dataset is accessible and the learning process is typically computationally intensive.
However, in practical engineering scenarios, the environment or system properties can change over time or vary across different tasks.
Offline-trained models often lack the adaptability required for such non-stationary conditions, and retraining with new data from scratch can be computationally prohibitive. Therefore, these challenges underscore the urgent need for online GPSSM methods that can efficiently and robustly adapt to evolving system dynamics.

However, achieving online learning of GPSSMs presents several challenges. As detailed in \cite{zheng2024recursive}, the main challenges include model nonlinearity, the nonparametric nature of GPs, the coupling between the system state and the model, and the difficulty of adapting the model hyperparameters.
Due to these challenges, existing online GPSSM methods are limited and can be broadly categorized into three types: Extended Kalman Filter (EKF)-based methods \cite{veiback2019learning,Kullberg2020,Kullberg2021}, Particle Filter (PF)-based methods \cite{Berntorp2021,zhaoStreamingVariationalMonte2023,liu2023sequential}, and Stochastic Variational Inference (SVI)-based methods \cite{frigola2014variational,Lin2024,Park2022}. 
Among these methods, both EKF-based and PF-based approaches first pre-parameterize the GP as a static structure with finite dimensional parameters by using inducing point approximation \cite{veiback2019learning,Kullberg2020} or spectrum approximation \cite{Kullberg2021}. After that, they use the EKF or PF to jointly infer the system state and GP parameters in a sequential manner from measurement data.
In contrast, SVI-based methods primarily extend the offline variational inference approach for GPSSMs by employing stochastic optimization techniques to enable online learning. 
Regarding performance, EKF-based methods are advantageous in computational efficiency, PF-based methods excel in nonlinear filtering capability, and SVI-based methods offer adaptability for the operating domain and GP hyperparameters. However, none of these methods achieves a comprehensive balance among these various performance aspects. For example, the pre-parameterization used in EKF-based and PF-based methods restricts the ability to learn outside the predefined operating domain. For the SVI-based method, online data usually do not satisfy the uniformly random condition required for stochastic optimization, which makes it difficult to ensure the convergence of the learning process.

Therefore, to address these limitations, our previous work \cite{zheng2024recursive} introduced the Recursive Gaussian Process State Space Model (RGPSSM) method. 
This method achieves recursive inference for GPSSMs based on the EKF framework and can adapt both the operating domain and the GP hyperparameters.
Specifically, the inducing points approximation is used to make inference tractable, and an online adding and discarding algorithm is proposed to select a representative set of inducing points, thus adapting to arbitrary operating domains. 
In addition, to optimize the GP hyperparameters without the need to store historical data, the likelihood model is extracted from the posterior distribution to serve as an information source for the optimization process, which improves learning accuracy.
Experimental results have demonstrated that RGPSSM achieves a desirable trade-off among learning flexibility, accuracy, and computational efficiency. 

However, the original RGPSSM method \cite{zheng2024recursive} mainly focused on improving learning flexibility. It did not fully address complex multi-output transition function learning or high-order nonlinearity. 
Consequently, there are two aspects requiring improvement:

(1) \textbf{Limited capability for learning heterogeneous multi-output transition functions:} 
To learn multi-output transition functions, the RGPSSM method \cite{zheng2024recursive} adopts the linear model of coregionalization (LMC) \cite{alvarez2012kernels} method to construct the multi-output kernel. In particular, it constructs a multi-output kernel by multiplying a single-output kernel with a signal covariance matrix.
In this configuration, the input variables and length-scale hyperparameters of the kernel, must be identical for all function dimensions. However, in many systems, the types of input variables for the different function dimensions can vary, and the sensitivity between function input and output can also differ. 
For example, the aerodynamic force and moment coefficients of the aircraft in different channels may depend on different variables and show different sensitivities.
As a result, the isomorphic learning setting of RGPSSM imposes significant limitations on both learning flexibility and accuracy for these heterogeneous function scenarios.


(2) \textbf{Limited accuracy of EKF-based moment matching:} To address the nonlinearity in GPSSMs and the nonparametric nature of GPs, the RGPSSM method uses first-order linearization to approximate the system model. This leads to a statistical moment matching approach similar to the EKF for recursively inferring the state and the GP. However, first-order linearization provides highly accurate approximations only when the system has low-order nonlinearity or when the uncertainty is small. As a result, when learning highly nonlinear dynamical systems or when there is significant measurement noise, the RGPSSM may experience severe performance degradation.

To address these challenges, this paper proposes an enhanced RGPSSM framework that supports heterogeneous transition function learning and incorporates advanced Bayesian filter techniques, significantly improving the flexibility and accuracy of online GPSSM learning. The main contributions are as follows:

\begin{itemize}
    \item To support learning of heterogeneous transition models, we introduce a heterogeneous multi-output kernel. This kernel allows each function dimension to use different input variables, kernel functions, and hyperparameters, which improves the representational capacity. Furthermore, a new management strategy for inducing points is developed, 
    enabling independent selection for each function dimension and pruning of 
    redundant inducing points, which improves computational efficiency and 
    numerical stability.
    \item To support the application of advanced Bayesian filtering techniques, a unified inference framework for RGPSSM is derived. This framework reformulates the inference problem into a finite-dimensional setting without relying on first-order linearization, making it suitable for general moment matching techniques. Based on this, several moment matching methods are integrated to enhance learning accuracy, including the Extended Kalman Filter (EKF) and the Unscented Kalman Filter (UKF) for general nonlinearities, as well as the Assumed Density Filter (ADF), which provides exact moment matching for the Gaussian kernel case.
    \item To ensure the numerical stability of the learning algorithm, we derive a stable implementation method based on Cholesky factorization, which applies to all three types of moment matching methods described above. Furthermore, the complete code for the proposed method is available at \href{https://github.com/TengjieZheng/rgpssm}{https://github.com/TengjieZheng/rgpssm} to support further development and application.
    \item Comprehensive experimental evaluations are conducted, covering tasks such as highly nonlinear transition function learning, time-varying parameter identification, hypersonic vehicle flight dynamics learning, and real-world quadrotor dynamics learning.
    The results indicate that, compared to state-of-the-art (SOTA) offline and online GPSSM approaches, the proposed method achieves higher learning accuracy and notable advantages in computational efficiency.
\end{itemize}


















\paragraph{Notation.} In this paper, we use $\{\bm e_i\}_{i=1}^n$ to denote a set of $n$ elements, and $[\bm v_i]_{i=1}^n$ indicates the column-wise concatenation of the $n$ column vectors.


\section{Problem Formulation}

In this section, we first review the fundamentals of Gaussian Processes (GPs), then introduce the Gaussian Process State Space Models (GPSSMs). Building on this foundation, we briefly summarize the Recursive Gaussian Process State Space Model (RGPSSM) \cite{zheng2024recursive} method and highlight its limitations, which are the focus of this paper.

\subsection{Gaussian Processes (GPs)}

Gaussian Process is a probability distribution over function space, which is widely used for regression tasks.
To learn a multi-input, multi-output function $\bm f(\cdot)$, we can first place a GP prior on it:

\begin{equation}
\begin{aligned}
\bm f(\bm z) \sim \mathcal{GP}\left(\bm m_0(\bm z), 
\bm K_0(\bm z, \bm z^\prime; \bm{\theta}) \right)
\end{aligned}
\end{equation} 
where $\bm m_0(\bm z) = \mathbb{E}_{p(\bm f)}[\bm f(\bm z)]$ denotes the prior mean function, and 
$\bm K_0(\bm z, \bm z^\prime;\bm{\theta}) = 
\mathbb E_{p(\bm f)}[(\bm f(\bm z) - \bm m_0(\bm z))(\bm f(\bm z^\prime) - \bm m_0(\bm z^\prime))^T]$ 
represents the prior covariance function (a.k.a. kernel function). 
Additionally, $\bm{\theta}$ denotes the kernel hyperparameters, which can be learned from data to capture properties such as output scale and input-output sensitivity.
As a common setting, we assume zero-mean GP prior, namely, $\bm m_0(\cdot) \equiv 0$. 
Given noise-free function values 
\( \bm{f}_{\mathcal{Z}} = [\bm f(\bm{z}_i)]_{i=1}^{n_\mathcal{Z}} \) 
at inputs \( \mathcal{Z} = \{\bm{z}_i\}_{i=1}^{n_\mathcal{Z}} \), 
the posterior distribution for test function values 
\( \bm{f}_* = [\bm f(\bm{z}_i^*)]_{i=1}^{n_{\mathcal{Z}^*}} \) 
at inputs \( \mathcal{Z}^* = \{\bm{z}_i^*\}_{i=1}^{n_{\mathcal{Z}^*}} \) 
can be obtained by the Bayesian principle \cite{rasmussenGaussianProcessesMachine2008}:

\begin{equation}\label{eq:GP_cond}
\begin{aligned}
    p(\bm{f}_*|\bm{f}_{\mathcal{Z}}) = \mathcal{N}(
    \bm{K}_{f_*f_\mathcal{Z}} 
    \bm{K}_{f_\mathcal{Z}f_\mathcal{Z}}^{-1} \bm{f}_{\mathcal{Z}}, \,
     \bm{K}_{f_*f_*} 
    - \bm{K}_{f_*f_\mathcal{Z}} \bm{K}_{f_\mathcal{Z}f_\mathcal{Z}}^{-1} 
    \bm{K}_{f_\mathcal{Z}f_*} )
\end{aligned}
\end{equation}
where $\mathcal N(\cdot)$ is the Gaussian distribution. Note that, this paper uses the function outputs to index the kernel matrices $\bm K$. For example, \( \bm{K}_{f_*f_*} = \mathrm{cov}_{p(\bm f)}[\bm{f}_*, \bm{f}_*] \) indicates the auto-covariance of \( \bm{f}_* \), and \( \bm{K}_{f_*f_\mathcal{Z}} = \mathrm{cov}_{p(\bm f)}[\bm{f}_*, \bm{f}_{\mathcal{Z}}] \) indicates the cross-covariance between \( \bm{f}_* \) and \( \bm{f}_{\mathcal{Z}} \).

\subsection{Gaussian Process State Space Models (GPSSMs)}

Consider the following discrete-time state space model (SSM) \cite{zheng2024recursive}:

\begin{equation}\label{eq:SSM}
\begin{aligned}
    \bm{x}_{t+1} &= \bm{F}(\bm{x}_t, \bm c_t, \bm{f}(\bm x_t, \bm c_t)) + \bm{\omega}_p, 
    & \bm{\omega}_p &\sim \mathcal N(\bm{0}, \bm{\Sigma}_p) \\
    \bm{y}_t &= \bm{g}(\bm{x}_t) + \bm{\omega}_m, 
    & \bm{\omega}_m &\sim \mathcal N(\bm{0}, \bm{\Sigma}_m)
\end{aligned}
\end{equation}
where the vector $\bm{x} \in \mathbb{R}^{d_x}$, $\bm{c} \in \mathbb{R}^{d_c}$ and $\bm{y} \in \mathbb{R}^{d_y}$ respectively denote the system state, control input and measurement. The vectors $\bm{\omega}_p\in \mathbb{R}^{d_x}$ and $\bm{\omega}_m \in \mathbb{R}^{d_y}$ represent the zero-mean process and measurement noise, with covariances $\bm{\Sigma}_p$ and $\bm{\Sigma}_m$, respectively.
Additionally, the function $\bm{g}: \mathbb{R}^{d_x} \to \mathbb{R}^{d_y}$ represents the known measurement function.
The function $\bm{F}: \mathbb{R}^{d_x} \times \mathbb{R}^{d_c} \times \mathbb{R}^{d_f} \to \mathbb{R}^{d_x}$ represents a known structure for the transition model, which includes a unknown $d_f$-dimensional latent function
$\bm f(\bm x, \bm c) = [f^k(\bm z^k)]_{k=1}^{d_f}$. 
Here, the variable $\bm z^k \in \mathbb R^{d_{z^k}}$ is the input to the $k$-th function dimension, which can be obtained via a known linear mapping, i.e., $\bm z^k = \varphi^k(\bm x, \bm c)$.
Note that throughout this paper, the function output dimensions are indexed using superscripts, e.g., $f^k$ and $\bm z^k$.
Besides, for simplicity and without loss of generality, we omit the control input $\bm c$ in the following derivation.

To obtain accurate SSMs, data-driven methods such as GPs have attracted significant attention because of their data efficiency and ability to quantify uncertainty. For the SSM described in \eqref{eq:SSM}, we can construct the following Gaussian Process State Space Models (GPSSMs) \cite{frigolaBayesianTimeSeries2015,zheng2024recursive}:

\begin{subequations}\label{eq:GPSSM}
    \begin{align}
        \bm{x}_0 
        &\sim p(\bm{x}_0) \\
        \bm f(\cdot) &\sim
        \mathcal{GP}(\bm 0, \bm K_0(\cdot, \cdot; \bm \theta)) \\
        \label{eq:tran_density}
        \bm{x}_{t+1} | \bm{x}_{t}, \bm f 
        &\sim \mathcal N \left(\bm{x}_{t+1} \, \big| \,  \bm{F}(\bm{x}_t, 
        \bm f(\bm x_t)), 
        \, \bm{\Sigma}_p \right) \\
        \label{eq:meas_density}
        \bm{y}_t | \bm{x}_t 
        &\sim \mathcal N \left(\bm{y}_t \, \big| \,  \bm{g}(\bm{x}_t), \, \bm{\Sigma}_m \right)
    \end{align}
    \end{subequations}
Here, $p(\bm{x}_0)$ represents a Gaussian prior distribution for the initial state $\bm{x}_0$, and $\bm f(\bm x_t) = [f^k(\varphi^k(\bm x_t))]_{k=1}^{d_f}$. In addition, $\bm{f} = [\bm f^k]_{k=1}^{d_f}$ and $\bm f^k = \{f^k(\bm x): \bm x \in \mathbb R^{d_x} \}$ denotes all the function values for the $k$-th dimension, which can be regarded as an infinite dimensional vector.


\subsection{Recursive Gaussian Process State Space Model (RGPSSM)}

Based on the model \eqref{eq:GPSSM}, there have been various offline learning methods \cite{doerrProbabilisticRecurrentStateSpace2018,Ialongo2019, lindinger2022laplace, Lin2024} for the latent function $\bm f(\cdot)$ from the measurement sequence $\bm y_{1:t}$. 
However, in engineering scenarios, shifts in the SSMs commonly occur due to equipment aging or changes in the environment or task, making online learning of SSMs necessary.
In the online setting, the measurement $\bm y$ arrives sequentially, and the inference task of GPSSMs can be defined as sequentially approximating the filtering distribution \( p(\bm{x}_t, \bm{f}|\bm{y}_{1:t}) \) by integrating the previous result \( p(\bm{x}_{t-1}, \bm{f}|\bm{y}_{1:t-1}) \) and the current measurement \( \bm{y}_t \).
To enable flexible and efficient online learning of GPSSMs, the Recursive Gaussian Process State Space Model (RGPSSM) method was recently proposed \cite{zheng2024recursive}, which achieves recursive inference by the following two-step format:

\begin{align}
    \label{eq:pred_step}
    \text{Prediction Step:} \quad
    &p(\bm{f}, \bm{x}_{t+1}|\bm{y}_{1:t}) = 
    \int  p(\bm{x}_{t+1}|\bm{x}_t, \bm{f}) 
    p(\bm{f}, \bm{x}_{t}|\bm{y}_{1:t}) \mathrm{d}\bm{x}_t \\
    \label{eq:corr_step}
    \text{Correction Step:} \quad
    &p(\bm{f}, \bm{x}_{t+1}|\bm{y}_{1:t+1}) = \dfrac{p(\bm{y}_{t+1}|\bm{x}_{t+1}) 
    p(\bm{f}, \bm{x}_{t+1}|\bm{y}_{1:t})}{p(\bm{y}_{t+1}|\bm{y}_{1:t})} 
\end{align}
where the prediction step aims to obtain the joint prior $p(\bm{f}, \bm{x}_{t+1}|\bm{y}_{1:t})$ by using the transition model \eqref{eq:tran_density}, while the correction step aims to obtain the joint posterior $p(\bm{f}, \bm{x}_{t+1}|\bm{y}_{1:t+1})$ through Bayesian inference. 
However, due to the model nonlinearity and the nonparametric nature of the GP (that is, $\bm f$ is infinite dimensional), exact inference for these two steps is intractable. 
To address these challenges, the RGPSSM method \cite{zheng2024recursive} applies first-order linearization to both the transition and measurement models, leading to a factorization theorem  summarized below:

\begin{definition}[First-order linearization]\label{defn1}
    Consider a Gaussian probabilistic model
    \[
    p(\bm a \mid \bm b) = \mathcal{N}\big(\bm a \mid \bm h(\bm b),\, \bm S\big),
    \]
    where $\bm h(\cdot)$ is a differentiable mapping and $\bm S$ is the covariance matrix. 
    The first-order linearization of this model around the mean $\bm m_b$ is defined as
    \[
    p(\bm a \mid \bm b) \approx 
    \mathcal{N}\!\left(\bm a \,\middle|\, \bm h(\bm m_b) + 
    \frac{\partial \bm h}{\partial \bm b}\bigg|_{\bm m_b} (\bm b - \bm m_b),\, \bm S\right).
    \]
\end{definition}

\begin{theorem}\label{theorem1}
    Let $q(\bm{f}, \bm{x}_t)$ denote an approximation to the joint distribution $p(\bm{f}, \bm{x}_t)$ obtained in the prediction step \eqref{eq:pred_step} and the correction step \eqref{eq:corr_step}, where, for brevity, $p(\bm{f}, \bm{x}_t)$ represents either $p(\bm{f}, \bm{x}_{t+1} \mid \bm{y}_{1:t})$ or $p(\bm{f}, \bm{x}_{t+1} \mid \bm{y}_{1:t+1})$.
    Let $\bm{u} \triangleq [ \bm u_i ]_{i=1}^{n_u}$ denote a finite set of inducing points, where each inducing point $\bm u_i = \bm f(\bm{x}_i) = [f^k(\varphi^k(\bm x_i))]_{k=1}^{d_f}$ is the function value evaluated at $\bm{x}_i$. 
    The set $\bm u$ can be initialized as an arbitrary finite set or as the empty set.
    
    In each prediction step and correction step, by applying the first-order linearization described in Definition \ref{defn1} to both the transition model \eqref{eq:tran_density} and the measurement model \eqref{eq:meas_density}, and by augmenting the set of inducing points $\bm{u}$ with the function value $\bm{f}(\mathbb{E}_{q(\bm{u}, \bm{x}_t)}[\bm{x}_t])$, the approximate distribution $q(\bm{f}, \bm{x}_t)$ takes the following factorized form:
        \begin{equation}\label{eq:approx_dist0}
            q(\bm{f}, \bm{x}_t) = p(\bm{f}_{\ne u}|\bm{u}) q(\bm{u}, \bm{x}_t)
        \end{equation}
    where $\bm f_{\ne u} = \bm{f} \setminus \bm{u}$ denotes all function values except $\bm u$, and $p(\bm{f}_{\ne u}|\bm{u})$ denotes the GP prior conditional, analogous to \eqref{eq:GP_cond}:
        \begin{equation}\label{eq:cond_prior}
        \begin{aligned}
        p(\bm{f}_{\ne u}|\bm{u})
        &= \mathcal N \left( 
            \bm{K}_{f_{\ne u}u} \bm{K}_{uu}^{-1} \bm{u}, \,
            \bm{K}_{f_{\ne u}f_{\ne u}} - \bm{K}_{f_{\ne u}u} \bm{K}_{uu}^{-1} \bm{K}_{uf_{\ne u}}
        \right)
        \end{aligned}
        \end{equation}
    Furthermore, $q(\bm{u}, \bm{x}_t)$ in \eqref{eq:approx_dist0} is a Gaussian distribution.

\end{theorem}


This theorem indicates that, through first-order linearization, the joint distribution $q(\bm f, \bm x)$ can be factorized and represented by a finite-dimensional Gaussian distribution $q(\bm u, \bm x)$. 
Therefore, it enables tractable inference for the prediction and correction steps via matching the statistical moments (mean and covariance) of $q(\bm u, \bm x)$. 
Based on this theoretical foundation, the RGPSSM method \cite{zheng2024recursive} introduces two techniques to improve the practicality of the learning algorithm, specifically:

(1) Adaptation of the inducing-point set: As described in Theorem~\ref{theorem1}, the inducing-point set needs to be continuously augmented to maintain the factored approximation. 
However, this process causes the dimension of the moments for $q(\bm u, \bm x)$ to continuously increase, and eventually results in an unacceptable computational burden. 
To address this, a management strategy for inducing points is developed based on the informative criterion, which includes rules for both adding points and removing points.

(2) Adaptation of GP hyperparameters: By recovering the likelihood model from the posterior distribution $q(\bm u, \bm x)$, the RGPSSM method enables online optimization of GP hyperparameters $\bm\theta$ without the need to restore historical data. This technique reduces the burden of hyperparameter tuning at algorithm initialization and improves learning accuracy.

\begin{figure}[h]
    \centering
    \includegraphics[width=0.99\textwidth]{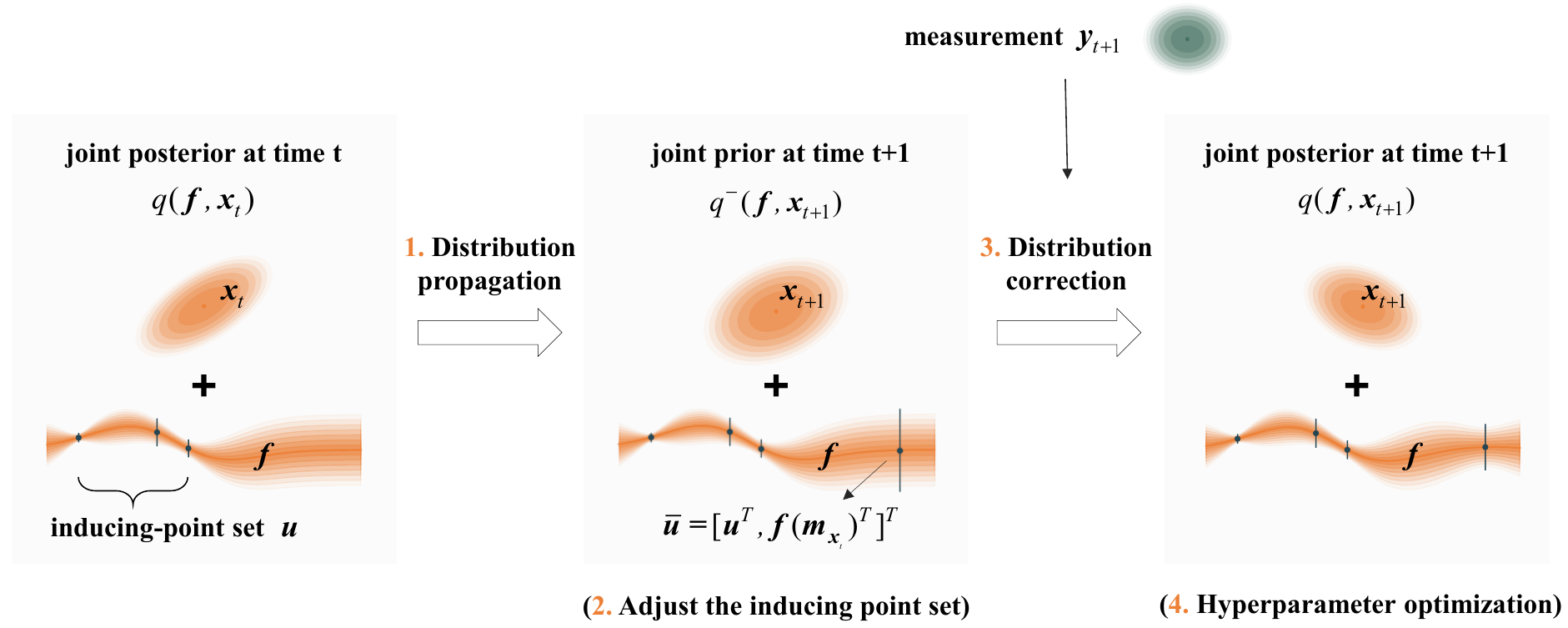}
    \caption{Algorithmic workflow of the RGPSSM \cite{zheng2024recursive}, where $\bm m_{x_t} = \mathbb{E}_{q(\bm{u}, \bm{x}_t)}[\bm{x}_t]$.}
    \label{fig:alg_flow}
\end{figure}

By integrating these two techniques, the complete RGPSSM algorithm consists of four procedures: the prediction step \eqref{eq:pred_step}, adjustment of the inducing-point set, the correction step \eqref{eq:corr_step}, and hyperparameter optimization, as illustrated in Fig.~\ref{fig:alg_flow}.

\subsection{Problem Statement}\label{subsec:problem}
With Theorem~\ref{theorem1} and the support of the two adaptation techniques, the RGPSSM framework achieves both sound theoretical foundations and flexible learning capabilities. 
Nevertheless, its flexibility and accuracy can be further improved through the following two aspects:

\textbf{(1) Heterogeneous multi-output transition function learning.} 
To enable the learning of the multi-output latent function $\bm f(\cdot)$ in SSMs \eqref{eq:SSM}, the RGPSSM applies the linear model of coregionalization (LMC) \cite{alvarez2012kernels} method to construct the multi-output kernel \cite{alvarez2012kernels}. Specifically, the kernel is constructed as $\bm{K}_0(\bm z_1, \bm z_2; \bm\theta) = {K}_0(\bm z_1, \bm z_2; \bm\theta) \bm{\Xi}$, where ${K}_0: \mathbb R^{d_z} \times \mathbb R^{d_z} \to \mathbb R$ is a single-output kernel and $\bm{\Xi}$ is the signal covariance matrix.
To distinguish it from the kernel introduced in this study, we refer to it as the isomorphic multi-output kernel. 
This isomorphic kernel is widely used and helps to simplify the RGPSSM algorithm, but it limits flexibility for heterogeneous multi-output function learning. In this paper, we define the heterogeneous multi-output learning setting as allowing different input variables $\bm z^k$, kernel types $K_0^k(\cdot, \cdot)$, and hyperparameters $\bm\theta^k$ for different function dimensions 
\footnote{
The need for heterogeneous multi-output function learning is common. For example, in a three-axis aircraft aerodynamic moment coefficient model, the input variables and the input-output sensitivities may differ for each axis.
}. It is evident that, the isomorphic kernel in the original RGPSSM method cannot accommodate this requirement.

\textbf{(2) Enhance learning accuracy for high-order nonlinear dynamics.} 
In RGPSSM, the first-order linearization leads to the fundamental factorization theorem~\ref{theorem1} and an Extended Kalman Filter (EKF)-like moment matching mechanism for lightweight learning \cite{zheng2024recursive}. However, the EKF only maintains high accuracy when the models exhibit low-order nonlinearity or when the uncertainty in $q(\bm u, \bm x)$ is small. 
Therefore, in scenarios with high-order nonlinearity or significant noise, the learning accuracy and the uncertainty quantification quality are significantly reduced, which can compromise the safety of downstream applications.

To improve the two capabilities mentioned above, we first present a framework to support heterogeneous multi-output latent function learning and general moment matching techniques in Section \ref{sec:heterogeneous}.
Subsequently, the application details of the three moment matching methods are provided in Section \ref{sec:ukf_adf}.

\section{Extension for Heterogeneous Learning and
General Moment Matching Methods}\label{sec:heterogeneous}

This section extends the RGPSSM in \cite{zheng2024recursive} to enable learning of heterogeneous multi-output functions and the application of general moment matching techniques. We first propose a heterogeneous multi-output kernel, then present a unified inference framework for general moment matching methods, and finally modify the inducing-point management method for the heterogeneous learning setting.

\subsection{Heterogeneous Multi-output Kernel}

Since the random variables inferred in RGPSSM are the function values $\bm f$, the learning algorithm is independent of the form of the function inputs and the kernel.
Therefore, to achieve heterogeneous multi-output learning, the core change is to extend the kernel function to accommodate this setting.
To address this, this paper designs a heterogeneous multi-output kernel, which expresses the prior covariance $\bm K_{f_1f_2} = \mathrm{cov}_{p(\bm f)}[\bm f_1, \bm f_2]$ between the two sets of function values $\bm f_1 = [\bm f_1^k]_{k=1}^{d_f}$ and $\bm f_2 = [\bm f_2^k]_{k=1}^{d_f}$ as:

\begin{equation}\label{eq:S0}
    \begin{aligned}
        \bm K_{f_1f_2} &=
        \begin{bmatrix}
            \bm K_{f^1_1f^1_2}  & \bm 0 & \cdots & \bm 0\\
            \bm 0 & \bm K_{f^2_1f^2_2} & \cdots & \bm 0 \\
            \vdots & \vdots & & \vdots \\
            \bm 0 & \bm 0 & \cdots & \bm K_{f^{d_f}_1f^{d_f}_2}
        \end{bmatrix} \\
    \end{aligned}
\end{equation}
Here, $\bm f_i^k$ for $i=1,2$ denotes the set of function values for the $k$-th dimension. In addition, $\bm K_{f^k_1f^k_2} = \mathrm{cov}_{p(\bm f^k)}[\bm f^k_1, \bm f^k_2]$ represents the prior covariance for the $k$-th function dimension, which can be computed via the corresponding single-output kernel function $K_0^k(\cdot, \cdot; \bm\theta^k)$:

\begin{equation}
\begin{aligned}
\bm K_{f^k_1f^k_2}
&= K_0^k(\mathcal{Z}^k_1, \mathcal{Z}^k_2; \bm\theta^k) \\
\end{aligned}
\end{equation}
where $\mathcal{Z}^k_i$ is the set of inputs corresponding to the $k$-th dimension function values $\bm f_i^k$.
Note that, here we assume the different function dimensions are independent in the prior, and $\bm f$ is stacked such that all values corresponding to the same function dimension are grouped together, which leads to a block diagonal kernel matrix, as shown in \eqref{eq:S0}.
When the set $\bm f$ is arranged arbitrarily, the kernel matrix can still be obtained by a straightforward procedure: first, reorder $\bm f$ according to function dimension to compute the kernel matrix; then, permute the rows and columns of the kernel matrix back to the original order.

Through the structure in \eqref{eq:S0}, this multi-output kernel allows the use of different kernel types ($K_0^k(\cdot,\cdot)$), kernel hyperparameters ($\bm\theta^k$), and input variables ($\bm z^k = \varphi^k(\bm x)$) for each function dimension. Therefore, this heterogeneous kernel possesses the following advantages. First, the inductive bias for different dimensions, such as input-output dependence, can be incorporated into the prior model via kernel types. 
Second, this kernel allows learning different kernel hyperparameters for each dimension, thus enhancing learning accuracy and generalization.
Third, it can adapt to cases where the input variables differ across different function dimensions, thus improving practical applicability.
In terms of the application of this kernel, one can replace the isomorphic kernel in the original RGPSSM method with the heterogeneous kernel \eqref{eq:S0}, thereby extending the method to a heterogeneous learning setting.

However, in the original RGPSSM method, an inducing point is defined as the all-dimensional function value $\bm f(\bm x) = \left[f^k(\bm x) \right]_{k=1}^{d_f}$, which means that the selection of inducing points is synchronized for all dimensions. 
To leverage the different characteristics of each dimension, it is preferable to independently select the inducing points for different dimensions, thereby improving both computational efficiency and learning accuracy.
In the following, we first present the necessary background on the GP prediction equation, and then derive a unified inference framework for general moment matching techniques. Finally, we provide the new inducing-point management method to enable independent addition and discarding operations for each dimension.


\subsection{GP Prediction Equation in the RGPSSM}

In RGPSSM, for an arbitrary set of function values $\bm f_*$, the posterior distribution $q(\bm f_*, \bm x_t) = \int p(\bm f_{\ne u}|\bm u) q(\bm u, \bm x_t) \, \mathrm{d} \bm f_{\ne *}$ can be evaluated by using $q(\bm u, \bm x_t)$. First, we express the moments of $q(\bm{u}, \bm{x}_t)$ and $q(\bm f_*, \bm x_t)$ as follows:

\begin{equation}\label{eq:approx_dist}
    \begin{aligned}
    q(\bm{u}, \bm{x}_t) & = \mathcal{N}\left(
    \begin{bmatrix} \bm{u} \\ \bm{x}_t \\ \end{bmatrix} \Bigg |
    \begin{bmatrix} \bm{m}_{u} \\ \bm{m}_{x_t} \\ \end{bmatrix},
    \begin{bmatrix}
    \bm{S}_{uu} & \bm{S}_{ux_t} \\ 
    \bm{S}_{ux_t}^T & \bm{S}_{x_t x_t} \\ 
    \end{bmatrix} \right) \\ 
    q(\bm{f}_*, \bm{x}_t) & = \mathcal{N}\left(
    \begin{bmatrix} \bm{f_*} \\ \bm{x}_t \\ \end{bmatrix} \Bigg |
    \begin{bmatrix} \bm{m}_{f_*} \\ \bm{m}_{x_t} \\ \end{bmatrix},
    \begin{bmatrix}
    \bm{S}_{f_*f_*} & \bm{S}_{f_* x_t} \\ 
    \bm{S}_{f_* x_t}^T & \bm{S}_{x_t x_t} \\ 
    \end{bmatrix} \right)
\end{aligned}
\end{equation}
In this paper, we consistently use the notation $\bm m$ to represent the mean of random variables and $\bm S$ to denote the covariance. And the subscripts of $\bm m$ and $\bm S$ indicate the corresponding random variables.
Using this notation, the moments of $q(\bm{f}_*, \bm{x}_t)$ can be expressed as follows:

\begin{equation}\label{eq:qf}
    \begin{aligned}
    &\bm{m}_{f_*} = \bm{K}_{f_*u} \bm{K}_{uu}^{-1} \bm{m}_u \\
    &\bm{S}_{f_* x_t} = \bm{K}_{f_* u} \bm{K}_{uu}^{-1} \bm{S}_{ux_t} \\
    &\bm{S}_{f_*f_*} = \bm{K}_{f_*f_*} + \bm{K}_{f_*u} \bm{K}_{uu}^{-1} 
    \left( \bm{S}_{uu} - \bm{K}_{uu} \right) \bm{K}_{uu}^{-1} \bm{K}_{uf_*} 
\end{aligned}
\end{equation}
which can be derived by using the factorized approximation \eqref{eq:approx_dist0} and the GP prior conditional \eqref{eq:cond_prior}. 
Based on this GP prediction equation, we now present the unified inference framework and the inducing-point management method.

\subsection{Unified Inference Framework Applicable to General Moment Matching Methods}\label{subsec:inference}

To achieve tractable inference, the original RGPSSM method \cite{zheng2024recursive} applies first-order linearization to the transition and measurement models, resulting in EKF-based moment matching equations for practical implementation.
However, this EKF-based method only provides first-order approximation accuracy, which can lead to significant loss of accuracy under strong nonlinearity and large uncertainty. 
To address this issue, this paper aims to apply more advanced moment matching methods, such as the sigma-point filter \cite{van2001square}.
However, the factorization theorem \ref{theorem1} is derived based on first-order linearization. There lacks a rigorous theoretical foundation and a straightforward framework for applying other moment matching methods. To address this limitation, this paper develops a new approximate inference framework that enables the use of general moment matching techniques. 
The derivation begins with the correction step \eqref{eq:corr_step}, which is straightforward, and is followed by the prediction step \eqref{eq:pred_step}.

First, for the correction step \eqref{eq:corr_step}, it can be easily shown that it is compatible with general moment matching techniques. Specifically, assume that the approximate joint prior $q^-(\bm f, \bm x_{t+1}) \approx p(\bm f, \bm x_{t+1}|\bm y_{1:t})$ attained in the prediction step \eqref{eq:pred_step} follows the factorized form \eqref{eq:approx_dist0}:

\begin{equation}
\begin{aligned}
    q^-(\bm f, \bm x_{t+1}) = p(\bm f_{\ne u}|\bm u) q^-(\bm u, \bm x_{t+1})
\end{aligned}
\end{equation}
where the superscript "$-$" denotes the predicted distribution, as is common in the Kalman filter literature. 
Then, substituting it into the correction step \eqref{eq:corr_step}, we have:


\begin{equation}\label{eq:corr_eq}
\begin{aligned}
    p(\bm f, \bm x_{t+1}|\bm y_{t+1}) 
    &\approx \dfrac{p(\bm y_{t+1}| \bm x_{t+1}) 
    p(\bm f_{\ne u}|\bm u) q^-(\bm u, \bm x_{t+1})}
    {q(\bm y_{t+1})} \\
    &= p(\bm f_{\ne u}|\bm u) 
    \left[ \dfrac{p(\bm y_{t+1}| \bm x_{t+1}) q^-(\bm u, \bm x_{t+1})}
    {q(\bm y_{t+1})} \right] \\
    &\approx p(\bm f_{\ne u}|\bm u) 
    q(\bm u, \bm x_{t+1})
    \\
\end{aligned}
\end{equation}
Therefore, general moment matching techniques can be directly used to approximate the finite-dimensional inference $q(\bm u, \bm x_{t+1}) = \tfrac{p(\bm y_{t+1}| \bm x_{t+1}) q^-(\bm u, \bm x_{t+1})}
{q(\bm y_{t+1})}$. 
Since the application is straightforward, we do not discuss its details further in the following.

Second, for the prediction step \eqref{eq:pred_step}, by substituting the factorized approximation \eqref{eq:approx_dist0} into \eqref{eq:pred_step}, we obtain:

\begin{equation}\label{eq:pred_eq0}
    \begin{aligned}
        p(\bm{f}, \bm{x}_{t+1}|\bm{y}_{1:t}) 
        &\approx \int  p(\bm{x}_{t+1}|\bm{x}_t, \bm{f}) 
        p(\bm f_{\ne u}|\bm u)
        q(\bm u, \bm x_t) 
        \mathrm{d}\bm{x}_t \\
\end{aligned}
\end{equation}
It can be seen that, since the transition model $p(\bm{x}_{t+1}|\bm{x}_t, \bm{f})$ depends on the function values $\bm f$, we cannot directly separate $p(\bm f_{\ne u}|\bm u)$ as in \eqref{eq:corr_eq}.
To obtain a factorized approximation, we introduce the latent function prediction $\bm f(\bm x_t)$ as a random variable $\bm h_t$, that is, $\bm h_t = \bm f(\bm x_t)$. Therefore, \eqref{eq:pred_eq0} can be rewritten as:

\begin{equation}\label{eq:p_h1}
\begin{aligned}
    p(\bm{f}, \bm{x}_{t+1}|\bm{y}_{1:t})
    &\approx \int  p(\bm{x}_{t+1}|\bm{x}_t, \bm{h}_t) \,
    p(\bm h_t| \bm f, \bm x_t) p(\bm f_{\ne u}|\bm u) \,
    q(\bm u, \bm x_t) 
    \,\mathrm{d}\bm h_t \mathrm{d}\bm{x}_t \\
\end{aligned}
\end{equation}
where the function prediction conditional $p(\bm h_t| \bm f, \bm x_t) = \mathcal N(\bm K_{h_t f} \bm K_{ff}^{-1} \bm f, \bm K_{h_t h_t} - \bm K_{h_t f} \bm K_{ff}^{-1} \bm K_{f h_t})$ is the GP prior conditional analogous to \eqref{eq:GP_cond}, and the state prediction conditional is:
\begin{equation}\label{eq:state_pred_model}
\begin{aligned}
    p(\bm{x}_{t+1}|\bm{x}_t, \bm{h}_t) = 
    \mathcal{N}(\bm x_{t+1}|\bm F(\bm x_t, \bm h_t), \bm\Sigma_p)
\end{aligned}
\end{equation}

Next, we address the issue arising from the function prediction conditional $p(\bm h_t | \bm f, \bm x_t)$, which depends on the infinite-dimensional function values $\bm f$. For clarity, we denote $\bm f_t = \bm f(\bm m_{x_t})$ and introduce the following approximation (its rationale will be discussed later):

\begin{equation}\label{eq:approx_ph}
\begin{aligned}
    p(\bm h_t| \bm f, \bm x_t) 
    &= p(\bm h_t|\bm f_{\ne (f_t, u)}, \bm f_t, \bm u, \bm x_t) \\
    &\approx p(\bm h_t| \bm f_t, \bm u, \bm x_t)
\end{aligned}
\end{equation}
This means that $\bm f_t$ and $\bm u$ together carry sufficient information for $\bm h_t$, and thereby $\bm h_t$ is conditionally independent of $\bm f_{\ne (f_t, u)}$ given $\bm f_t$ and $\bm u$.
Therefore, the function prediction conditional is approximated to only rely on finite dimensional function values, thereby can make the inference tractable. 

Based on the approximation \eqref{eq:approx_ph}, we can augment $\bm f_t$ into the inducing-point set $\bm u$, that is, $\bar{\bm{u}} = [\bm{u}^T, \bm f_t^T]^T$, and substitute \eqref{eq:approx_ph} into \eqref{eq:p_h1} to obtain the approximate inference framework:

\begin{equation}
\begin{aligned}
    p(\bm{f}, \bm{x}_{t+1}|\bm{y}_{1:t})
    &\approx \int  p(\bm{x}_{t+1}|\bm{x}_t, \bm{h}_t) \,
    p(\bm h_t| \bm f_t, \bm u, \bm x_t) \,
    p(\bm f_{\ne u}|\bm u) \,
    q(\bm u, \bm x_t) 
    \,\mathrm{d}\bm h_t \mathrm{d}\bm{x}_t \\
    &= \int  p(\bm{x}_{t+1}|\bm{x}_t, \bm{h}_t) \,
    p(\bm h_t| \bar{\bm u}, \bm x_t) \,
    p(\bm f_{\ne \bar u}|\bar{\bm u}) \,
    p(\bm f_t|\bm u) \,
    q(\bm u, \bm x_t) 
    \,\mathrm{d}\bm h_t \mathrm{d}\bm{x}_t \\
    &= p(\bm f_{\ne \bar u}|\bar{\bm u}) 
    \left[
    \int  p(\bm{x}_{t+1}|\bm{x}_t, \bm{h}_t) \,
    p(\bm h_t| \bar{\bm u}, \bm x_t) \,
    q(\bar{\bm u}, \bm x_t) 
    \,\mathrm{d}\bm h_t \mathrm{d}\bm{x}_t 
    \right] \\
    &= p(\bm f_{\ne \bar u}|\bar{\bm u})
    q^-(\bar{\bm u}, \bm x_{t+1}) 
\end{aligned}
\end{equation}
where $q(\bar{\bm u}, \bm x_t) = p(\bm f_t|\bm u) q(\bm u, \bm x_t)$ and $q^-(\bar{\bm u}, \bm x_{t+1})$ is a Gaussian approximation of the inference result inside the brackets.
Therefore, we obtain a new approximate inference framework that maintains the factorized form \eqref{eq:approx_dist0} without relying on linearization, which consists of two steps:

(1) Augment the inducing-point set as $\bar{\bm{u}} = [\bm{u}^T, \bm f_t^T]^T$ and compute the moments of the new joint distribution $q(\bar{\bm u}, \bm x_t)$, which can be obtained from \eqref{eq:qf} by setting $\bm f_* = \bar{\bm u}$.

(2) Propagate the moments of the distribution $q(\bar{\bm u}, \bm x_t)$ to obtain the approximate predicted distribution $q^-(\bar{\bm u}, \bm x_{t+1})$, namely:
\begin{equation}\label{eq:pred_task}
\begin{aligned}
    q^-(\bar{\bm u}, \bm x_{t+1}) &\approx
    \int
    p(\bm{x}_{t+1}|\bm{x}_t, \bm{h}_t) 
    p(\bm h_t| \bm x_t, \bar{\bm u})
    q(\bar{\bm u}, \bm x_t) 
    \mathrm{d}\bm h_t \mathrm{d}\bm{x}_t 
\end{aligned}
\end{equation}
In this paper, we provide EKF-, Unscented Kalman Filter (UKF)-, and Assumed Density Filter (ADF)-based approaches for the second step, which are described in detail in Section \ref{sec:ukf_adf}.

Thus far, we have derived a tractable approximate inference framework that maintains the factorized form \eqref{eq:approx_dist0}. This framework does not rely on first-order linearization and is therefore applicable to any moment matching technique.
In the derivation, only the approximation \eqref{eq:approx_ph} is introduced, and we now analyze its reasonableness. Qualitatively, when the function values $\bm f_t$ and $\bm u$ can provide sufficient information for $\bm h_t = \bm f(\bm x_t)$, the approximation is accurate. This condition is satisfied in most cases. 
Specifically, on the one hand, when the uncertainty of $\bm x_t$ is small, $\bm h_t$ is statistically close to $\bm f_t = \bm f(\bm m_{\bm x_t})$, and $\bm f_t$ can provide sufficient information for $\bm h_t$.
On the other hand, after a period of learning, the inducing-point set $\bm u$ can be sufficiently distributed in the space and thus can also provide adequate information. Therefore, only in the early stage of learning and when the uncertainty of $\bm x_t$ is large, this approximation may have a large error. 
According to our experimental results in Section \ref{sec:experiment}, this approximation error does not significantly affect the ability to achieve accurate learning, even in the case of large noise.

In summary, this subsection presents a new inference framework for the RGPSSM that does not rely on first-order linearization. This framework divides the prediction step into two parts: augmenting the inducing-point set and performing moment matching for the joint variable $(\bar{\bm u}, \bm x)$. In the next subsection, we extend the inducing-point management method to support independent addition and discarding operations for different function dimensions.

\subsection{Inducing Points Management}\label{subsec:inducing_points_management}

As shown in Section \ref{subsec:inference}, the inducing-point set $\bm u$ needs to be continuously augmented in order to preserve the factorized form \eqref{eq:approx_dist0}. This augmentation results in a rapid increase in the dimension of the moments of $q(\bm u, \bm x)$, which in turn leads to higher computational cost.
To address this, the original RGPSSM method \cite{zheng2024recursive} developed an inducing-point management strategy that primarily consists of adding and discarding operations.
Specifically, (1) adding rule: new inducing points are added when the novelty of the new points exceeds a certain threshold $\varepsilon_{tol}$; (2) discarding rule: the least important points are discarded once the size of $\bm u$ exceeds the budget $M$, where the importance is quantified by a score criterion $s_d$.
Through this mechanism, the growth of the inducing-point set can be slowed and its maximum size can be limited within the budget $M$, thereby maintaining manageable computational complexity.

However, in the original RGPSSM method, for implementation convenience, the multi-output kernel is assumed to be isomorphic, and an inducing point is defined as the all-dimensional function values $\bm f(\bm x) = [f^k(\bm x)]_{k=1}^{d_f}$. 
When employing a heterogeneous multi-output kernel \eqref{eq:S0}, inducing points must be selected uniformly in order to accommodate the most complex function dimension that requires numerous points. Therefore, the original approach will reduce both computational efficiency and representational flexibility.
To address this limitation, we aim to perform the addition and discarding of inducing points independently for each output dimension.
For convenience in the derivation, we denote the inducing point to be deleted as $u_\mathrm{d}$, which is a scalar in this paper, and the remaining inducing points as $\bm u_{\mathrm{l}}$. 
In the following, we show that the deletion operation can basically be inherited from the original method, while the addition operation requires some modifications.
Moreover, a pruning method for redundant inducing points, inspired by the new adding criterion, is developed to enhance computational efficiency and numerical stability.

\textbf{Rule for discarding inducing points when the size of the inducing-point set exceeds the budget.}
To make the inference tractable after discarding inducing points, \cite{zheng2024recursive} provides the following definition. The approximate distribution after discarding is assumed to belong to the factorized distribution family $\hat{q}(\bm{f}, \bm{x}) = p(\bm{f}_{\ne u_{\mathrm{l}}}|\bm{u}_{\mathrm{l}}) \hat{q}(\bm{u}_{\mathrm{l}}, \bm{x})$.
Based on this definition, \cite{zheng2024recursive} derive the optimal approximate distribution by  minimizing the inclusive KL divergence (\( \mathrm{KL}[q \| \hat{q}] = \int q \log \left( q/\hat{q} \right) \)) between the original and approximated distribution.
Specifically,

\begin{equation}\label{eq:res_proj}
\begin{aligned}
\hat{q}^*(\bm{u}_{\mathrm{l}}, \bm{x}) 
&= \arg\min\limits_{\hat{q}(\bm{u}_{\mathrm{l}}, \bm{x})} \mathrm{KL} 
\left[ q(\bm{f}, \bm{x}) \| \hat{q}(\bm{f}, \bm{x}) \right] \\
&= \arg\min\limits_{\hat{q}(\bm{u}_{\mathrm{l}}, \bm{x})} \mathrm{KL} 
\left[ p(\bm{f}_{\ne u}|\bm{u})q(\bm{u}, \bm{x}) \| p(\bm{f}_{\ne u_{\mathrm{l}}}|\bm{u}_{\mathrm{l}}) \hat{q}(\bm{u}_{\mathrm{l}}, \bm{x}) \right] \\
&= \arg\min\limits_{\hat{q}(\bm{u}_{\mathrm{l}}, \bm{x})} \mathrm{KL} 
\left[q(\bm{u}, \bm{x}) \| p(u_{\mathrm{d}}|\bm{u}_{\mathrm{l}}) \hat{q}(\bm{u}_{\mathrm{l}}, \bm{x})\right] \\
&= \int q(\bm{u}, \bm{x}) \, \mathrm{d} u_{\mathrm{d}}
\end{aligned}
\end{equation}
where $\hat{q}^*(\bm{u}_{\mathrm{l}}, \bm{x})$ denotes the optimal deletion distribution, whose moments can be easily obtained by removing the dimension corresponding to $u_d$.
Furthermore, by setting \( \hat{q}(\bm{u}_{\mathrm{l}}, \bm{x}) = \hat{q}^*(\bm{u}_{\mathrm{l}}, \bm{x}) \) in the third row of \eqref{eq:res_proj}, one can obtain a KL divergence that quantifies the accuracy loss incurred by discarding \( \bm u_{\mathrm{d}} \), specifically, \( D^* = \mathrm{KL} \left[q(\bm{u}, \bm{x}) \| p(u_{\mathrm{d}}|\bm{u}_{\mathrm{l}}) \hat{q}^*(\bm{u}_{\mathrm{l}}, \bm{x})\right] \). Utilizing this KL divergence, \cite{zheng2024recursive} derive the importance score for selecting the inducing point to discard:

\begin{equation}\label{eq:score}
    \begin{aligned}
    s_d 
    &= \Delta_1 + \Delta_2 + \Delta_3\\
    \Delta_1 &= \bm{m}_u^T \bm{Q}_{du}^T \bm{Q}_{dd}^{-1} \bm{Q}_{du} \bm{m}_u \\
    \Delta_2 &= \mathrm{tr}\left(\bm{Q}_{du} \bm{S}_{uu} \bm{Q}_{du}^T \bm{Q}_{dd}^{-1} \right) \\
    \Delta_3 &= \log\vert \bm{\Omega}_{dd} \vert - \log\vert \bm{Q}_{dd} \vert
    \end{aligned}
\end{equation}
where \( \bm{\Omega} = \bm{\Sigma}_t^{-1} \) is the inverse of the joint covariance $\bm{\Sigma}_t = \mathrm{var}[(\bm u, \bm x_t)]$. \( \bm{\Omega}_{dd} \) is the diagonal element of the matrix \( \bm{\Omega} \) corresponding to the discarded point \( u_{\mathrm{d}} \). In addition, \( \bm{Q}_{dd} \) and \( \bm{Q}_{du} \) are the diagonal element and the row of the inverse kernel matrix \( \bm{Q} = \bm{K}_{uu}^{-1} \) corresponding to the discarded point \( u_{\mathrm{d}} \). 
It can be seen that using the scalar $u_d$ in the above derivation presents no difficulty. Therefore, evaluating the score and deleting the inducing point in a scalar-wise manner can be directly inherited. 
As a result, the new deletion rule is as follows: evaluate the score for each scalar inducing point and remove the $n_u - M$ points with the lowest scores when the size of $\bm u$ exceeds the budget $M$.

\textbf{Rule for adding new inducing points.} 
For the adding operation, \cite{zheng2024recursive} propose to use the GP prior conditional variance $\bm\gamma = \mathrm{var}_{p(\bm f_t|\bm u)}[\bm f_t]$ as the selection criterion, where:

\begin{equation}
\begin{aligned}
    \bm\gamma = \bm K_{f_tf_t} - \bm K_{f_t u} \bm K_{uu}^{-1} \bm K_{u f_t}
\end{aligned}
\end{equation}
where $\bm\gamma$ is a $d_f \times d_f$ matrix. 
Theoretically, $\bm\gamma$ measures both approximation accuracy and numerical stability.
Specifically, \cite{zheng2024recursive} shows that as $\bm\gamma \to 0$, the score $s_d \to -\infty$, indicating that omitting the new point results in negligible accuracy loss. At the same time, $\bm\gamma$ is the Schur complement of the updated kernel matrix $\bm K_{\bar u \bar u}$ with respect to the block $\bm K_{uu}$, so if $\bm\gamma \to 0$, the updated kernel matrix becomes nearly singular. Therefore, \cite{zheng2024recursive} suggests adding the new points$\bm f_t$ (across all function dimensions) only when $\mathrm{tr}(\bm\gamma) > \varepsilon_{tol}$, where $\varepsilon_{tol}$ is a predefined threshold. 
This adding criterion is theoretically appropriate and practically effective, and we aim to adapt it to support independent addition operations for each dimension. This adaptation can be achieved by taking advantage of the special structure of the heterogeneous kernel in \eqref{eq:S0}. In particular, due to the prior independence among different function dimensions, the novelty matrix is diagonal, namely, $\bm\gamma = \mathrm{diag}([\gamma^k]_{k=1}^{d_f})$, where

\begin{equation}\label{eq:gam_k}
    \begin{aligned}
        \gamma^k = \bm K_{f_t^k f_t^k} - \bm K_{f_t^k u^k} \bm K_{u^k u^k}^{-1} \bm K_{u^k f_t^k}
    \end{aligned}
\end{equation}
Therefore, $\gamma^k$ serves as the accuracy and singularity metric for each dimension and can be used as the adding criterion. However, this criterion makes the selection of the adding threshold $\varepsilon_{tol}$ challenging.
Specifically, in the heterogeneous learning setting, different output dimensions may have different kernel types and hyperparameters, which makes it difficult to set a unified threshold $\varepsilon_{tol}$ for each dimension. 
Although for stationary kernels (such as the Gaussian kernel defined later in \eqref{eq:kernel_rbf}), the signal variance can be used to normalize $\gamma^k$ and thus achieve a unified threshold $\varepsilon_{tol}$, this approach is not suitable for non-stationary kernels\footnote{A \emph{stationary kernel} satisfies $K_0(\bm z_1,\bm z_2)=k(\bm z_1-\bm z_2)$, and its autocovariance $K_0(\bm z,\bm z)=k(0)$ is independent of $\bm z$. 
In contrast, a \emph{non-stationary kernel} may have autocovariance $K_0(\bm z,\bm z)$ that varies with $\bm z$. Therefore, stationary kernels admit a global signal variance (scale), $\sigma^2=k(0)$, whereas non-stationary kernels do not.}.
To address this, we use the singularity metric meaning of $\gamma^k$ to normalize the novelty $\gamma^k$ by the maximum diagonal element of the kernel matrix $\bm K_{u^k u^k}$, which leads to the following criterion:

\begin{equation}\label{eq:add_criterion}
\begin{aligned}
    \gamma^k / \mathrm{max}\left[ \mathrm{diag}(\bm K_{u^k u^k}) \right]> \varepsilon_{tol}
\end{aligned}
\end{equation}
Here, $\mathrm{max}\left[ \mathrm{diag}(\bm K_{u^k u^k}) \right]$ reflects the typical scale of the $k$-th dimensional kernel matrix $\bm K_{u^k u^k}$, allowing a unified threshold $\varepsilon_{tol}$ to be used across all function dimensions and thus keeping parameter tuning simple. 

\textbf{Rule for pruning the redundant inducing points.} 
This singularity-based criterion is effective (as demonstrated in Section \ref{sec:experiment}), and it inspires us to design a method for pruning redundant points.
Specifically, the RGPSSM involves online adaptation of the kernel hyperparameters, and there are cases where the length scale of the kernel increases during adaptation. In such cases, the kernel matrix $\bm K_{u^k u^k}$ can easily become singular, which means that some inducing points are no longer important. 
To address this, we apply the conditional variance in \eqref{eq:gam_k} to all retained inducing points. 
For example, for the $k$-th function dimension, if there exist inducing points $u_i^k$ that satisfy:
\begin{equation}\label{eq:singularity_test}
\begin{aligned}
    \gamma^k_i / \mathrm{max}\left[ \mathrm{diag}(\bm K_{u^k u^k}) \right] < \rho \cdot \varepsilon_{tol}, \quad \rho < 1 \\
\end{aligned}
\end{equation}
where $\gamma^k_i = \bm K_{u_i^k u_i^k} - \bm K_{u_i^k u^k} \bm K_{u^k u^k}^{-1} \bm K_{u^k u_i^k}$,
we discard the point with the smallest $\gamma^k_i$. 
Note that, here we use a smaller threshold, i.e., $\rho \cdot \varepsilon_{tol}$, than the adding criterion because the $\gamma^k_i$ for retained points can become smaller after the inducing-point set is augmented. We found that $\rho = 0.1$ is a suitable decay factor. In addition, at each step, only one point is discarded for each dimension, because the $\gamma^k_i$ for the remaining points can increase after discarding a point. 
In terms of implementation, this procedure can be performed efficiently because $\gamma^k_i$ is equal to the inverse of the $i$-th diagonal element of the inverse matrix $\bm K_{u^k u^k}^{-1}$ (this result can be found in Appendix A of \cite{zheng2024recursive}).
Since we store the Cholesky factor of $\bm K_{u^k u^k}$, its inverse matrix $\bm K_{u^k u^k}^{-1}$ can be efficiently computed.

Thus far, we have extended the original inducing-point management method to allow independent addition and discarding operations for different function dimensions. Specifically, we have shown that the discarding operation can be directly inherited from the original method, while the adding operation requires modification by using a normalized novelty measure. Furthermore, the singularity-based criterion is employed to prune redundant inducing points during the  hyperparameter adaptation, thereby improving computational efficiency and numerical stability.

In summary, this section extends the standard RGPSSM method to handle heterogeneous multi-output function learning and to support general moment matching techniques. First, a heterogeneous multi-output kernel is proposed to allow different GP priors for each dimension. Second, a unified inference framework is derived without using first-order linearization to support general moment matching methods. Third, the inducing points management algorithm is extended to enable independent selection for each dimension and to prune redundant inducing points.
In the next section, we will present the application details for several moment matching techniques.

\section{Three Moment Matching Methods for the Prediction Step}\label{sec:ukf_adf}

In this section, we re-derive the EKF-based moment matching method using the new approximate inference framework introduced in Section \ref{subsec:inference}, which provides new insights. Furthermore, we develop two more accurate moment matching techniques for the RGPSSM, including a UKF-based method for handling general nonlinearity and an ADF-based method, which provides exact moment matching for the Gaussian kernel case.

\textbf{Preliminaries.} 
As illustrated in Section \ref{subsec:inference}, the task of different moment matching methods is to solve the inference problem in \eqref{eq:pred_task}. 
It is important to note that, considering the inducing points management method described earlier, the set of inducing points $\bar{\bm u}$ in the inference task \eqref{eq:pred_task} has a new definition. Specifically, $\bar{\bm u}$ now refers to the inducing-point set that has been updated according to the addition criterion in Section \ref{subsec:inducing_points_management}. This may involve adding inducing points for all dimensions, for some dimensions, or not adding any at all.
In addition, for convenience in the following derivation, rewrite the function prediction conditional $p(\bm h_t| \bm x_t, \bar{\bm u})$ as follows:

\begin{equation}\label{eq:ph_xu}
\begin{aligned}
    &p(\bm h_t| \bm x_t, \bar{\bm u})
        = \mathcal N\left(\bm h_t \mid \bm \mu_{\mathrm{gp}}(\bm{x}_t, \bar{\bm u}), 
        \bm \Sigma_{\mathrm{gp}}(\bm{x}_t)\right) \\
    &\bm \mu_{\mathrm{gp}}(\bm{x}_t, \bar{\bm u}) = \bm{K}_{h_t \bar u}\bm{K}_{\bar u\bar u}^{-1} \bar{\bm u} \\
    &\bm \Sigma_{\mathrm{gp}}(\bm{x}_t) =  \bm{K}_{h_th_t} - \bm{K}_{h_t\bar u}\bm{K}_{\bar u\bar u}^{-1}\bm{K}_{\bar uh_t}\\
\end{aligned}
\end{equation}
where $\bm K_{h_th_t}$ and $\bm K_{h_t\bar u}$ are the function of $\bm x_t$, which can be evaluated by using the kernel \eqref{eq:S0} and the GP input mapping $\bm z = \bm\varphi(\bm x)$ as defined in \eqref{eq:SSM}.

\subsection{Extended Kalman Filter (EKF)-based Moment Matching}\label{subsec:ekf}

The EKF-based method approximates the inference problem in \eqref{eq:pred_step} by linearizing the transition model, thereby enabling closed-form moment matching. 
According to the inference problem definition, the state prediction conditional \eqref{eq:state_pred_model} and the function prediction conditional \eqref{eq:ph_xu}, the transition model for the state $\bm x$ can be written as follows:

\begin{equation}\label{eq:tran_x}
    \begin{aligned}
        &\bm x_{t+1}|\bm x_t, \bar{\bm u}, \bm\varepsilon \sim
        \mathcal{N} \left(\bm x_{t+1} \,\Big|\, 
        \bm F \left(
        \bm x_t, \bm\mu_{\mathrm{gp}}(\bm{x}_t, \bar{\bm u}) 
        + \bm\Sigma^{1/2}_{\mathrm{gp}}(\bm {x}_t) \bm \varepsilon
        \right)
        ,\, \bm\Sigma_p
        \right) \\
        &\bm \varepsilon \sim \mathcal{N}(\bm 0, \bm I_{d_f}) \\
    \end{aligned}
    \end{equation}
where the auxiliary random variable $\bm\varepsilon \in \mathbb R^{d_f}$ is introduced to represent the GP conditional uncertainty in \eqref{eq:ph_xu}. Then, the transition model can be approximated by a first-order Taylor expansion as follows:

\begin{equation}\label{eq:linearized_model}
    \begin{aligned}
       \bm x_{t+1}|\bm x_t, \bar{\bm u}, \bm\varepsilon
        &\approx \mathcal{N} \left(\bm x_{t+1} \,\Big|\, 
        \bm F_t 
        + \bm A_x \left(\bm x_t - \bm m_{x_t} \right)
        + \bm A_{\bar u} \left(\bar{\bm u} - \bm m_{\bar u} \right)
        + \bm A_{\varepsilon}\bm\varepsilon
        ,\, \bm\Sigma_p
        \right)
    \end{aligned}
\end{equation}
where $\bm F_t = \bm F(\bm m_{x_t}, \bm m_{f_t})$, $\bm f_t$ is shorthand for $\bm f(\bm m_{x_t})$, and its mean is $\bm m_{f_t} = \bm\mu_{\mathrm{gp}}(\bm{m}_{x_t}, \bm m_{\bar u})$. Furthermore, the Jacobian matrices in \eqref{eq:linearized_model} can be given by:

\begin{equation}
\begin{aligned}
    \bm A_x &= \frac{\partial \bm{F}(\bm{x}, \bm m_{f_t})}{\partial \bm{x}} \bigg|_{\bm{m}_{x_t}}
    + \bm{A}_f \frac{\partial \bm\mu_{\mathrm{gp}}(\bm x, \bm m_{\bar u})}
    {\partial \bm{x}} \bigg|_{\bm{m}_{x_t}} \\
    \bm A_f &= \frac{\partial \bm{F}(\bm{m}_{x_t}, \bm h)}{\partial \bm h} \bigg|_{m_{f_t}}\\
    \bm A_{\bar u}      & = \bm{A}_f \frac{\partial \bm\mu_{\mathrm{gp}}(\bm m_{x_t}, \bar{\bm u})}
    {\partial \bar{\bm u}} \bigg|_{\bm m_{\bar u}} \\
    \bm A_\varepsilon   &= \bm A_f \bm\Sigma^{1/2}_{\mathrm{gp}}(\bm m_{x_t}) \\
\end{aligned}
\end{equation}
Based on the linearized model \eqref{eq:linearized_model}, the moment matching equations can be derived.
For simplicity in the resulting equations, we denote the joint of the inducing points $\bar{\bm u}$ and the state $\bm x_t$ as an augmented state $\bar{\bm X}_t = [\bar{\bm u}^T, \bm x_t^T]^T$, and its moments as $\bar{\bm\xi}_t = \mathbb{E}_{q(\bar{\bm u}, \bm x_t)}[\bar{\bm X}_t]$ and $\bar{\bm\Sigma}_t = \mathrm{var}_{q(\bar{\bm u}, \bm x_t)}[\bar{\bm X}_t]$. Therefore, the moment matching equations can be written as follows:

\begin{equation}\label{eq:pred_ekf}
    \begin{aligned}
    &\bar{\bm{\xi}}_{t+1}^- = [\bm{m}_{\bar{u}}^T, \bm{F}_t^T]^T \\
    &\bar{\bm{\Sigma}}_{t+1}^- = \bm{\Phi} \bar{\bm{\Sigma}}_t \bm{\Phi}^T 
    + \bm{\Sigma}_{p, \bar{X}} \\
    &\bm{\Phi} = \begin{bmatrix}
        \bm{I}_{n_{\bar u}}  & \bm{0} \\
        \bm{A}_{\bar u}  & \bm{A}_x \\
    \end{bmatrix} \\
    & \bm{\Sigma}_{p, \bar{X}} = \begin{bmatrix}
    \bm{0} & \bm{0} \\
    \bm{0} & \bm{\Sigma}_p + \bm A_f \bm \Sigma_{\mathrm{gp}}(\bm m_{x_t}) \bm A_f^T \\
    \end{bmatrix}
    \end{aligned}
\end{equation}
where $\bm F_t$ has been defined in \eqref{eq:linearized_model}, \( \bar{\bm{\xi}}_{t+1}^- = \mathbb{E}_{q^-(\bar{\bm u}, \bm x_{t+1})}[\bar{\bm X}_{t+1}]\) and \( \bar{\bm{\Sigma}}_{t+1}^- = \mathrm{var}_{q^-(\bar{\bm u}, \bm x_{t+1})}[\bar{\bm X}_{t+1}]\) represent the predicted mean and covariance for the augmented state \( \bar{\bm{X}}_{t+1} = [\bar{\bm u}^T, \bm x_{t+1}^T]^T \) respectively, 
matrix \( \bm{\Phi} \) denotes its transition Jacobian, 
and \( \bm{\Sigma}_{p, \bar{X}} \) is its process noise covariance.

So far, we have derived the EKF-based moment matching equation. It is worth noting that, although the above derivation requires introducing the auxiliary variable $\bm\varepsilon$ (which is not required in \cite{zheng2024recursive}), the resulting equation~\eqref{eq:pred_ekf} covers all possible cases of adding inducing points, namely, adding all, adding some, or adding none. In contrast, the original RGPSSM needs to separately derive the moment matching equations for the cases with and without adding inducing points. 
Therefore, the inference framework proposed in this paper provides a clearer theoretical perspective. In terms of computational efficiency, considering \eqref{eq:pred_ekf} and assuming a low-dimensional system learning condition $M \gg d_x, d_f, d_z$, this method has complexity $\mathcal O(M^2)$.

\subsection{Unscented Kalman Filter (UKF)-based Moment Matching}\label{subsec:ukf}


The UKF is a well-established nonlinear filtering method that uses sigma points for moment matching. Unlike the EKF, which is only accurate to first order, the UKF achieves third-order accuracy (in a Taylor series sense) for Gaussian inputs and arbitrary nonlinearities~\cite{van2001square}. 
Based on the transition model summarized in \eqref{eq:tran_x}, UKF-based moment matching can be performed in a straightforward manner. First, sample sigma points from the joint distribution of $(\bar{\bm u}, \bm x_t, \bm\varepsilon)$. Next, propagate these sigma points through the transition model \eqref{eq:tran_x} to obtain samples for the state $\bm x_{t+1}$. Finally, estimate the moments of the predicted distribution $q^-(\bar{\bm u}, \bm x_{t+1})$ empirically from these propagated samples. The practical implementation is provided in the following pseudocode.

\setlength{\intextsep}{0pt}%
\setlength{\textfloatsep}{0pt}%
\begin{algorithm}[h]
\caption{UKF algorithm for moment propagation}\label{alg:ukf}
\begin{algorithmic}[1] 
\item[] \textbf{Input:} mean $\bm m_{x_t}$ of the state $\bm x_t$, mean $\bm m_{\bar u}$ of the inducing points $\bar{\bm u}$, joint covariance $\bm\Sigma_t = \mathrm{var}[(\bar{\bm u}, \bm x_t)]$.
    \STATE Initialize the hyperparameters of UKF, including $\alpha$ (is usually set to $1e-4 \le \alpha \le 1$), $\beta$ (=2 is optimal for Gaussian distribution), the dimension of sigma-points $d_s = d_x + n_{\bar u} + d_f$. 
    \STATE Calculate the scale parameters $\lambda = L(\alpha^2-1)$, $\eta = \sqrt{d_s + \lambda}$ and the weight  $W_0^{(m)} = \lambda/(d_s+\lambda), W_0^{(c)}=\lambda/(d_s+\lambda) + (1-\alpha^2+\beta)$, $W_i^{(m)}=W_i^{(c)}=1/(2d_s+2\lambda), \, i=1,2,\dots,2d_s$. 
    \STATE Calculate the mean and covariance of the joint variable ($\bar{\bm u},\bm x_t,\bm\varepsilon$):
        \begin{equation*}
        \bm m_{\bar u x_t \varepsilon} = \left[\bm m_{\bar u}^T, \bm m_{x_t}^T, \bm 0^T \right]^T, \quad
        \bm S_{\bar u x_t \varepsilon} = \begin{bmatrix}
            \bm\Sigma_t & \bm 0 \\
            \bm 0 & \bm I_{d_f} \\
        \end{bmatrix}
        \end{equation*}
    \STATE Sample sigma points for the joint variable ($\bar{\bm u},\bm x_t,\bm\varepsilon$):
        \begin{equation*}
        \bm{\mathcal X}_{\bar u x_t \varepsilon} = \begin{bmatrix}
            \bm m_{\bar u x_t \varepsilon} 
            & \bm m_{\bar u x_t \varepsilon} + \eta \bm S_{\bar u x_t \varepsilon}^{1/2}
            & \bm m_{\bar u x_t \varepsilon} - \eta \bm S_{\bar u x_t \varepsilon}^{1/2}
        \end{bmatrix}
        \end{equation*}
    where $\bm S_{\bar u x_t \varepsilon}^{1/2}$ is the Cholesky factor of $\bm S_{\bar u x_t \varepsilon}$.

    \STATE Attain the sigma points of the predicted state $\bm x_{t+1}$ using the transition model \eqref{eq:tran_x} :
        \begin{equation*}
        \bm{\mathcal X}_{x_{t+1}, i}^- = \bm F \left(
            \bm{\mathcal X}_{x_{t}, i}, \bm\mu_{\mathrm{gp}}(\bm{\mathcal X}_{x_{t}, i}, \bm{\mathcal X}_{\bar u, i}) 
            + \bm\Sigma^{1/2}_{\mathrm{gp}}(\bm{\mathcal X}_{x_{t}, i})\bm{\mathcal X}_{\varepsilon, i}
            \right), \quad i=0,1,\dots, 2d_s  
        \end{equation*}
    \STATE Calculate the moments of the predicted joint distribution $q^-(\bar{\bm u}, \bm x_{t+1})$:
        \begin{equation*}
        \begin{aligned}
            \bm m_{x_{t+1}}^- &= \sum\limits_{i=0}^{2 d_s} W_{i}^{(m)} 
            \bm{\mathcal X}_{x_{t+1}, i}^-  \\
            \bm S_{x_{t+1}, x_{t+1}}^- &= \sum\limits_{i=0}^{2 d_s} W_i^{(c)}\left[
                \bm{\mathcal X}_{x_{t+1}, i}^- - \bm m_{x_{t+1}}^-
            \right] \left[
                \bm{\mathcal X}_{x_{t+1}, i}^- - \bm m_{x_{t+1}}^-
            \right]^T + \bm\Sigma_p \\
            \bm S_{\bar u, x_{t+1}}^- &= \sum\limits_{i=0}^{2 d_s} W_i^{(c)}\left[
                \bm{\mathcal X}_{\bar u, i} - \bm m_{\bar u}
            \right] \left[
                \bm{\mathcal X}_{x_{t+1}, i}^- - \bm m_{x_{t+1}}^-\right]^T
        \end{aligned}
        \end{equation*}
        and $\bm m_{\bar u}^- = \bm m_{\bar u}$, $\bm S_{\bar u\bar u}^- = \bm S_{\bar u\bar u}$.
\end{algorithmic}
\end{algorithm}
By using UKF-based moment matching, higher filtering accuracy can be achieved compared to the EKF. However, the computational complexity remains of the same order, namely $\mathcal O(M^2)$, similar to the conclusion of the standard UKF \cite{van2001square}.
Furthermore, it is clear that other sigma-point filters, such as the Cubature Kalman Filter (CKF) \cite{arasaratnam2009cubature} and the Gauss-Hermite Kalman Filter (GHKF) \cite{arasaratnam2007discrete}, can be implemented similarly.

\subsection{Assumed Density Filter (ADF)-based Moment Matching}\label{subsec:adf}

Assumed Density Filter (ADF) is an approximate Bayesian inference method that recursively updates the posterior by projecting it onto a tractable family of distributions, typically via moment matching. 
In the context of GPs, ADF is particularly effective when using the Gaussian kernel (a.k.a. radial basis function (RBF) kernel) because this kernel allows closed-form moment evaluation.
Specifically, the Gaussian kernel is defined as follows.
\begin{equation}\label{eq:kernel_rbf}
\begin{aligned}
    K_{\mathrm{RBF}}(\bm{z}_1, \bm{z}_2; \bm\Lambda) = \sigma^2
    \exp\left(-\frac{1}{2} (\bm{z}_1 - \bm{z}_2)^T \bm{\Lambda}^{-1} (\bm{z}_1 - \bm{z}_2)\right)
\end{aligned}
\end{equation}
where $\sigma^2$ is the signal variance (which captures the signal magnitude), and $\bm{\Lambda} = \mathrm{diag}(l_1^2, l_2^2, \ldots, l_{d_z}^2)$ contains the squared length scales $l_i$ (which capture the input-output dependence).
The Gaussian kernel is widely used in GPs due to its flexibility and has a close relationship with the Gaussian probability density function, which enables analytic computation of the prediction moments when the GP input is uncertain.
 We summarize the analytic property of the Gaussian kernel as follows.

First, a single Gaussian kernel (e.g., used for the $k$-th function dimension) can be written as a scaled Gaussian density:
\begin{equation}\label{eq:property1}
\begin{aligned}
    K_{\mathrm{RBF}}(\bm{z}_1^k, \bm{z}_2^k; \bm \Lambda^k)
    &= \sigma_k^2 (2\pi)^{d_{z^k}/2} \vert \bm\Lambda^k \vert^{1/2} 
    \cdot \mathcal{N}(\bm{z}_1^k \mid \bm{z}_2^k, \bm{\Lambda}^k) \\
    &= c^k \cdot \mathcal{N}(\bm{z}_1^k \mid \bm{z}_2^k, \bm{\Lambda}^k) \\
    &= c^k \cdot \mathcal{N}(\bm{z}_2^k \mid \bm{z}_1^k, \bm{\Lambda}^k)
\end{aligned}
\end{equation}
where $c^k = \sigma_k^2 (2\pi)^{d_{z^k}/2} \vert \bm\Lambda^k \vert^{1/2}$, and $\sigma_k^2$ and $\bm\Lambda^k$ are the kernel hyperparameters for the $k$-th dimension.

Second, the product of two Gaussian kernels (e.g., for the $k$-th and $l$-th function dimensions) can also be written as a scaled Gaussian density over an augmented variable:
\begin{equation}\label{eq:property2}
\begin{aligned}
    K_{\mathrm{RBF}}(\bm z^k, \bm z_i^k; \bm\Lambda^k) 
    \cdot K_{\mathrm{RBF}}(\bm z^l, \bm z_j^l; \bm\Lambda^l)
    &= c^k c^l \cdot \mathcal{N}(\bm z_i^k \mid \bm z^k, \bm \Lambda^k)  \cdot
    \mathcal{N}(\bm z_j^l \mid \bm z^l, \bm \Lambda^l) \\
    &= c^{kl} \cdot  \mathcal{N}
    \left(\bm Z^{kl}_{ij} \mid \bm Z^{kl}, \bm\Lambda^{kl} \right)
\end{aligned}
\end{equation}
where $\bm Z^{kl} = [(\bm z^{k})^T, (\bm z^{l})^T]^T$, $\bm Z^{kl}_{ij} = [(\bm z^{k}_i)^T, (\bm z^{l}_j)^T]^T$, 
$\bm\Lambda^{kl} = \mathrm{blkdiag}(\bm{\Lambda}^k, \bm\Lambda^l)$, and $c^{kl} = \sigma_k^2 \sigma_l^2 (2\pi)^{(d_{z^k} + d_{z^l})/2} \vert \bm\Lambda^{kl} \vert^{1/2}$.
Through the two analytical properties in \eqref{eq:property1} and \eqref{eq:property2}, we can evaluate some of the moments required for the inference task \eqref{eq:pred_task} analytically. In the following, we first decompose the ADF-based moment matching into the function prediction step and the state prediction step, and then derive the exact moment matching formula for the function prediction step.

\textbf{Step decomposition of the ADF-based moment matching.} 
As shown in \eqref{eq:pred_task}, the GP prediction appears only in the function prediction conditional $p(\bm h_t | \bm x_t, \bm u)$, whereas the state prediction conditional $p(\bm x_{t+1} | \bm x_t, \bm h_t)$ can be an arbitrary nonlinear function. 
Therefore, the analytical moment evaluation is applicable only to the function prediction, resulting in the following step decomposition:


\text{(1) Function prediction.} This step involves the function prediction in \eqref{eq:pred_task}, which can be formulated as:
\begin{equation}\label{eq:pred_h}
\begin{aligned}
    q(\bar{\bm u}, \bm h_t, \bm x_t) \approx p(\bm h_t | \bm x_t, \bar{\bm u}) q(\bar{\bm u}, \bm x_t)
\end{aligned}
\end{equation}
Here, $q(\bar{\bm u}, \bm h_t, \bm x_t)$ is a Gaussian approximation, and its moments can be computed exactly when using the Gaussian kernel. 
Since most of the moments of $q(\bar{\bm u}, \bm h_t, \bm x_t)$ can be inherited from $q(\bar{\bm u}, \bm x_t)$, it is necessary to evaluate only the moments related to $\bm h_t$, specifically including the mean $\bm m_{h_t}$ and the cross-covariances $\bm S_{h_t x_t}$ and $\bm S_{h_t \bar u}$.

\text{(2) State prediction.}
By decomposing the function prediction result $q(\bar{\bm u}, \bm h_t, \bm x_t)$ obtained in \eqref{eq:pred_h} as $q(\bm h_t, \bm x_t) q(\bar{\bm u}| \bm h_t, \bm x_t)$, this step can be formulated as:

\begin{equation}\label{eq:u_cond_hx}
\begin{aligned}
    q^-(\bar{\bm u}, \bm x_{t+1}) 
    &= \int p(\bm x_{t+1} | \bm x_t, \bm h_t) q(\bar{\bm u}, \bm h_t, \bm x_t) 
    \mathrm{d}\bm h_t \mathrm{d}\bm x_t \\
    &= \int p(\bm x_{t+1} | \bm x_t, \bm h_t) q(\bm h_t, \bm x_t) q(\bar{\bm u}| \bm h_t, \bm x_t) 
    \mathrm{d}\bm h_t \mathrm{d}\bm x_t \\
    &\approx \int q(\bm x_{t+1}, \bm h_t, \bm x_t) q(\bar{\bm u}| \bm h_t, \bm x_t)
    \mathrm{d}\bm h_t \mathrm{d}\bm x_t
\end{aligned}
\end{equation}
where $q(\bm x_{t+1}, \bm h_t, \bm x_t)$ is a Gaussian approximation of $p(\bm x_{t+1} | \bm x_t, \bm h_t) q(\bm h_t, \bm x_t)$, and $q(\bar{\bm u}| \bm h_t, \bm x_t)$ can be obtained from the joint distribution $q(\bar{\bm u}, \bm h_t, \bm x_t)$. Specifically, using the conditional probability formula for the Gaussian distribution, we have:
\begin{equation}\label{eq:qu_hx}
\begin{aligned}
    q(\bar{\bm u}| \bm h_t, \bm x_t) &=
    \mathcal N \left(
    \bm m_{\bar u} + 
    \begin{bmatrix} \bm{S}_{\bar uh_t} & \bm{S}_{\bar ux_t} \\ \end{bmatrix}
    \begin{bmatrix}
    \bm S_{h_t h_t} & \bm{S}_{h_t x_t} \\
    \bm{S}_{h_t  x_t}^T & \bm{S}_{x_t x_t}
    \end{bmatrix}^{-1}
    \begin{bmatrix} \bm h_t - \bm{m}_{h_t} \\ \bm x_t - \bm{m}_{x_t} \\ \end{bmatrix}, 
    \, *
    \right)
\end{aligned}
\end{equation}
where the variance is not provided because it is not used in the following. 

Therefore, given \eqref{eq:u_cond_hx}, the state prediction consists of two substeps. Firstly, evaluate the joint distribution:

\begin{equation}\label{eq:pred_x}
\begin{aligned}
    q(\bm x_{t+1}, \bm h_t, \bm x_t) \approx p(\bm x_{t+1} | \bm x_t, \bm h_t) q(\bm h_t, \bm x_t)
\end{aligned}
\end{equation}
which is a general moment propagation task and can be handled by the standard UKF, similar to Algorithm \ref{alg:ukf}.


Secondly, compute all the moments of $q^-(\bar{\bm u}, \bm x_{t+1})$. Since most of the moments of $q^-(\bar{\bm u}, \bm x_{t+1})$ can be directly obtained from $q(\bm x_{t+1}, \bm h_t, \bm x_t)$ and $q(\bar{\bm u}, \bm x_t)$, only the cross-covariance $\bm S_{\bar ux_{t+1}}^-$ need to additionally evaluate. 
Considering the last row of \eqref{eq:u_cond_hx} and the conditional distribution \eqref{eq:qu_hx}, this moment can be evaluated by:

\begin{equation}\label{eq:Sux}
\begin{aligned}
    \bm S_{\bar ux_{t+1}}^- &=     
    \begin{bmatrix} \bm{S}_{\bar uh_t} & \bm{S}_{\bar ux_t} \\ \end{bmatrix}
    \begin{bmatrix}
    \bm S_{h_t h_t} & \bm{S}_{h_t x_t} \\
    \bm{S}_{h_t  x_t}^T & \bm{S}_{x_t x_t}
    \end{bmatrix}^{-1}
    \begin{bmatrix}
    \bm{S}_{h_t x_{t+1}}^- \\ \bm{S}_{x_t x_{t+1}}^- \\
    \end{bmatrix}
\end{aligned}
\end{equation}

\textbf{Exact moment matching for the function prediction step.} 
In view of the above step decomposition, the remaining task to obtain the moment evaluation formulas for the function prediction step \eqref{eq:pred_h}.
For brevity, we simplify the notation and rewrite the function prediction conditional \eqref{eq:ph_xu}. Specifically, omit the time index $t$ in the variables $\bm x_t$ and $\bm h_t$, and abbreviate $\bar{\bm u}$ as $\bm u$. 
In addition, we denote 

\begin{equation}
\begin{aligned}
    \bm v^k &= \bm K_{u^k u^k}^{-1} \bm u^k, \quad
    \bm Q^{kl} &=
    \begin{cases}
        \bm K_{u^k u^k}^{-1}, & \text{if } k = l \\
        \bm 0, & \text{if } k \neq l
    \end{cases}
\end{aligned}
\end{equation}
and $\bm v = [\bm v^k]_{k=1}^{d_f}$. Based on these, we rewrite the prediction conditional \eqref{eq:ph_xu} as:

\begin{equation}\label{eq:ph_xu_1}
\begin{aligned}
    &p(\bm h| \bm x, \bm u)
        = \mathcal N\left(\bm h \mid \bm \mu_{\mathrm{gp}}(\bm{z}, \bm{v}), 
        \bm \Sigma_{\mathrm{gp}}(\bm{z})\right) \\
    &\bm \mu_{\mathrm{gp}}(\bm{z}, \bm v) = \left[ \bm K_{h^k u^k} \, \bm v^k \right]_{k=1}^{d_f}\\
    &\bm \Sigma_{\mathrm{gp}}(\bm{z}) = \mathrm{diag}\left( \left[
    \bm{K}_{h^kh^k} - \bm K_{h^k u^k} \, \bm{Q}^{kk} \, \bm K_{u^k h^k} \right]_{k=1}^{d_f} \right)\\
\end{aligned}
\end{equation}
Here, $\bm z = \{\bm z^k\}_{k=1}^{d_f}$ represents the set of GP inputs for the function value $\bm h$ in all dimensions. The input for the $k$-th dimension is obtained by the linear mapping $\bm z^k = \varphi^k(\bm x)$. 
In addition, the elements of the kernel matrix $\bm K_{h^k u^k}$ in \eqref{eq:ph_xu_1} can be given by:

\begin{equation}\label{eq:phi}
\begin{aligned}
    \bm K_{h^k u^k_j} &= K_{\mathrm{RBF}}(\bm z^k, \bm z_j^k; \bm\Lambda^k)
\end{aligned}
\end{equation}
where $\bm z_j^k$ denotes the input of the $j$-th inducing point $\bm u_j^k$. 
Given these expressions and the function prediction task defined in \eqref{eq:pred_h}, the moments that need to be evaluated can be expressed as follows:

\begin{equation}\label{eq:moment}
    \begin{aligned}
        \bm m_h &= \mathbb{E}[\bm \mu_{\mathrm{gp}}(\bm{z}, \bm v)] \\
        \bm S_{hh} &= \mathbb{E}[\bm \Sigma_{\mathrm{gp}}(\bm{z})] + \mathrm{var}[\bm \mu_{\mathrm{gp}}(\bm{z}, \bm v)] \\
                  &= \mathbb{E}[\bm \Sigma_{\mathrm{gp}}(\bm{z})] + \mathbb{E}[\bm \mu_{\mathrm{gp}}(\bm{z}, \bm v)\bm \mu_{\mathrm{gp}}(\bm{z}, \bm v)^T] - \bm m_h \bm m_h^T \\
        \bm S_{hx} &= \mathrm{cov}[\bm \mu_{\mathrm{gp}}(\bm{z}, \bm v), \bm x] \\
        &= \mathbb{E}[\bm \mu_{\mathrm{gp}}(\bm{z}, \bm v)\bm x^T] - \bm m_h \bm m_{x}^T \\
        \bm S_{hu} &= \mathrm{cov}[\bm \mu_{\mathrm{gp}}(\bm{z}, \bm u), \bm u] \\
        &= \mathbb{E}[\bm \mu_{\mathrm{gp}}(\bm{z}, \bm v)\bm u^T] - \bm m_h \bm m_u^T \\
    \end{aligned}
\end{equation}
where the operations $\mathbb{E}$, $\mathrm{var}$, and $\mathrm{cov}$ are taken with respect to the joint distribution $q(\bm{u}, \bm{x}, \bm{v}, \bm{z})$, whose moments can be attained by the following method. 
When the input mapping $\varphi^k(\cdot)$ is linear, the moments of $q(\bm{u}, \bm{x}, \bm{v}, \bm{z})$ can be evaluated from the moments of $q(\bm u, \bm x)$ by utilizing the linear relationship $\bm v^k = \bm K_{u^k u^k}^{-1} \bm u^k$ and $\bm z^k = \varphi^k(\bm x)$. If $\varphi^k(\cdot)$ is nonlinear, it can be achieved via first-order linearization, although this may result in a loss of accuracy.
By combining \eqref{eq:moment}, \eqref{eq:ph_xu_1}, and the two analytic properties \eqref{eq:property1} and \eqref{eq:property2} of the Gaussian kernel, we obtain the evaluation formulas for the desired moments (the derivations are provided in Appendix \ref{app:adf}):

\begin{equation}\label{eq:moment_adf}
\begin{aligned}
    \bm m_{h^k} &= \sum_{j=1}^{n_{u^k}} 
    b_j^{k} \, \bm{m}_{v_j^k|z_j^k} \\
    \bm S_{h^kh^l} &= 
    -\sum_{i=1}^{n_{u^k}} \sum_{j=1}^{n_{u^l}} 
    \bm{Q}_{ij}^{kl} \cdot B_{ij}^{kl} \\
    &\quad + \sum_{i=1}^{n_{u^k}} \sum_{j=1}^{n_{u^l}} 
    B_{ij}^{kl} \left(\bm S_{v_i^k v_j^k|Z_{ij}^{kl}} 
    + \bm m_{v_i^k|Z_{ij}^{kl}} \bm m_{v_j^k|Z_{ij}^{kl}}^T \right) \\
    &\quad + \bm K_{h^kh^k} - \bm m_{h^k} \bm m_{h^k}^T \\
    \bm S_{h^kx} &= \sum_{j=1}^{n_{u^k}} b_j^{k} \left( \bm{S}_{v_j ^k x|z_j^k} 
    + \bm{m}_{v_j^k|z_j^k} \bm{m}_{x|z_j^k}^{T} \right) 
    - \bm m_{h^k} \bm m_{x}^T \\
    \bm S_{h^ku} &= \sum_{j=1}^{n_{u^k}} b_j^{k} \left( \bm{S}_{v_j^k u|z_j^k} + \bm{m}_{v_j^k|z_j^k} \bm{m}_{u|z_j^k}^T \right)- \bm m_{h^k} \bm m_{u}^T \\
\end{aligned}
\end{equation}
where $\bm h^k$ denotes the $k$-th element of the vector $\bm h$, so $\bm m_{h^k}$ is the $k$-th element of the mean vector $\bm m_h$. Similarly, $\bm S_{h^k h^l}$ is the $(k,l)$-th entry of the covariance matrix $\bm S_{hh}$, and $\bm S_{h^k x}$ and $\bm S_{h^k u}$ are the $k$-th rows of the matrices $\bm S_{hx}$ and $\bm S_{hu}$, respectively.
In addition, the variable $\bm Z_{ij}^{kl} = [(\bm z_i^{k})^T, (\bm z_j^{l})^T]^T$ as defined in \eqref{eq:property2}. 
The constants $b_j^k$ and $B_{ij}^{kl}$ are defined as:

\begin{equation}
\begin{aligned}
b_j^k &= \sigma^2_k \cdot 
\left| \bm I + \bm S_{z^kz^k} (\bm\Lambda^{k})^{-1} \right|^{-1/2} \cdot
\exp\left[-\frac{1}{2} \left(\bm{z}_j^k - \bm{m}_{z^k}\right)^T 
\left(\bm{\Lambda}^k + \bm S_{z^kz^k}\right)^{-1} 
\left(\bm{z}_j^k - \bm{m}_{z^k}\right)\right] \\
B_{ij}^{kl} &= \sigma_k^2 \sigma_l^2 \cdot \vert \bm I + \bm S_{Z^{kl}Z^{kl}} (\bm\Lambda^{kl})^{-1} \vert^{-1/2} \cdot
\exp\left[-\frac{1}{2} \left(\bm{Z}_{ij}^{kl} - \bm{m}_{Z^{kl}} \right)^T 
\left(\bm{\Lambda}^{kl} + \bm S_{Z^{kl}Z^{kl}} \right)^{-1} 
\left(\bm{Z}_{ij}^{kl} - \bm{m}_{Z^{kl}} \right)\right]
\end{aligned}
\end{equation}
where the mean $\bm{m}_{z^k}$ and $\bm{m}_{Z^{kl}}$, and the covariance $\bm S_{z^kz^k}$ and $\bm S_{Z^{kl}Z^{kl}}$ can be attained from the moments of $q(\bm{u}, \bm{x}, \bm{v}, \bm{z})$. In addition, the conditional moments in \eqref{eq:moment_adf}, such as $\bm{m}_{v_j^k|z_j^k}$ and $\bm S_{v_i^k v_j^k|Z_{ij}^{kl}}$, can be computed using Kalman filter-like equations:

\begin{equation}\label{eq:moment_kf12}
    \begin{aligned}
        \bm{m}_{a|z_j^k} &= 
        \bm m_a + \bm{S}_{az^k} (\bm{S}_{z^kz^k} + \bm{\Lambda}^k)^{-1} (\bm z_j^k - \bm m_{z^k}) \\
        \bm{S}_{ab|z_j^k} &= 
        \bm{S}_{ab} - \bm{S}_{az^k} (\bm{S}_{z^kz^k} + \bm{\Lambda}^k)^{-1} \bm{S}_{z^kb} \\
        \bm m_{a|Z_{ij}^{kl}} &= \bm m_{a} + \bm S_{aZ^{kl}} (\bm S_{Z^{kl}Z^{kl}} + \bm{\Lambda}^{kl})^{-1} (\bm Z_{ij}^{kl} - \bm{m}_{Z^{kl}}) \\
        \bm S_{a b|Z_{ij}^{kl}} 
        &= \bm S_{a b} - \bm S_{a Z^{kl}} 
        (\bm S_{Z^{kl}Z^{kl}} + \bm{\Lambda}^{kl})^{-1} \bm S_{Z^{kl}b} \\
\end{aligned}
\end{equation}
where $\bm a$ and $\bm b$ can be replaced by $\bm x$, $\bm z$, $\bm u$, or $\bm v$ to obtain the corresponding moments.

So far, we have derived the ADF-based moment matching method. In theory, when the state prediction conditional $p(\bm x_{t+1}|\bm x_t, \bm h_t)$ is linear, it can provide exact moment matching for the prediction step. In terms of computational efficiency, considering \eqref{eq:moment_adf} and assuming a low-dimensional system learning condition $M \gg d_x, d_f, d_z$, this method has complexity $\mathcal O(M^2)$, which is of the same order as the EKF-based and UKF-based methods.

In summary, this section applies three moment matching techniques for the prediction step of the RGPSSM. First, the EKF-based method is derived using the new inference framework presented in Section \ref{subsec:inference}, resulting in a simpler equation compared to that of the original RGPSSM method. Second, based on the transition model \eqref{eq:tran_x}, the UKF is employed to achieve more accurate moment matching for general nonlinearity. Third, the analytic properties of the Gaussian kernel are exploited to achieve exact moment matching. The computational complexity, learning accuracy and novelty of the proposed methods are summarized as follows.

\begin{itemize}
    \item \textbf{Computational Efficiency:} All three methods have computational complexity $\mathcal O(M^2)$.

    \item \textbf{Learning accuracy:} For a linear state prediction conditional $p(\bm x_{t+1}|\bm x_t, \bm h_t)$, the accuracies are ordered as
    \begin{center}
        EKF $<$ UKF $<$ ADF.
    \end{center}

    \item \textbf{Novelty:} The method proposed in this paper has the following key differences compared with existing works applying Bayesian filters in GPSSMs, such as \cite{Ko2009} (using UKF) and \cite{deisenroth2009analytic} (using ADF):
    \begin{enumerate}
        \item \textbf{Joint inference:} In this work, both the GP and the state are estimated simultaneously. In contrast, in \cite{Ko2009} and \cite{deisenroth2009analytic}, although the system model is represented as a GP, it is not updated during the state filtering process. 
        Therefore, our method uses a dynamic set of inducing points, and requires additional evaluation of the GP-relevant moments $\bm m_u$, $\bm S_{ux}$, and $\bm S_{uu}$ to achieving GP learning.

        \item \textbf{State-GP correlation:} The correlation between the state and the GP is not considered in either \cite{Ko2009} or \cite{deisenroth2009analytic}. This may potentially cause underestimated uncertainty, as discussed in \cite{hewing2020simulation}. Our method explicitly considers this correlation by incorporating the covariance between the state and the GP inducing points, i.e., $\bm S_{xu}$. 
        Through this covariance, our method achieves more accurate uncertainty propagation and can learn the GP from the measurements $\bm y$.

        \item \textbf{Auxiliary variable $\bm \varepsilon$:} This paper introduces an auxiliary random variable $\bm\varepsilon$ to account for the GP prior conditional prediction uncertainty (variance $\bm \Sigma_{\mathrm{gp}}(\bm{x}_t)$ in \eqref{eq:ph_xu}), which offers a more theoretically sound basis than that in \cite{Ko2009}. Specifically, \cite{Ko2009} omits GP uncertainty during sigma-point propagation of the UKF, and instead accounts for it by simply adding $\bm \Sigma_{\mathrm{gp}}(\bm m_{x_t})$ to the predicted state variance $\bm S_{x_{t+1}x_{t+1}}^-$. These two approaches are not equivalent, and our method is more theoretically appropriate.

    \end{enumerate}
\end{itemize}

Therefore, this paper further develops the moment matching techniques in GPSSMs, particularly in online learning scenarios.

\subsection{Algorithm Summary}

In summary, this paper extends the original RGPSSM method proposed in \cite{zheng2024recursive} to support heterogeneous multi-output latent function learning and the use of general moment matching techniques, thereby improving both flexibility and accuracy. We refer to the extended algorithm as RGPSSM-H, where "H" denotes its heterogeneous learning capability.
The modifications in the extended method do not change the overall algorithmic flow, as illustrated in Fig.~\ref{fig:alg_flow}. They only affect the process of inducing-point management and moment matching.
We summarize the RGPSSM-H method in Algorithm \ref{alg:rgpssm_h}.

\begin{algorithm}[h]
    \caption{Recursive Gaussian Process State Space Model for Heterogeneous Multi-Output Latent Function Learning with EKF-, UKF-, and ADF-Based Moment Matching}\label{alg:rgpssm_h}
    \begin{algorithmic}[1] 
    \item[] \textbf{Input:} Initial inducing points $\bm{u}$, threshold for adding points $\varepsilon_{tol}$, and budget for the inducing-point set $M$. 
    \REPEAT
        \FOR{$k = 1$ \TO $d_f$}
            \STATE Assess the novelty $\gamma^k$ for the $k$-th function dimension using \eqref{eq:gam_k}.
            \IF{$\gamma^k / \mathrm{max}\left[ \mathrm{diag}(\bm K_{u^k u^k}) \right] > \varepsilon_{tol}$}
                \STATE Add the new point $f_t^k$ to the inducing-point set $\bm u$.
            \ENDIF
        \ENDFOR
        \STATE Perform moment matching for the prediction step using the EKF-, UKF-, or ADF-based method as described in Section \ref{sec:ukf_adf}.

        \IF{the number of inducing points $n_u > M$}
            \STATE Identify the least important $n_u - M$ points based on the score in \eqref{eq:score} and remove it by deleting corresponding dimensions for the mean and covariance of $q(\bm u, \bm x)$.
        \ENDIF
        \IF{a new measurement $\bm y_t$ is available}
            \STATE Correct the moments by using EKF- or UKF-based moment matching based on \eqref{eq:corr_step}.
        \ENDIF  
        \STATE Update GP hyperparameters using the method detailed in \cite{zheng2024recursive}.
        \STATE Prune redundant inducing points if necessary according to the criterion shown in \eqref{eq:singularity_test}.
    \UNTIL{termination}
\end{algorithmic}
\end{algorithm}

In addition, to maintain the numerical stability, the original RGPSSM method provides a stable implementation method by performing Cholesky factorization for the joint covariance matrix $\bm\Sigma_t = \mathrm{var}[(\bm u, \bm x_t)]$. To adapt to the modified inducing-point management and new moment matching techniques, we develop the corresponding Cholesky version for the new algorithm, which is described in Appendix \ref{subsec:chol}. According to the derivation in Appendix \ref{subsec:chol}, the Cholesky version for EKF-based, UKF-based, and ADF-based moment matching in the prediction step all maintain quadratic complexity $\mathcal O(M^2)$. 
The computational complexity of each procedure in the Cholesky version of RGPSSM and RGPSSM-H is summarized in Table \ref{tab:complexity}.
Here, the maximum size of the inducing points $M$ follows the definition in this paper, that is, the total number of points across all dimensions, which is different from the definition in \cite{zheng2024recursive}. In addition, the complexity analysis assumes the low-dimensional system learning condition where $M \gg d_x, d_f, d_z$, and therefore only the highest order of complexity with respect to $M$ is retained.

\begin{table}[h]
    \centering
    \caption{Computational complexity of the Cholesky version of the original RGPSSM and RGPSSM-H}
    \begin{tabular}{lcc}
    \hline
    Algorithm step & Cholesky RGPSSM & Cholesky RGPSSM-H \\
    \hline
    GP prediction & $\mathcal{O}\left( M^2 \right)$ & $\mathcal{O}\left( M^2 \right)$ \\
    Moment propagation & $\mathcal{O}\left( M^3 \right)$ & $\mathcal{O}\left( M^2 \right)$ \\
    Inducing point discarding & $\mathcal{O}\left( M^3 \right)$ & $\mathcal{O}\left( M^3 \right)$ \\
    Moment correction & $\mathcal{O}\left( M^2 \right)$ & $\mathcal{O}\left( M^2 \right)$ \\
    Hyperparameter update & $\mathcal{O}\left( M^3 \right)$ & $\mathcal{O}\left( M^3 \right)$ \\
    \hline
    Total complexity & $\mathcal{O}\left( M^3 \right)$ & $\mathcal{O}\left( M^3 \right)$ \\
    \hline
    \end{tabular}
    \label{tab:complexity}
\end{table}

As shown in Table \ref{tab:complexity}, the total complexity of RGPSSM-H is of the same order as that of RGPSSM, while its flexibility and accuracy are further improved. 
In addition, the complexity of moment propagation for RGPSSM-H is $\mathcal O(M^2)$, which is reduced from the RGPSSM complexity of $\mathcal O(M^3)$ because of a new modification to the algorithm, as discussed in Appendix \ref{subsec:chol}. Specifically, the order of $\bm x$ and $\bm u$ in the covariance matrix $\bm\Sigma$ is exchanged. This results in the upper left block of the Cholesky factor of $\bm\Sigma$ remaining unchanged in the prediction step and not requiring recomputation, which reduces the computational cost.

\subsection{Related Works}


\textbf{Multi-output Gaussian processes (MOGPs).}
Extensive research has been conducted on MOGPs (see reviews \cite{liu2018remarks, alvarez2012kernels}). Existing studies mainly focus on three aspects.
The first is capturing the correlations among different function outputs to improve learning accuracy \cite{alvarez2012kernels, mcdonald2023nonparametric, alvarez2011computationally}.
The second is addressing situations where different outputs depend on different input spaces \cite{liu2023learning, hebbal2021multi}.
The third is improving computational efficiency for high-dimensional outputs \cite{lin2025efficient, zhe2019scalable, higdon2008computer}.
Among these, the first two aspects are most relevant to this paper.

For the first aspect, the main research direction is designing kernels that can model dependencies among outputs. Representative approaches include the linear model of coregionalization (LMC) \cite{alvarez2012kernels}, the process convolution (CONV) model \cite{alvarez2011computationally, mcdonald2023nonparametric}, and the spectral mixture (SM) kernel \cite{altamirano2022nonstationary, chen2019multioutput}.
The LMC framework represents a multi-output function as a linear combination of latent functions, resulting in a kernel of the form
$\bm{K}_0(\bm z_1, \bm z_2) = \sum\limits_{i=1}^Q {K}_{0,i}(\bm z_1, \bm z_2) \bm{\Xi}_i$
where ${K}_{0,i}(\cdot)$ is a scalar-valued kernel and $\bm{\Xi}_i$ is the coregionalization matrix.
LMC is computationally convenient and is used in the original RGPSSM.
In contrast, CONV models each output as a convolution of latent functions, while SM can be viewed as a special case of CONV that defines the kernel via the inverse Fourier transform of a matrix-valued spectral density function.
Compared to LMC, both CONV and SM are more expressive, as they can learn richer inter-output dependencies through the convolution kernel \cite{mcdonald2023nonparametric} or spectral density function \cite{chen2019multioutput}. However, they introduce more hyperparameters and are generally harder to train, making them more suitable for offline applications.

For the second aspect, which concerns modeling outputs with heterogeneous input spaces, existing research is relatively limited. Most approaches attempt to learn a mapping that projects different input spaces into a shared latent representation. For example, \cite{liu2023learning} and \cite{hebbal2021multi} employ variational GP to align the inputs of different outputs. 
It should be noted that, although the CONV model theoretically supports heterogeneous inputs through the convolution kernel, a learning-based input alignment is still necessary for effective modeling. 
Another related work is the multiview GP \cite{mao2020multiview}, which addresses scenarios where multiple input views correspond to a single output. This method combines information from different views by introducing an auxiliary GP whose input is the concatenation of all view-specific inputs.

Compared to the aforementioned methods, the heterogeneous multi-output kernel \eqref{eq:S0} proposed in this paper is mainly designed for online learning applications. 
In online scenarios, the amount of available data is often limited. As a result, explicitly modeling correlations among different outputs can easily lead to overfitting and instability during learning. To address this issue, we design an independent multi-output kernel, which simplifies the handling of heterogeneous input spaces, kernel types, and hyperparameters.
This independence also leads to a block diagonal structured kernel matrix as in \eqref{eq:S0}, which improves computational efficiency, an essential requirement for online applications.
The improvement in efficiency is mainly due to the ability to independently add inducing points for each output dimension, as described in \eqref{eq:add_criterion}, and to perform independent GP predictions during inference. As a result, the computational complexity with respect to the output dimensions for these operations is reduced from cubic to linear.
Overall, the proposed heterogeneous multi-output kernel provides high flexibility for modeling heterogeneous outputs, while maintaining superior computational efficiency and stability for online learning scenarios.

\textbf{Uncertainty propagation (UP) for Gaussian process dynamical systems (GPDS).}
Propagating uncertainty in GPDSs is a fundamental task that underpins many learning algorithms and downstream applications such as state estimation \cite{Ko2009,deisenroth2009analytic,wolff2025gaussian}, model predictive control \cite{pan2017prediction,liu2025learning}, and reinforcement learning \cite{deisenroth2011pilco}.
The most common strategy is to recursively propagate the state distribution while treating the GP prediction uncertainty as independent process noise. 
This approach is computationally efficient and generally accurate for short-term prediction.
For example, \cite{Ko2009} applies Bayesian filtering method including EKF, UKF, and PF to implement UP in GPDSs. 
In addition, \cite{deisenroth2009analytic} introduces ADF to achieve exact moment matching in GPDSs for kernels including the Gaussian, trigonometric, and polynomial kernels.
However, this category of methods has a theoretical limitation. The predicted state, such as $x_{t+1}$, is generated by the GP model and is therefore statistically correlated with the GP itself. In principle, this correlation should be taken into account in subsequent state predictions.
However, these approaches ignore this correlation, which can result in underestimated uncertainty as discussed in \cite{hewing2020simulation}.

To address this issue, two main solutions have been proposed: sample-based methods and linearization-based methods.
The sample-based method \cite{frigolaBayesianTimeSeries2015} propagates uncertainty by sequentially generating state trajectories and treating all past states as GP training data for each prediction. Although this approach is theoretically exact, its computational cost grows cubically with the prediction horizon. 
To reduce the computational burden, \cite{beckers2021prediction} proposes conditioning only on a window of recent states. 
This approximation improves computational efficiency and only produces overestimated uncertainty, which is more desirable for ensuring control safety.
In contrast, linearization-based methods \cite{ridderbusch2022approximate, hewing2020simulation} approximate the GP locally by linearizing it around the mean of the state distribution. In this way, the state $x_t$ can be expressed as a linear function of the GP evaluated at all previous state means, i.e., $\bm m_{x_{0}},\dots,\bm m_{x_{t-1}}$, allowing analytical computation of the mean and covariance.

The prediction step in the RGPSSM also involves uncertainty propagation. To handle this, we provide a third and more efficient solution. Instead of relying on the linearization-based approach that depends on function values at all previous state means, we introduce a set of inducing points to capture the key correlations. This design significantly reduces computational cost while preserving high accuracy.
Moreover, the proposed framework supports multiple moment-matching schemes, including EKF-based (linearization), UKF-based, and ADF-based formulations, which further enhance estimation accuracy. Consequently, the proposed method achieves a favorable balance between computational efficiency and predictive performance.

\section{Experiment}\label{sec:experiment}

This section first graphically demonstrates the capability of the proposed method through two simple illustrative experiments, which are presented in Section \ref{subsec:kink} and Section \ref{subsec:param_iden}. 
Next, we conduct two experiments to demonstrate the performance of the algorithm in a complex task and a real-world application, namely learning the flight dynamics of a hypersonic vehicle in Section \ref{subsec:hypersonic} and learning the real-world quadrotor dynamics in Section \ref{subsec:exp_quadrotor}.
The learning algorithm is implemented on a desktop computer running Python 3.9, with an Intel(R) Core(TM) i7-14700F 2.10 GHz processor and 32 GB of RAM. The associated code is available online\footnote{\href{https://github.com/TengjieZheng/rgpssm}{https://github.com/TengjieZheng/rgpssm}}. 
In all experiments, unless otherwise specified, the heterogeneous multi-output kernel \eqref{eq:S0} with the Gaussian kernel \eqref{eq:kernel_rbf} as its single-output kernel is used for the RGPSSM-H.
In addition, when considering a continuous-time SSM with the transition model $\dot{\bm x} = \bm F(\bm x, \bm f(\bm x))$, we transform it into a discrete-time SSM in the form $\bm x_{t+1} = \bm x_t + \bm F(\bm x_t, \bm f(\bm x_t)) \Delta t$ to facilitate the learning process.

\subsection{Kink Transition Function Learning}\label{subsec:kink}

\begin{figure}[H]
    \centering
    \begin{subfigure}[b]{0.48\textwidth}
        \centering
        \includegraphics[width=\textwidth]{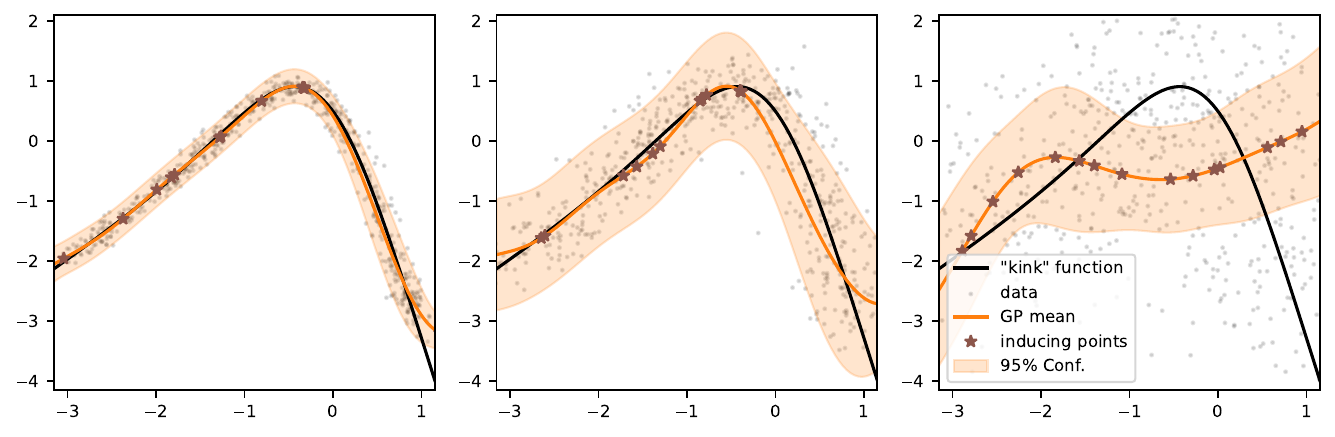}
        \caption{offline: VCDT \cite{Ialongo2019}}
        \label{fig:subfig1}
    \end{subfigure}
    \hspace{0.01\textwidth}
    \begin{subfigure}[b]{0.48\textwidth}
        \centering
        \includegraphics[width=\textwidth]{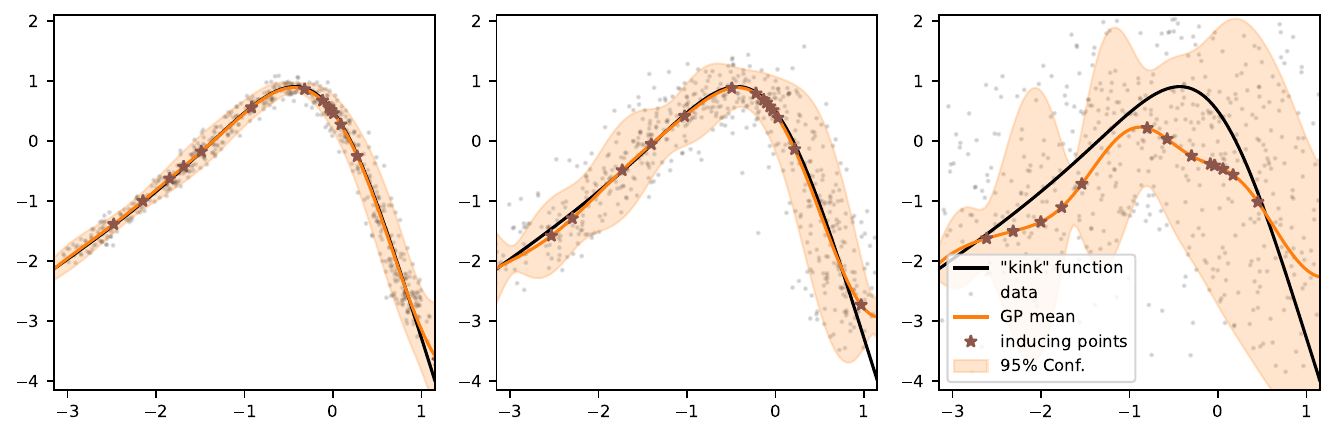}
        \caption{offline: EnVI \cite{Lin2024}}
        \label{fig:subfig2}
    \end{subfigure}
    \vspace{0.5em}
    \begin{subfigure}[b]{0.48\textwidth}
        \centering
        \includegraphics[width=\textwidth]{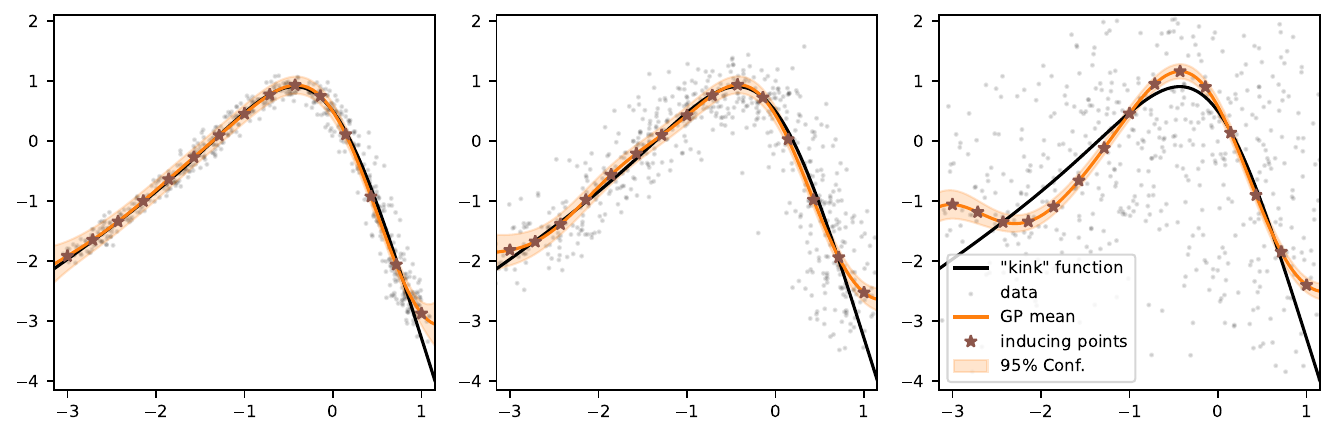}
        \caption{online: SVMC-GP \cite{zhaoStreamingVariationalMonte2023}}
        \label{fig:subfig3}
    \end{subfigure}
    \hspace{0.01\textwidth}
    \begin{subfigure}[b]{0.48\textwidth}
        \centering
        \includegraphics[width=\textwidth]{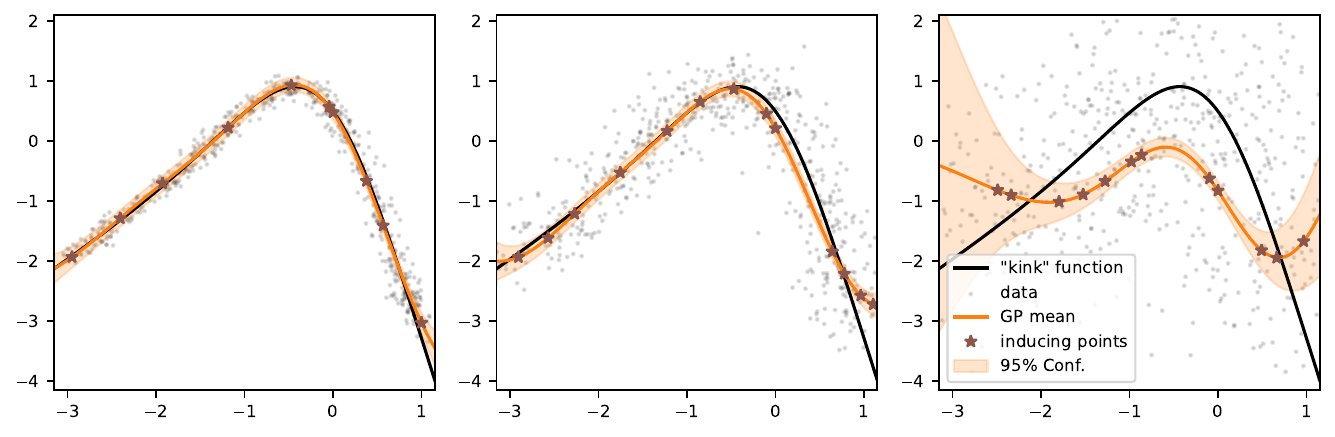}
        \caption{online: RGPSSM-H (EKF)}
        \label{fig:subfig4}
    \end{subfigure}
    \vspace{0.5em}
    \begin{subfigure}[b]{0.48\textwidth}
        \centering
        \includegraphics[width=\textwidth]{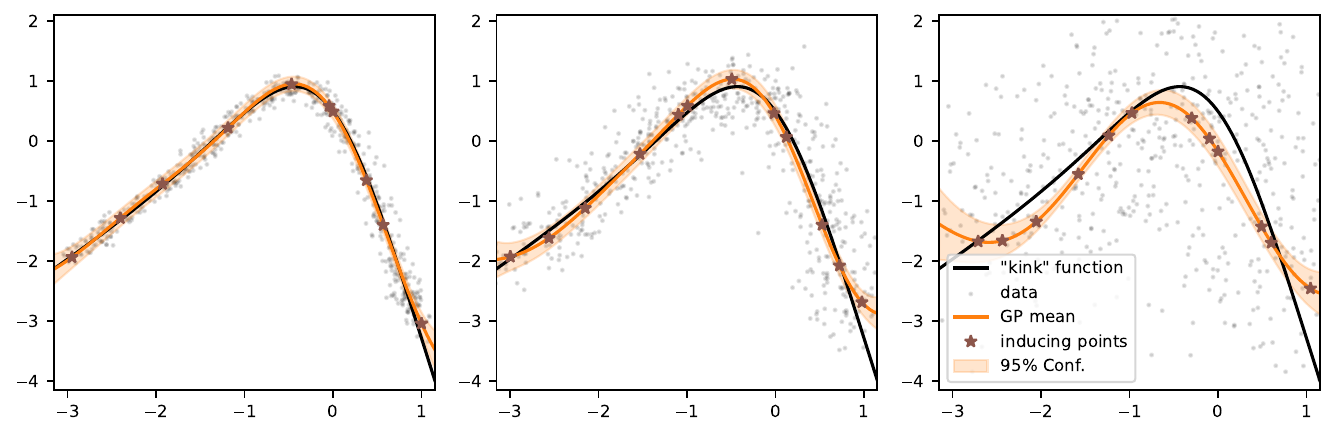}
        \caption{online: RGPSSM-H (UKF)}
        \label{fig:subfig5}
    \end{subfigure}
    \hspace{0.01\textwidth}
    \begin{subfigure}[b]{0.48\textwidth}
        \centering
        \includegraphics[width=\textwidth]{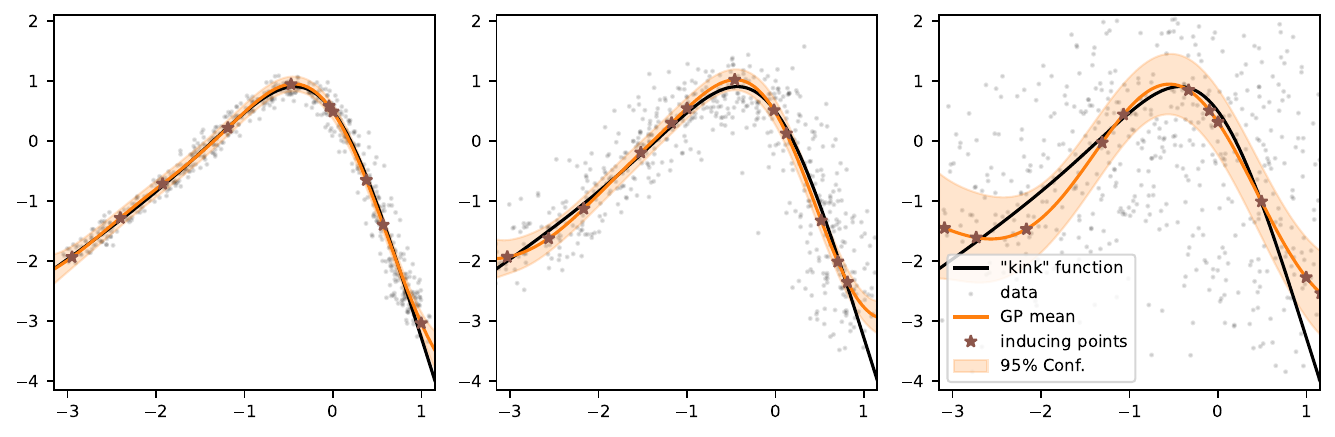}
        \caption{online: RGPSSM-H (ADF)}
        \label{fig:subfig6}
    \end{subfigure}
    \caption{Best learning performance of the kink transition function over five random experiments using various methods across different levels of measurement noise ($\sigma^2_m \in \{0.008,\, 0.08,\, 0.8\}$, from left to right).}
    \label{fig:kink}
\end{figure}

This section mainly evaluates the accuracy and efficiency of RGPSSM-H with different moment matching techniques. To benchmark the performance, we use two state-of-the-art (SOTA) offline GPSSM methods: VCDT \cite{Ialongo2019} and EnVI \cite{Lin2024}, as well as a SOTA online GPSSM method, SVMC-GP \cite{zhaoStreamingVariationalMonte2023}, for comparison.
This experiment aims to learn a one-dimensional highly nonlinear function known as the kink function, which is a widely used benchmark for evaluating GPSSM methods \cite{Ialongo2019, Lin2024}. The system model is given by:

\begin{equation}
\begin{aligned}
    x_{t+1} &= 0.8 + (x_t + 0.2)
    \left[1 - \dfrac{5}{1+\exp(-2x_t)} \right] + \omega_p, \quad 
    &\omega_p \sim \mathcal N(0, \sigma_p^2) \\
    y_t &= x_t + \omega_m, \quad 
    &\omega_m \sim \mathcal N(0,\sigma_m^2) \\
\end{aligned}
\end{equation}
The process noise variance is set to $\sigma_p^2=0.05$, and we simulate three measurement noise variances $\sigma^2_m \in \{0.008,\, 0.08,\, 0.8\}$, where high noise levels are challenging for EKF-based moment matching.
All methods are trained on a measurement sequence of 600 data points, and the maximum size of the inducing points is set to 15.
Regarding the learning process, the offline methods can repeatedly utilize the entire batch of data, whereas the online method faces a more challenging setting in which data are processed sequentially and it is not allowed to reuse past data.
The final GP learning results are evaluated using normalized mean squared error (nMSE), which measures the accuracy of mean prediction, and mean negative log-likelihood (MNLL), which measures the accuracy of uncertainty quantification. 
Table~\ref{tab:noise_comparison} presents the prediction accuracy and total training time for five random runs, and Fig.~\ref{fig:kink} shows the best GP prediction case.

\begin{table}[htbp]
    \centering
    {\footnotesize
    \resizebox{\columnwidth}{!}{%
    \begin{tabular}{llcccc}
        \toprule
        \multicolumn{2}{c}{Method} & $\sigma^2_m = 0.008$ (nMSE $\downarrow$ $\mid$ MNLL $\downarrow$) & $\sigma^2_m = 0.08$ (nMSE $\downarrow$ $\mid$ MNLL $\downarrow$) & $\sigma^2_m = 0.8$ (nMSE $\downarrow$ $\mid$ MNLL $\downarrow$) & Time (s) $\downarrow$ \\
        \cmidrule(lr){1-2} \cmidrule(lr){3-6}
        \multirow{2}{*}{\textbf{Offline}} & VCDT \cite{Ialongo2019} & 0.0199$\pm$0.0038 $\mid$ -0.4802$\pm$0.0569 & 0.6479$\pm$0.7185 $\mid$ 1.0672$\pm$0.6545 & 1.8253$\pm$0.4850 $\mid$ 2.5484$\pm$0.5455 & 5587.10$\pm$914.80 \\
                                           & EnVI \cite{Lin2024} & \cellcolor[HTML]{9E86B0}0.0035$\pm$0.0005 $\mid$ -1.0763$\pm$0.0219 & \cellcolor[HTML]{9E86B0}0.0332$\pm$0.0095 $\mid$ 0.0641$\pm$0.0845 & \cellcolor[HTML]{CABCD4}0.3204$\pm$0.0587 $\mid$ 1.1839$\pm$0.3600 & 292.22$\pm$4.21 \\
        \midrule
        \multirow{4}{*}{\textbf{Online}} & SVMC-GP \cite{zhaoStreamingVariationalMonte2023} & 0.0198$\pm$0.0027 $\mid$ -0.5137$\pm$0.1734 & 0.0651$\pm$0.0187 $\mid$ 5.6546$\pm$2.2685 & 0.7811$\pm$0.7582 $\mid$ 93.4310$\pm$71.1789 & 84.20$\pm$34.86 \\
                                         & RGPSSM-H (EKF) & 0.0075$\pm$0.0011 $\mid$ -1.1780$\pm$0.0978 & 0.0579$\pm$0.0095 $\mid$ 4.8183$\pm$1.6441 & 0.8441$\pm$0.1476 $\mid$ 43.7460$\pm$5.7688 & \cellcolor[HTML]{9E86B0}2.16$\pm$0.14 \\
                                         & RGPSSM-H (UKF) & \cellcolor[HTML]{EDE8F0}0.0068$\pm$0.0011 $\mid$ -1.2770$\pm$0.0748 & \cellcolor[HTML]{EDE8F0}0.0402$\pm$0.0078 $\mid$ 1.3220$\pm$0.7826 & \cellcolor[HTML]{EDE8F0}0.3767$\pm$0.1692 $\mid$ 18.2753$\pm$10.2788 & \cellcolor[HTML]{CABCD4}2.45$\pm$0.15 \\
                                         & RGPSSM-H (ADF) & \cellcolor[HTML]{CABCD4}0.0066$\pm$0.0010 $\mid$ -1.2763$\pm$0.0763 & \cellcolor[HTML]{CABCD4}0.0365$\pm$0.0078 $\mid$ 0.7455$\pm$0.6168 & \cellcolor[HTML]{9E86B0}0.2236$\pm$0.0747 $\mid$ 1.1824$\pm$0.4585 & \cellcolor[HTML]{EDE8F0}3.22$\pm$0.15 \\
        \bottomrule
    \end{tabular}%
    }
    }
    \caption{Kink transition function learning performance using various methods across different levels of measurement noise ($\sigma^2_m \in \{0.008,\, 0.08,\, 0.8\}$, from left to right), reported as mean $\pm$ standard deviation. The top three results are highlighted in shades of purple, with darker shades indicating better performance.}
    \label{tab:noise_comparison}
\end{table}

\textbf{RGPSSM-H with different moment matching methods.} As shown in Table \ref{tab:noise_comparison}, in terms of accuracy, the ADF variant performs the best, followed by UKF, while EKF performs the worst. In particular, at the highest noise level $\sigma_m^2=0.8$, the learning performance of EKF degrades significantly, whereas RGPSSM-H (UKF) and RGPSSM-H (ADF) still maintain good performance. 
In terms of computation time, EKF is the fastest, followed closely by UKF, while ADF requires approximately 1.5 times the computation time of EKF.
These results demonstrate that the integration of UKF and ADF improve learning accuracy under highly nonlinear and high-noise conditions while maintaining similar computational efficiency.

\textbf{RGPSSM-H vs SOTA offline GPSSM methods.} In terms of accuracy, the best RGPSSM-H variant, namely RGPSSM-H (ADF), achieves performance comparable to the SOTA offline method EnVI. 
Notably, at the highest noise level ($\sigma_m^2=0.8$), RGPSSM-H (ADF) achieves lower nMSE and MNLL than EnVI.
The reason why EnVI performs poorly under high-noise conditions may be that it approximates the smoothing distribution with the filtering distribution, which is accurate only when the measurement noise is small.
In contrast, our method leverages the covariance between the state and the GP, which enables more effective aggregation of measurement information across multiple time steps, thereby improving learning accuracy in the presence of large noise.
Regarding computation time, thanks to its recursive learning paradigm, the RGPSSM-H method is two orders of magnitude faster than the best offline method EnVI.
These results demonstrate that by incorporating high-precision moment matching techniques, RGPSSM-H can achieve competitive accuracy while maintaining superior computational efficiency.

\textbf{RGPSSM-H vs SOTA online GPSSM methods.} 
Compared with the SOTA online method SVMC-GP, RGPSSM-H achieves better accuracy in all cases except at the highest noise level $\sigma_m^2=0.8$, where SVMC-GP outperforms the RGPSSM-H (EKF).
In terms of computation time, RGPSSM-H requires less than $1/20$ of the runtime of SVMC-GP.
The reason for the inferior performance of SVMC-GP is that, although SVMC-GP is based on particle filtering and can theoretically handle arbitrarily complex nonlinearities and noise levels, it requires numerous particles and often suffers from particle degeneracy, which leads to degraded performance. Therefore, RGPSSM-H outperforms SVMC-GP in both learning accuracy and computational efficiency in this experiment.

\subsection{Time-varying Parameter Online Identification}\label{subsec:param_iden}

This subsection demonstrates the heterogeneous learning capacity of the proposed method. The experiment is to online identify time-varying model parameters, which possess different time varying rates. Specifically, the system model is given by:

\begin{equation}
\begin{aligned}
    \dot x &= \theta_1 x + \theta_2 +  c \\
    y &= x + \omega_m \\
\end{aligned}
\end{equation}
where $x \in \mathbb{R}$ is the state, $c \in \mathbb{R}$ denotes the control input, and $y \in \mathbb{R}$ is the measurement with a noise standard deviation of $\sigma_{m} = 0.05$.
Additionally, $\theta_1 = \cos(t)$ and $\theta_2= \cos(0.2t)$ are model parameters that vary over time at different rates. 
The RGPSSM and RGPSSM-H are implemented to identify the time-varying parameter $\bm\theta(t) = [\theta_1(t), \theta_2(t)]^T$, which is modeled as latent functions $\bm f(\cdot)$ and assigned a GP prior, namely, $\bm \theta(t) \sim \mathcal{GP}(\bm 0, \bm K_0(t, t^\prime))$. Three learning settings are simulated in this experiment:

\textbf{RGPSSM:} As the baseline, we first simulate the RGPSSM, which uses an isomorphic Gaussian kernel with a shared length scale $l$ for the two function dimensions.

\textbf{RGPSSM-H:} To demonstrate the advantages of allowing different kernel hyperparameters and independently selecting inducing points, we simulate the RGPSSM-H, which uses a heterogeneous Gaussian kernel with independent length scales $l^1$ and $l^2$ for the two function dimensions.

\textbf{RGPSSM-H-P:} To demonstrate the effectiveness of the new adding criterion \eqref{eq:add_criterion} for non-stationary kernels, we also simulate another RGPSSM-H method in which the first dimension uses a Gaussian kernel and the second dimension uses an additive kernel as follows:

\begin{equation}
\begin{aligned}
    &K^2(t, t^\prime) = K_{\mathrm{RBF}}(t, t^\prime) + \bm\phi(t) \bm I \bm\phi(t^\prime) \\
    &\bm\phi(t) = 
    \begin{bmatrix}
        \cos(0.2t) & \cos(0.5t) & \cos(t) \\
    \end{bmatrix}
\end{aligned}
\end{equation}
This is a kernel that combines a Gaussian kernel with a basis function kernel. The basis function kernel is a non-stationary kernel and is used to represent cases where some structural prior knowledge is available.

\begin{figure}[H]
    \centering
    \includegraphics[width=0.98\linewidth]{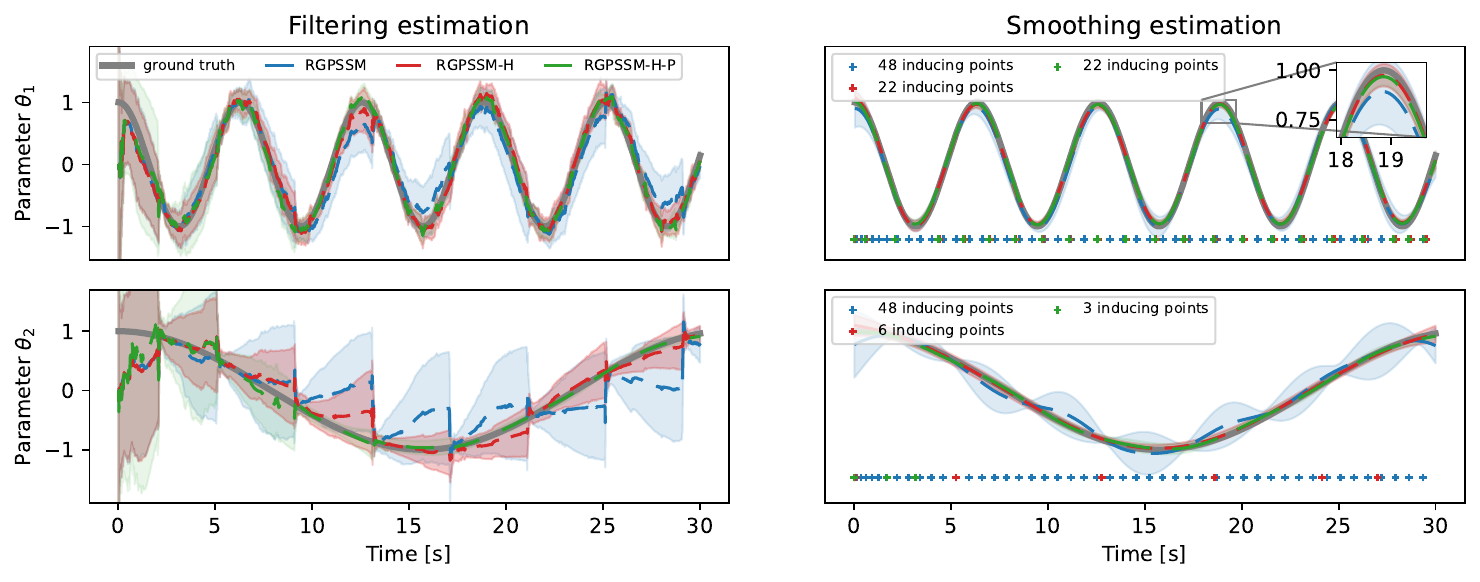}
    \caption{Filtering and smoothing estimation of time-varying parameters. Filtering estimation refers to $p(\bm\theta_t|\bm y_{1:t})$ and smoothing estimation refers to $p(\bm\theta_t|\bm y_{1:T})$, where $T$ is the total number of measurements. The shaded area indicates the 95\% confidence interval, and the '+' marker indicates the locations of the inducing points.}
    \label{fig:param_iden}
\end{figure}

\begin{figure}[H]
    \centering
    \includegraphics[width=0.6\linewidth]{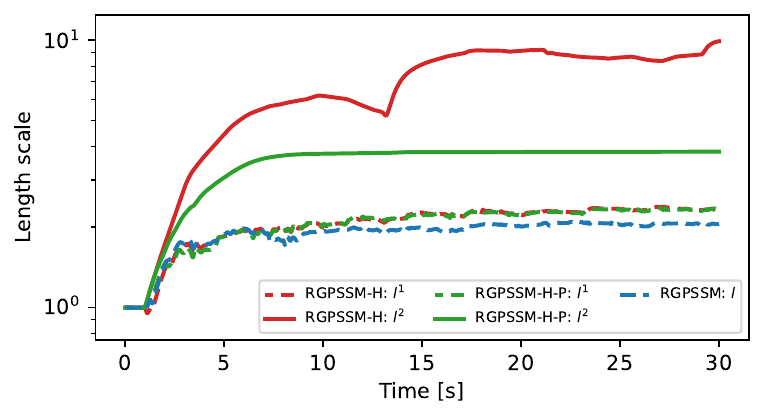}
    \caption{Evolution curves of the length scales of the kernel.}
    \label{fig:hyperparam}
\end{figure}

In this experiment, the length scales of Gaussian kernel are all initialized to $1$ and are optimized during learning by using the method developed in RGPSSM \cite{zheng2024recursive}. For fairness, the RGPSSM-H is implemented with EKF-based moment matching. 
The filtering and smoothing estimation profiles as well as the quantified accuracy are shown in Fig.~\ref{fig:param_iden} and Table \ref{tab:param_iden}, respectively, while the hyperparameter profiles are presented in Fig.~\ref{fig:hyperparam}.

\begin{table}[htbp]
    \centering
    \footnotesize
    \renewcommand{\arraystretch}{1.3}
    \begin{tabular}{lcccc}
        \toprule
        \multirow{2}{*}{\textbf{Methods}} & \multicolumn{2}{c}{\textbf{Filtering estimation}} & \multicolumn{2}{c}{\textbf{Smoothing estimation}} \\
        \cmidrule(lr){2-3} \cmidrule(lr){4-5}
         & $\boldsymbol{\theta_1}$~$(\text{nMSE}\downarrow \mid \text{MNLL}\downarrow)$ 
         & $\boldsymbol{\theta_2}$~$(\text{nMSE}\downarrow \mid \text{MNLL}\downarrow)$ 
         & $\boldsymbol{\theta_1}$~$(\text{nMSE}\downarrow \mid \text{MNLL}\downarrow)$ 
         & $\boldsymbol{\theta_2}$~$(\text{nMSE}\downarrow \mid \text{MNLL}\downarrow)$ \\
        \midrule
        RGPSSM & \cellcolor[HTML]{EDE8F0}0.0932 $\mid$ -0.3078 & \cellcolor[HTML]{EDE8F0}0.3778 $\mid$ 0.3089 & \cellcolor[HTML]{EDE8F0}0.0056 $\mid$ -1.6364 & \cellcolor[HTML]{EDE8F0}0.0167 $\mid$ -1.0665 \\
        RGPSSM-H & \cellcolor[HTML]{CABCD4}0.0380 $\mid$ -0.9138 & \cellcolor[HTML]{CABCD4}0.0950 $\mid$ -0.6587 & \cellcolor[HTML]{CABCD4}0.0019 $\mid$ -2.0674 & \cellcolor[HTML]{CABCD4}0.0010 $\mid$ -2.0624 \\
        RGPSSM-H-P & \cellcolor[HTML]{9E86B0}0.0251 $\mid$ -1.1853 & \cellcolor[HTML]{9E86B0}0.0500 $\mid$ -1.4870 & \cellcolor[HTML]{9E86B0}0.0016 $\mid$ -2.1465 & \cellcolor[HTML]{9E86B0}0.0009 $\mid$ -2.2616 \\
        \bottomrule
    \end{tabular}
    \normalsize
    \caption{Accuracy of filtering and smoothing estimation for time-varying parameters. The three lowest nMSE values are highlighted in shades of purple, with darker shades indicating lower nMSEs.}
    \label{tab:param_iden}
\end{table}

As shown in Fig.~\ref{fig:param_iden} and Table \ref{tab:param_iden}, for both filtering estimation and smoothing estimation, RGPSSM-H and RGPSSM-H-P provides more accurate mean estimates and smaller confidence intervals, compared to RGPSSM. This improvement is due to the heterogeneous learning capability of RGPSSM-H. 
Specifically, as shown in Fig.~\ref{fig:hyperparam}, RGPSSM-H can accurately learn the length scales for different function dimensions. This enables it to learn a larger length scale for $\theta_2$, thereby capturing its strong temporal correlation.
In contrast, RGPSSM uses only a shared length scale, which tends to fit the most sensitive $\theta_1$ dimension and limits its generalization ability. 
In addition, as shown in Fig.~\ref{fig:param_iden}, the locations of the inducing points are indicated by the "+" marker, and the total number for each dimension is shown in the legend. For the $\theta_1$ dimension, RGPSSM-H and RGPSSM-H-P have fewer inducing points than RGPSSM, although the length scales for this dimension are similar for these three methods, as shown in Fig.~\ref{fig:hyperparam}. This is achieved by pruning redundant points after the length scale increases. For the $\theta_2$ dimension, RGPSSM-H and RGPSSM-H-P have only a single-digit number of inducing points, which is significantly fewer than RGPSSM. This improvement is due to the independent adding criterion and redundant inducing point pruning procedure.
Furthermore, it is clear that RGPSSM-H-P achieves more accurate estimation and uses fewer inducing points. This demonstrates that the proposed method can effectively combine the computational efficiency and generalization benefits of traditional parametric approaches by incorporating prior structural knowledge into the kernel.
In summary, these results demonstrate that the proposed heterogeneous learning improvements effectively enhance both the learning accuracy and the efficiency of inducing points, compared to the original RGPSSM.





\subsection{Learning Hypersonic Vehicle Dynamics}\label{subsec:hypersonic}

Online learning of the aerodynamic model is a core component of most adaptive control methods for hypersonic vehicles \cite{2025Experience, xu2022predefined, bin2007adaptive}.
This subsection evaluates the practical learning capability of RGPSSM-H for the aerodynamic moment coefficient model of a hypersonic vehicle using the winged cone dataset \cite{shaughnessy1990hypersonic, keshmiri2005development}. 
The system model is given as follows:

\begin{equation}
\begin{aligned}
\dot{\bm \omega} = \bm I^{-1} \left[ -\bm\omega \times (\bm I \bm \omega) + \bm M \right]
\end{aligned}
\end{equation}
where $\bm\omega = [p, q, r]^T$ is the angular rate, with $p$, $q$, and $r$ representing the roll, pitch, and yaw rates, respectively. $\bm I$ is the inertia matrix, and "$\times$" denotes the cross product. $\bm M= [M_x, M_y, M_z]^T$ is the aerodynamic moment, which can be expressed as follows:

\begin{equation}\label{eq:C}
\begin{aligned}
    M_x&= Q S_{\mathrm{ref}} b C_l \\
    M_y&= Q S_{\mathrm{ref}} c C_m \\ 
    M_z&= Q S_{\mathrm{ref}} b C_n \\
    C_l& 
    = f_l\left(\alpha, \beta, \delta_a, \delta_e, \delta_r, p, r\right) 
    =C_{l,\beta}\beta + C_{l,\delta_a} + C_{l,\delta_e} + C_{l,\delta_r} 
    + C_{l,p}\frac{pb}{2V} + C_{l,r}\frac{rb}{2V}
    \\
    C_m& 
    = f_m\left(\alpha, \delta_a, \delta_e, \delta_r, q\right) 
    =C_{m,\alpha}\alpha + C_{m,\delta_a} + C_{m,\delta_e} + C_{m,\delta_r} 
    + C_{m,q}\frac{qc}{2V}
    \\
    C_n& 
    = f_n\left(\alpha, \beta, \delta_a, \delta_e, \delta_r, p, r\right) 
    =C_{n,\beta}\beta + C_{n,\delta_a} + C_{n,\delta_e} + C_{n,\delta_r} 
    + C_{n,p}\frac{pb}{2V} + C_{n,r}\frac{rb}{2V}
    \\
\end{aligned}
\end{equation}
where $Q$ is the dynamic pressure, $V$ is the velocity, $S_{\mathrm{ref}}$ is the reference area, $b$ is the span, $c$ is the mean aerodynamic chord, and $C_l$, $C_m$, and $C_n$ are the roll, pitch, and yaw moment coefficients, respectively. These coefficients depend on the angle of attack $\alpha$, sideslip angle $\beta$, right aileron deflection $\delta_a$, left aileron deflection $\delta_e$, rudder deflection $\delta_r$, and the angular rates $p$, $q$, and $r$. 
In this experiment, we simulate the vehicle under hypersonic cruise conditions with constant altitude ($h=110000\mathrm{ft}$) and constant speed (Ma=15).
Therefore, the Mach number and velocity are not included in the input variables of the moment coefficient $\bm C$ in \eqref{eq:C}. 
In addition, we assume that several states and model parameters are precisely known, including the states $Q$, $\alpha$, and $\beta$, as well as the parameters $S_{\mathrm{ref}} = 3603 \mathrm{ft}^2$, $b = 60\mathrm{ft}$, $c = 80\mathrm{ft}$, and the inertia matrix $\bm I$, which is given as follows:

\begin{equation}
\begin{aligned}
\bm I = \begin{bmatrix}
    I_{xx} & -I_{xy} & -I_{xz} \\
    -I_{xy} & I_{yy} & -I_{yz} \\
    -I_{xz} & -I_{yz} & I_{zz} \\
    \end{bmatrix}
\end{aligned}
\end{equation}
where $I_{xx} = 6.4\times 10^5 \mathrm{slug}\cdot\mathrm{ft}^2$, $I_{yy}=7\times 10^6 \mathrm{slug}\cdot\mathrm{ft}^2$, $I_{zz} = 9\times 10^6 \mathrm{slug}\cdot\mathrm{ft}^2$, $I_{xz}=1\times 10^5 \mathrm{slug}\cdot\mathrm{ft}^2$, and $I_{xy} = I_{yz} = 0$. 
To simulate aerodynamic uncertainty, the actual moment coefficient is set as $\bm C(\cdot) = \bm C_0(\cdot) + \Delta \bm C(\cdot)$, where $\bm C_0(\cdot)$ is the known baseline model constructed from the winged cone dataset \cite{shaughnessy1990hypersonic, keshmiri2005development}. 
Additionally, the model bias $\Delta \bm C(\cdot)$ is constructed by adjusting the sub-coefficients in \eqref{eq:C}.
Specifically, the sub-coefficients related to control surface deflections $\delta_a, \delta_e, \delta_r$ are increased by $30$\%, while the sub-coefficients related to aerodynamic angles $\alpha$, $\beta$, and angular rates $\bm\omega$ are decreased by $30$\%. 
This bias setting ensures that the actual dynamics remain strongly controllable.  
Therefore, given this uncertainty condition, the learning objective is to identify the model bias $\Delta \bm C(\cdot)$ from  angular rate measurements with moderate noise intensity, where the signal-to-noise ratio $\mathrm{SNR}=15\,\mathrm{dB}$.

\begin{figure}[h]
    \centering
    \includegraphics[width=0.99\linewidth]{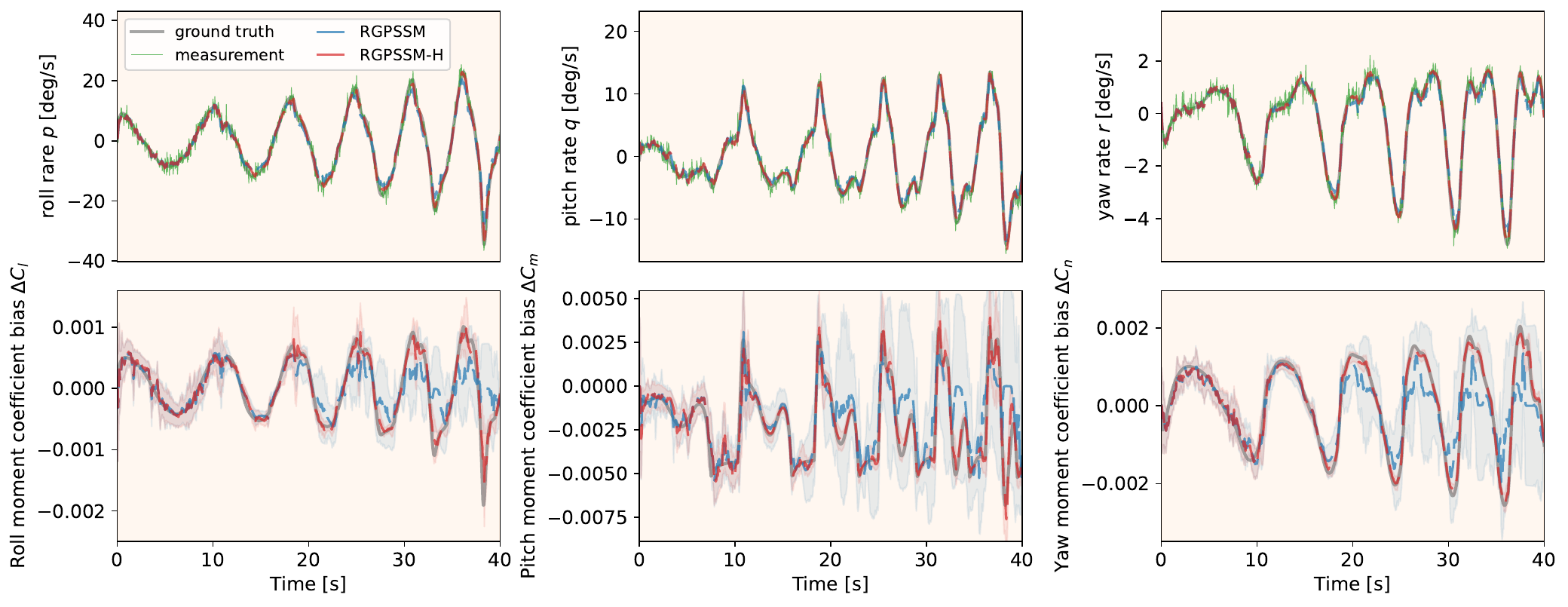}
    \caption{First row: filtering estimation of the angular rate $\bm\omega$. Second row: prediction of the moment coefficient model bias $\Delta \bm C(\cdot)$ at each prediction step. The shaded area indicates the 95\% confidence interval.}
    \label{fig:wc_learn}
\end{figure}

\begin{figure}[h]
    \centering
    \includegraphics[width=0.98\linewidth]{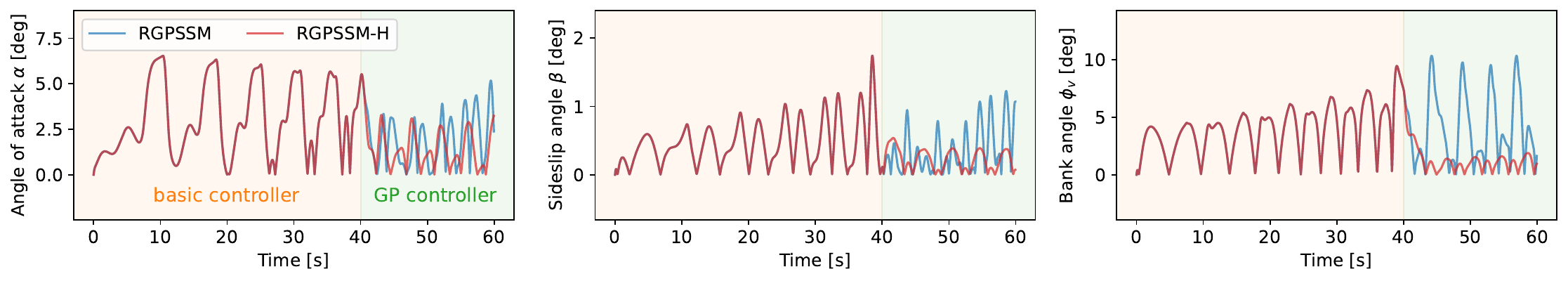}
    \caption{Angle tracking error. The controller in the first 40 seconds is based on the baseline moment coefficient model $\bm C_0(\cdot)$, and after the 40-th second, it is based on the real-time learned model $\bm C_0(\cdot) + \bm m(\cdot)$.}
    \label{fig:wc_ctrl}
\end{figure}

Furthermore, to evaluate the advantages of online learning for the aerodynamics model, we incorporate the model learned in real time into the attitude controller. 
The control task is to track sinusoidal signals for the angle of attack $\alpha$, sideslip angle $\beta$, and bank angle $\phi_v$.
Since the design of a sophisticated adaptive controller is not the focus of this paper, we use a switching controller with a simple adaptive mechanism to demonstrate the benefits of the learned model.
Specifically, we use a dynamic inversion controller based on the following approximation for the angular acceleration model:

\begin{equation}\label{eq:model_wc}
\begin{aligned}
    \dot{\bm \omega} (\bm s, \bm \delta) \approx \dot{\bm \omega} (\bm s, \bm 0) + \bm G \bm \delta
\end{aligned}
\end{equation}
where $\bm s$ represents the model input variables excluding the control surface deflections $\bm\delta= [\delta_a, \delta_e, \delta_r]^T$. $\bm G = \frac{\partial \dot{\bm \omega}(\bm s, \bm 0)}{\partial \bm \delta}$ denotes the control effectiveness matrix, which is computed via numerical differentiation for convenience. 
This linearization is justified by the near-linear dependence between the moment coefficient $\bm C$ and the control surface deflections $\bm\delta$ observed in the winged cone dataset.
 Given the model \eqref{eq:model_wc}, the control law is designed as follows:

\begin{equation}\label{eq:delta_wc}
\begin{aligned}
    \bm\delta = \bm G^{-1} \left(\dot{\bm \omega}_d - \dot{\bm \omega}(\bm s, \bm 0) \right)
\end{aligned}
\end{equation}
where $\dot{\bm \omega}_d$ is the desired angular acceleration, which can be obtained using the dynamic surface method \cite{bin2007adaptive}. 
To warm up the learning process, during the first 40 seconds, the baseline aerodynamic moment model $\bm C_0(\cdot)$ is used to compute the control surface deflection $\bm \delta$. After the 40-th second, the controller switches to using the learned model $\bm C_0(\cdot) + \bm m(\cdot)$, where $\bm m(\cdot)$ is the mean function of the GP obtained in real time. 
The RGPSSM and RGPSSM-H (ADF) are employed for the learning task. 
To address the differences in input variables among the different channels, as shown in \eqref{eq:C}, the RGPSSM uses an isomorphic kernel whose input variable is the union of all the channels. In contrast, the RGPSSM-H addresses this in a straightforward manner by using the heterogeneous kernel in \eqref{eq:S0} with the relevant variables for each dimension.
The maximum size of the inducing points is limited to 90. 
The learning and control performance are depicted in Fig.~\ref{fig:wc_learn} and Fig.~\ref{fig:wc_ctrl}.

As shown in Fig.~\ref{fig:wc_learn}, RGPSSM-H achieves more accurate moment coefficient predictions than RGPSSM. 
This is primarily because RGPSSM-H allows the input variables to be tailored to each output dimension, which reduces the requirements for data and inducing points compared to RGPSSM.
In addition, RGPSSM-H allows the use of different kernel length scales and inducing points for each dimension, which further enhances learning accuracy. Based on the learned model, the control performance is illustrated in Fig.~\ref{fig:wc_ctrl}. It is evident that when using the baseline aerodynamic model to evaluate control surface deflection $\bm\delta$, the tracking accuracy is limited. 
In contrast, when the learned model from RGPSSM or RGPSSM-H is used, the control accuracy improves in most channels.
In particular, the controller based on RGPSSM-H achieves better control performance because it provides a more accurate aerodynamic model.
These results demonstrate the effectiveness of the proposed method in improving sample efficiency and learning accuracy for multi-output transition model learning, as well as its potential benefits for enhancing the adaptive capability of the hypersonic vehicle controller.

\subsection{Learning Real World Quadrotor Dynamics}\label{subsec:exp_quadrotor}

\begin{figure}[htbp]
    \centering
    \begin{minipage}[t]{0.24\textheight}
        \centering
        \includegraphics[width=\textwidth]{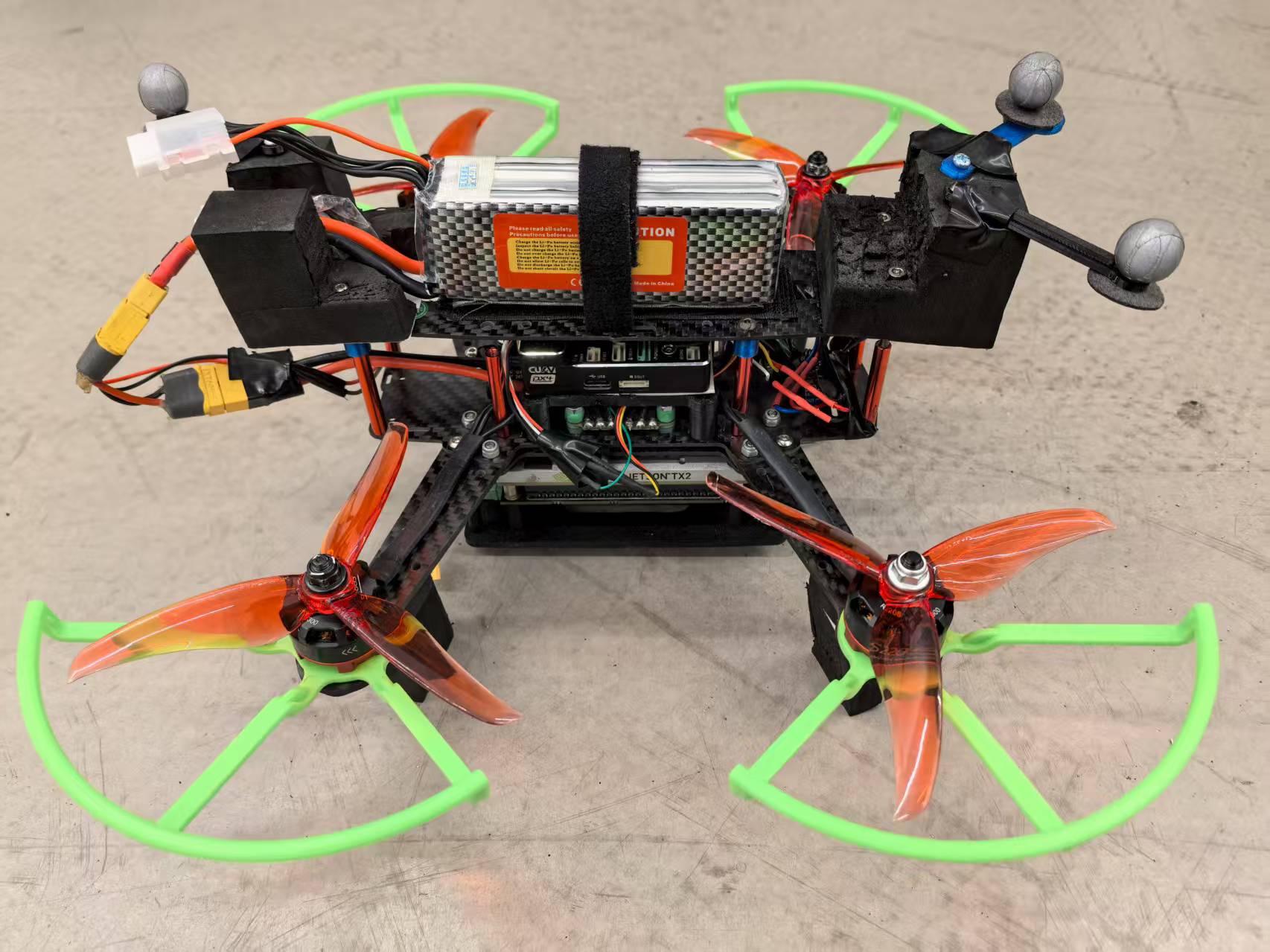}
        \caption*{(a) Quadrotor}
    \end{minipage}
    \begin{minipage}[t]{0.205\textheight}
        \centering
        \includegraphics[width=\textwidth]{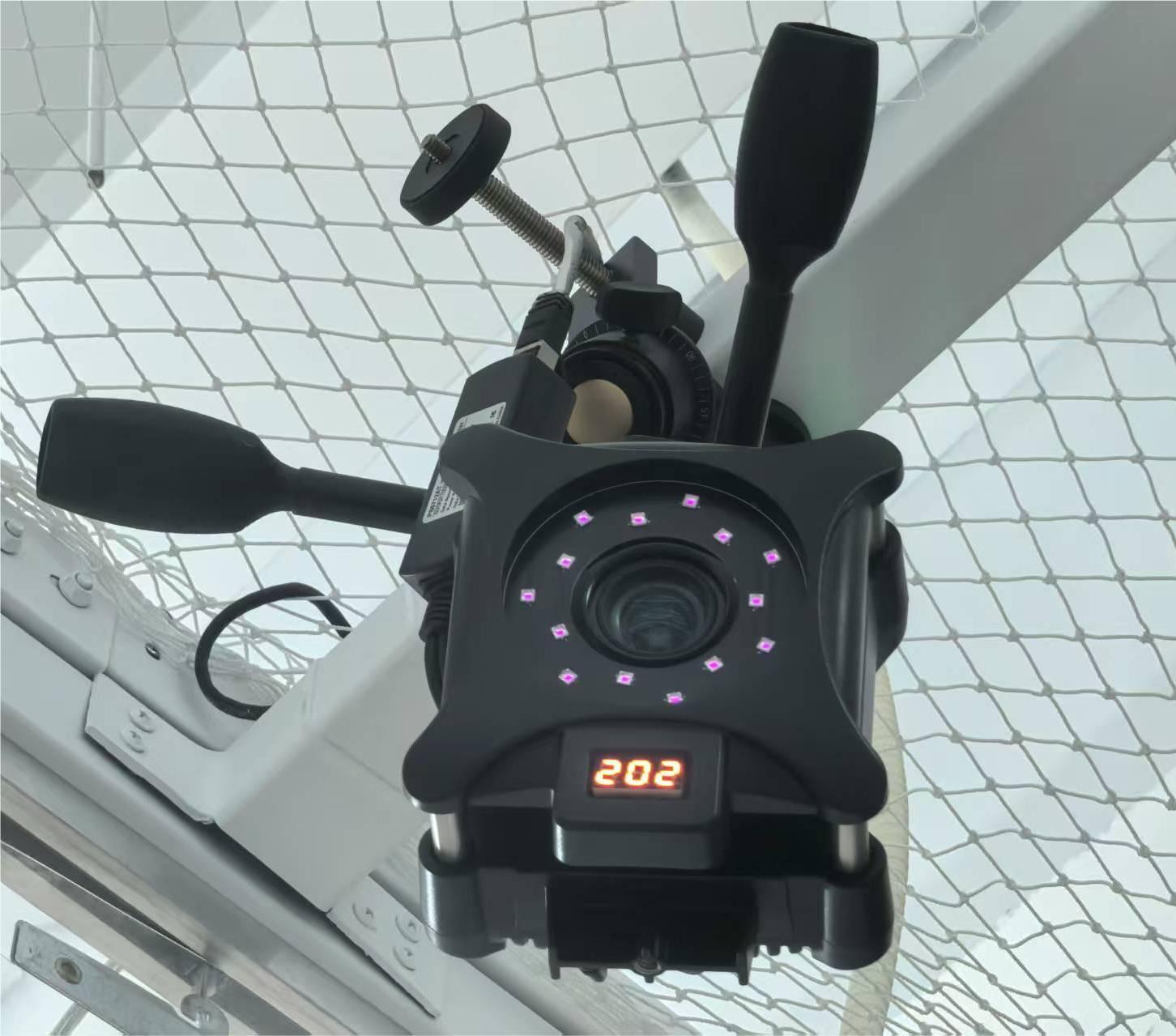}
        \caption*{(b) Motion capture system}
    \end{minipage}
    \begin{minipage}[t]{0.22\textheight}
        \centering
        \includegraphics[width=\textwidth]{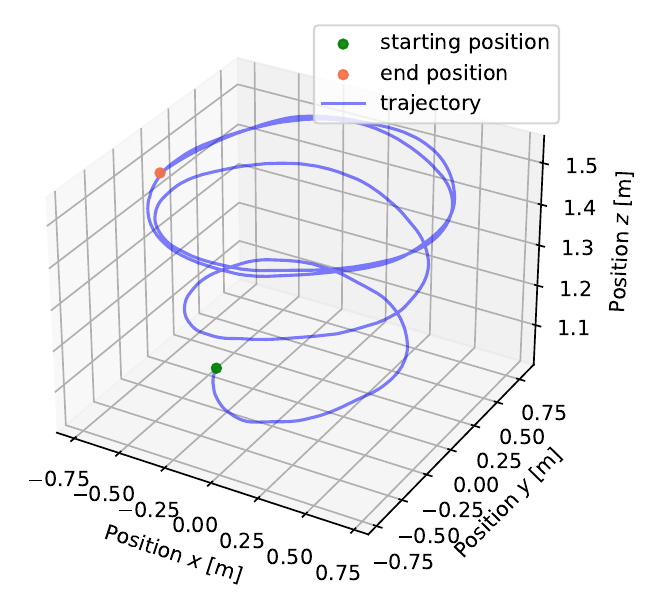}
        \caption*{(c) Flight trajectory}
    \end{minipage}
    \caption{Real-world quadrotor, data acquisition equipment, and flight trajectory}
    \label{fig:quadrotor}
\end{figure}

To further demonstrate the practical capability of the proposed method, we apply it to a real-world quadrotor dataset. Specifically, the learning data are collected from a self-assembled quadrotor, which is controlled to follow a helical trajectory using the PX4 autopilot control algorithm.
Learning data are obtained through onboard sensors and a motion capture system. The actual device and flight trajectory are shown in Fig.~\ref{fig:quadrotor}.

\begin{figure}[htbp]
    \centering
    \includegraphics[width=0.99\linewidth]{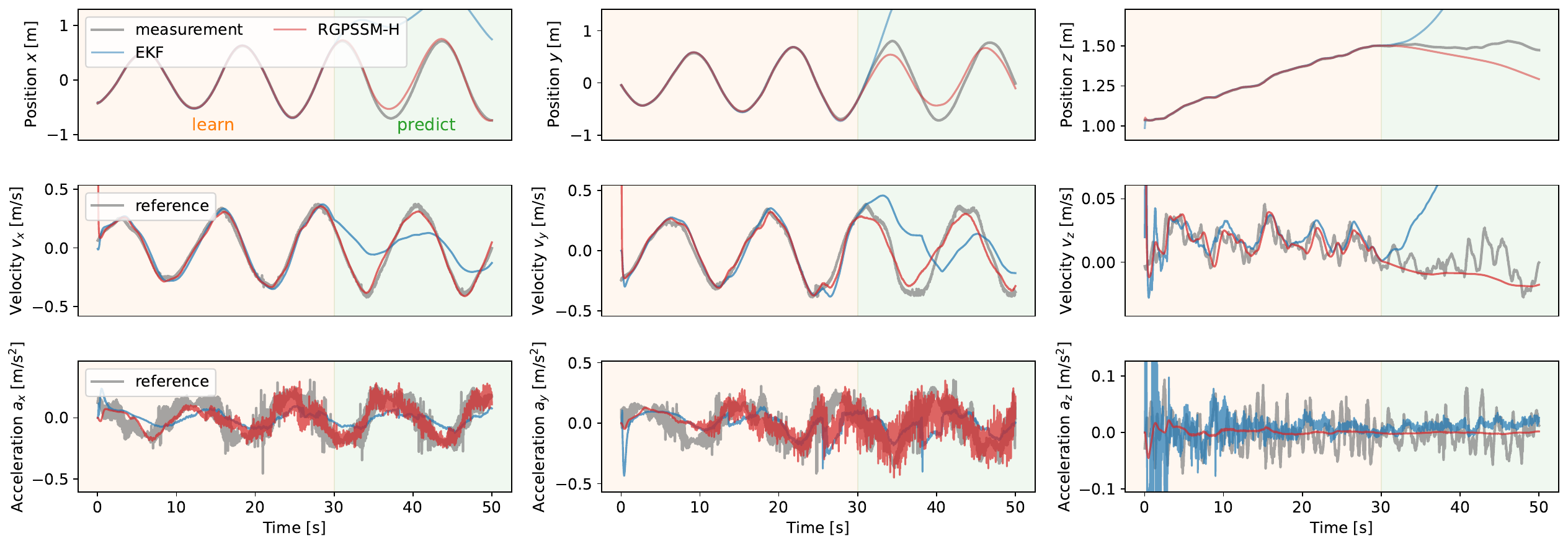}
    \caption{State estimation and prediction results for the quadrotor. The reference velocity and acceleration are computed by numerically differentiating the smoothed position and velocity trajectories, where smoothing is performed using a moving average filter.}
    \label{fig:learn_quadrotor}
\end{figure}

In this experiment, the learning objective is the translational acceleration model and the system model is given by:

\begin{equation}
\begin{aligned}
    \dot{\bm p} &= \bm v \\
    \dot{\bm v} &= \bm a(\bm u_{\mathrm{pwm}}, \bm q) \\
\end{aligned}
\end{equation}
where $\bm p \in \mathbb R^3$ and $\bm v \in \mathbb R^3$ are the three-axis position and velocity in the world coordinate system. The acceleration model $\bm a(\cdot)$ is defined as a function of the PWM command $\bm u_{\mathrm{pwm}} \in \mathbb{R}^4$ and the attitude quaternion $\bm q \in \mathbb{R}^4$. 
Since the flight is at low speed due to space constraints, the velocity-dependent drag terms are not considered.
The RGPSSM-H and an EKF-based parametric learning method are employed to learn the acceleration model from position measurements, which are obtained by the motion capture system at 50 Hz. Specifically, for the RGPSSM-H, the joint variable $(\bm u_{\mathrm{pwm}}, \bm q)$ is used as model input for all three function dimensions, but different length scales can be learned for each dimension. 
In addition, since the learned dynamics are not strongly nonlinear, EKF-based moment matching is employed to pursue real-time performance.
As for the EKF-based parametric method, the following model structure is used:

\begin{equation}
\begin{aligned}
    &\bm a = a_{\mathrm{thrust}} \cdot \vec{\bm k} + \bm b_{\mathrm{ekf}} \\
    &a_{\mathrm{thrust}} = \begin{bmatrix}
        1 & \sum\limits_{i=1}^4 \bm u_{\mathrm{pwm}, i} & \sum\limits_{i=1}^4 \bm u_{\mathrm{pwm}, i}^2 \\
        \end{bmatrix}
    \bm w_{\mathrm{ekf}}
\end{aligned}
\end{equation}
where $a_{\mathrm{thrust}} \in \mathbb R$ denotes the thrust acceleration magnitude, which is modeled as a quadratic polynomial with respect to the PWM command $\bm u_{\mathrm{pwm}}$.
$\vec{\bm k}$ is the unit vector of the body $z$-axis expressed in the world coordinate system, which is obtained from the quaternion $\bm q$.
Therefore, by identifying the parameters $\bm w_{\mathrm{ekf}} \in \mathbb{R}^3$ and $\bm b_{\mathrm{ekf}} \in \mathbb{R}^3$ using the EKF, we can learn the quadrotor dynamics. 
Specifically, these parameters are augmented into the state space and are jointly inferred with the quadrotor state via the EKF.
In this experiment, we use a dataset with a total duration of 50 seconds, consisting of data pairs $(\bm u_{\mathrm{pwm}}, \bm q, \bm p)$. 
To evaluate the learning accuracy, the first 30 seconds of measurements are used for learning. While after the 30-th second, the algorithm begins to perform rolling prediction, which is implemented by executing only the prediction step without the correction step.
The learning results are summarized in Fig.~\ref{fig:learn_quadrotor} and Table \ref{tab:learn_quadrotor}.

\begin{table}[htbp]
    \centering
    \footnotesize
    \begin{tabular}{lcccc}
        \toprule
        Method 
        & $x$ (nMSE$\downarrow$ $\mid$ MNLL$\downarrow$) 
        & $y$ (nMSE$\downarrow$ $\mid$ MNLL$\downarrow$) 
        & $z$ (nMSE$\downarrow$ $\mid$ MNLL$\downarrow$) & Time (s)$\downarrow$ \\
        \midrule
        EKF & 5.5722 $\mid$ 2.6556 & 11.8852 $\mid$ 2.1156 & 4003.8604 $\mid$ 0.8583 & \textbf{4.27} \\
        RGPSSM-H & \textbf{0.0303} $\mid$ \textbf{0.9171} & \textbf{0.1185} $\mid$ \textbf{0.9857} & \textbf{55.3195} $\mid$ \textbf{0.7163} & 15.67 \\
        \bottomrule
    \end{tabular}
    \normalsize
    \caption{Prediction performance for the position trajectory and total learning time for quadrotor dynamics using EKF and RGPSSM-H. The column "Time" indicates the total runtime required to process 30 seconds of measurements.}
    \label{tab:learn_quadrotor}
\end{table}

As shown in Fig.~\ref{fig:learn_quadrotor}, the RGPSSM-H achieves accurate trajectory prediction after the $30$-th second, whereas the prediction of the parametric-based method diverges rapidly. The prediction accuracy for the position trajectory is further illustrated in Table \ref{tab:learn_quadrotor}.
The difference in performance for the two methods is mainly due to the fact that the GP provides more flexible representational capability, and the RGPSSM-H algorithm offers excellent learning accuracy and sample efficiency.
In contrast, for the EKF-based parametric method, although it also has superior sample efficiency and generalization ability, it is difficult to establish an accurate model structure for real-world systems due to factors such as installation deviations and component-to-component variability. For example, there may be variations among the four motors of the quadrotor. 
Therefore, the parametric method finds it difficult to accurately capture the system dynamics in real-world scenarios.
In terms of computational efficiency, Table \ref{tab:learn_quadrotor} shows that the RGPSSM-H requires only four times the runtime of the EKF. This demonstrates its superior computational efficiency and strong potential for real-world applications.
In summary, this experiment demonstrates the flexible and accurate learning capability of RGPSSM-H for real-world quadrotor dynamics, as well as its potential for real-world deployment.

\section{Conclusion}

In this paper, we develop an extension of the RGPSSM method \cite{zheng2024recursive} for the heterogeneous multi-output learning setting and general moment matching techniques. First, to enhance heterogeneous learning, we propose a heterogeneous multi-output kernel that enables different kernel types, hyperparameters, and input variables for each function dimension. 
Second, to improve learning efficiency and numerical stability, 
the management algorithm for inducing points is improved to allow independent selection for each dimension and to prune redundant points. 
Third, to support the application of general moment matching techniques, we derive a new approximate inference framework that does not rely on first-order linearization.
Fourth, we present detailed implementations for several moment matching methods, including the EKF, UKF, and ADF, all of which are more theoretically appropriate and general compared to existing studies. 
Fifth, we provide a Cholesky-based stable implementation method, and complete code to facilitate practical application. 
Finally, experiments on synthetic and real-world datasets demonstrate that RGPSSM-H offers greater flexibility, sample efficiency, and learning accuracy. The results in benchmark tests (Section \ref{subsec:kink}) indicate that, compared to SOTA offline GPSSM approaches, the proposed method achieves comparable accuracy with only $1/100$ of the runtime. In addition, compared to SOTA online GPSSM methods, the proposed method achieves approximately $70\%$ higher accuracy under significant noise conditions and requires only $1/20$ of the runtime.

\section{Acknowledgement}
This work is supported by the National Natural Science Foundation of China (Grant No. 12572052) and the Beijing Natural Science Foundation (Grant L251013).

\section*{Appendix}
\appendix                                           
\renewcommand{\thesubsection}{\Alph{subsection}}

\setcounter{subsection}{0}
\subsection{Evaluation Metrics: MSE and MNLL}\label{subsec:metric}

\textbf{Normalized Mean Squared Error (nMSE)} is the mean squared error (MSE) normalized by the variance of the true values:
\begin{equation}
    \mathrm{nMSE} = \frac{\frac{1}{n} \sum_{i=1}^n (x_i - \hat{x}_i)^2}{\mathrm{var}(x)}
\end{equation}
where $x_i$ is the true value, $\hat{x}_i$ is the predicted value, $n$ is the number of samples, and $\mathrm{var}(x)$ is the variance of the true values.

\textbf{Mean Negative Log-Likelihood (MNLL)} evaluates how well the predicted probability distribution matches the observed data. Assuming the model predicts a Gaussian distribution for each observation with mean $\hat{x}_i$ and standard deviation $\sigma_i$, the MNLL is defined as:
\begin{equation}
    \mathrm{MNLL} = \frac{1}{n} \sum_{i=1}^n \left[
        \frac{1}{2} \left(
            \frac{(x_i - \hat{x}_i)^2}{\sigma_i^2}
            + 2\log \sigma_i
            + \log(2\pi)
        \right)
    \right]
\end{equation}
where $x_i$ is the true value, $\hat{x}_i$ is the predicted mean, and $\sigma_i$ is the predicted standard deviation. Lower MNLL values indicate that the predicted distribution is closer to the true data.







\subsection{Derivation of the Moments in the ADF-based Moment Matching}\label{app:adf}

This subsection provides detailed derivations for each of the moments in \eqref{eq:moment}.

\subsubsection{\texorpdfstring{\(\bm m_h\)}{m_h}}
    
Given \eqref{eq:ph_xu_1} and \eqref{eq:moment}, the $k$-th element of the moment $\bm m_h$ can be given by:

\begin{equation}
\begin{aligned}
    \bm m_{h^k} 
    &= \mathbb{E}[\bm K_{h^k u^k} \bm{v}^k] \\
    &= \mathbb{E} \left[\sum_{j=1} \bm K_{h^k u^k_j} \bm{v}^k_j \right] \\
    &= \sum_{j=1}^{n_{u^k}} \mathbb{E} [\bm K_{h^k u^k_j} \bm{v}^k_j] \\
\end{aligned}
\end{equation}
Using the definition of $\bm K_{h^k u^k_j}$ as illustrated in \eqref{eq:phi} and the property \eqref{eq:property1}, we have:

\begin{equation}\label{eq:phi_v}
\begin{aligned}
    \mathbb{E} [\bm K_{h^k u^k_j} \bm{v}^k_j] 
    &= c^k\, \mathbb{E} \left[
        \mathcal{N}(\bm z_j^k \mid \bm z^k, \bm{\Lambda}^k)\, \bm{v}_j^k
    \right] \\[1.ex]
    &= c^k 
    \int \left[ \mathcal{N}(\bm z_j^k \mid \bm z^k, \bm{\Lambda}^k)\;
    q(\bm{z}^k, \bm{v}^k)\;
    \bm{v}_j^k\; \right]
    \, \mathrm{d}\bm{z}^k\, \mathrm{d}\bm v^k \\[1.ex]
\end{aligned}
\end{equation}
Here, $\mathcal{N}(\bm z_j^k \mid \bm z^k, \bm{\Lambda}^k)$ can be regarded as the data likelihood, and $q(\bm{z}^k, \bm{v}^k)$ as the prior distribution. Therefore, we can use Bayesian inference to perform the following transformation:

\begin{equation}\label{eq:bayes}
\begin{aligned}
    \mathcal{N}(\bm z_j^k \mid \bm z^k, \bm{\Lambda}^k)\; q(\bm{z}^k, \bm{v}^k)
    = q(\bm z^k_j)\; q(\bm{z}^k, \bm{v}^k \mid \bm z_j^k)
\end{aligned}
\end{equation}
where $q(\bm z^k_j)$ is the marginal likelihood and $ q(\bm{z}^k, \bm{v}^k \mid \bm z_j^k)$ is the posterior of $(\bm{z}^k, \bm{v}^k)$ given $\bm z_j^k$.
Hence, by substituting \eqref{eq:bayes} into \eqref{eq:phi_v}, we obtain:

\begin{equation}\label{eq:phi_v1}
    \begin{aligned}
        \mathbb{E} [\bm K_{h^k u^k_j} \bm{v}^k_j] 
        &= c^k\, q(\bm z^k_j)\;
        \int \left[ q(\bm{z}^k, \bm{v}^k \mid \bm z_j^k)\;
        \bm{v}_j^k\; \right]
        \, \mathrm{d}\bm{z}^k\, \mathrm{d}\bm v^k \\[1.ex]
        &= b_j^{k}\; \bm m_{v_j^k|z_j^k}
    \end{aligned}
\end{equation}
where the constant $b_j^{k} = c^k q(\bm z_j^k)$, whose expression can be obtained by using the definition of $c^k$ as illustrated in \eqref{eq:property1}: 
\begin{equation}\label{eq:b}
    \begin{aligned}
        b_j^k
        &= \sigma^2_k \cdot (2\pi)^{d_{z^k}/2} \, \vert \bm \Lambda^k \vert^{1/2} \cdot
        \mathcal N(\bm z_j^k \mid \bm{m}_{z^k}, \bm{\Lambda}^k + \bm S_{z^kz^k}) \\
        &= \sigma^2_k \cdot \left| \bm I + \bm S_{z^kz^k} (\bm\Lambda^{k})^{-1} \right|^{-1/2} \cdot
        \exp\left[-\frac{1}{2} \left(\bm{z}_j^k - \bm{m}_{z^k}\right)^T 
        \left(\bm{\Lambda}^k + \bm S_{z^kz^k}\right)^{-1} 
        \left(\bm{z}_j^k - \bm{m}_{z^k}\right)\right]
    \end{aligned}
\end{equation}
and $\bm m_{v_j^k|z_j^k} = \mathbb{E}_{q(\bm{z}^k, \bm{v}^k \mid \bm z_j^k)}[\bm v_j^k]$, which can be evaluated using the Kalman filter equation. 
Generally, the moments of $q(\bm a, \bm b \mid \bm z_j^k)$ can be obtained by the following equation:

\begin{equation}\label{eq:moment_kf1}
    \begin{aligned}
        \bm{m}_{a|z_j^k} &= 
        \bm m_a + \bm{S}_{az^k} (\bm{S}_{z^kz^k} + \bm{\Lambda}^k)^{-1} (\bm z_j^k - \bm m_{z^k}) \\
        \bm{S}_{ab|z_j^k} &= 
        \bm{S}_{ab} - \bm{S}_{az^k} (\bm{S}_{z^kz^k} + \bm{\Lambda}^k)^{-1} \bm{S}_{z^kb} \\
\end{aligned}
\end{equation}
Here, $\bm m_{a|z_j^k} = \mathbb{E}_{q(\bm{a}, \bm b \mid \bm z_j^k)}[\bm a]$ and $\bm{S}_{ab|z_j^k} = \mathrm{cov}_{q(\bm{a}, \bm{b} \mid \bm z_j^k)}[\bm a, \bm b]$, where the random variables $\bm a$ and $\bm b$ can be replaced by $\bm x$, $\bm z$, $\bm u$, or $\bm v$ in order to obtain the corresponding moments.


\subsubsection{\texorpdfstring{\(\bm S_{hh}\)}{S_hh}}

Given \eqref{eq:moment}, $\bm S_{hh}$ can be decomposed into three terms: the first term $\bm S_{hh,1} = \mathbb{E}[\bm \Sigma_{\mathrm{gp}}(\bm{z})]$, the second term $\bm S_{hh,2} = \mathbb{E}[\bm \mu_{\mathrm{gp}}(\bm{z}, \bm v)\bm \mu_{\mathrm{gp}}(\bm{z}, \bm v)^T]$, and the third term $\bm S_{hh,3} = \bm m_h \bm m_h^T$. We derive the first two terms below.

For the first term $\bm S_{hh,1}$, according to \eqref{eq:ph_xu_1}, the off-diagonal elements are zero, and for the diagonal elements, we have:

\begin{equation}
\begin{aligned}
    \bm S_{h^kh^k,1}
    &= \bm K_{h^kh^k} - \mathbb{E} [\bm K_{h^k u^k} \bm{Q}^{kk} \bm K_{u^k h^k}] \\
\end{aligned}
\end{equation}
where $\bm K_{h^kh^k} = \sigma_k^2$ is a constant and:

\begin{equation}
\begin{aligned}
    \mathbb{E} [\bm K_{h^k u^k} \bm{Q}^{kk} \bm K_{u^k h^k}] = 
    \sum_{i=1}^{n_{u^k}} \sum_{j=1}^{n_{u^l}}
    \bm{Q}_{ij}^{kk} \cdot \mathbb{E} [\bm K_{h^k u^k_i} \bm K_{h^k u^k_j}] \\
\end{aligned}
\end{equation}
The expectation term can be evaluated by using the property \eqref{eq:property2}:

\begin{equation}\label{eq:dual_phi}
\begin{aligned}
    \mathbb{E} [\bm K_{h^k u^k_i} \bm K_{h^k u^k_j}]
    &= \int c^{kl} \mathcal{N}({\bm Z}_{ij}^{kl} | \bm Z^{kl}, \bm\Lambda^{kl}) \cdot
        q(\bm{Z}^{kl})  \mathrm{d}\bm{Z}^{kl} \\
    &= c^{kl} q({\bm Z}_{ij}^{kl}) \\
    &= B_{ij}^{kl} \\
\end{aligned}
\end{equation}
where $\bm Z^{kl}_{ij} = [(\bm z^{k}_i)^T, (\bm z^{l}_j)^T]^T$ and $q({\bm Z}_{ij}^{kl})= \int \mathcal{N}({\bm Z}_{ij}^{kl} | \bm Z^{kl}, \bm\Lambda^{kl})  q(\bm{Z}^{kl}) \mathrm{d}\bm{Z}^{kl}$ is the marginal likelihood. Given the definition of $c^{kl}$ as illustrated in \eqref{eq:property2}, we have:

\begin{equation}\label{eq:B}
\begin{aligned}
    B_{ij}^{kl} &= \sigma_k^2 \sigma_l^2 \cdot \vert \bm I + \bm S_{Z^{kl}Z^{kl}} (\bm\Lambda^{kl})^{-1} \vert^{-1/2} \cdot
    \exp\left[-\frac{1}{2} \left(\bm{Z}_{ij}^{kl} - \bm{m}_{Z^{kl}} \right)^T 
    \left(\bm{\Lambda}^{kl} + \bm S_{Z^{kl}Z^{kl}} \right)^{-1} 
    \left(\bm{Z}_{ij}^{kl} - \bm{m}_{Z^{kl}} \right)\right]
\end{aligned}
\end{equation}
can be derived similar to \eqref{eq:b}.

For the second term $\bm S_{hh,2}$, according to \eqref{eq:ph_xu_1}, the $(k,l)$-th element of the matrix can be given by:

\begin{equation}
\begin{aligned}
    S_{h^kh^l, 2} 
    &= \mathbb{E} [\bm K_{h^k u^k} \bm{v}^k \bm{v}^{l,T} \bm K_{h^l u^l}^T] \\
    &= \sum_{i=1}^{n_{u^k}} \sum_{j=1}^{n_{u^l}} \mathbb{E} [ 
        \bm K_{h^k u^k_i} \bm K_{h^l u^l_j} \bm{v}_i^k \bm{v}_j^l] \\
\end{aligned}
\end{equation}
The expectation term can be evaluated by using the property \eqref{eq:property2} and the result in \eqref{eq:dual_phi}, specifically:

\begin{equation}
\begin{aligned}
    \mathbb{E} [ \bm K_{h^k u^k_i} \bm K_{h^l u^l_j} \bm{v}_i^k \bm{v}_j^l]
    &= \int c^{kl} \mathcal{N}({\bm Z}_{ij}^{kl} | \bm Z^{kl}, \bm\Lambda^{kl}) \cdot
    q(\bm Z^{kl}, \bm v) 
    \, \bm v_i^k \bm v_j^l \mathrm{d}\bm{Z}^{kl} \mathrm{d}\bm{v} \\
    &= c^{kl} q({\bm Z}_{ij}^{kl}) 
    \int q(\bm Z^{kl}, \bm v | \bm Z_{ij}^{kl}) 
    \, \bm v_i^k \bm v_j^l \mathrm{d}\bm{Z}^{kl} \mathrm{d}\bm{v} \\
    &= B_{ij}^{kl} \int  q(\bm Z^{kl}, \bm v | \bm Z_{ij}^{kl}) \, \bm v_i^k \bm v_j^l \mathrm{d}\bm{Z}^{kl} \mathrm{d}\bm{v} \\
    &= B_{ij}^{kl} \left( \bm{S}_{v_i^k v_j^l|Z_{ij}^{kl}} 
    + \bm{m}_{v_i^k|Z_{ij}^{kl}} \bm{m}_{v_j^l|Z_{ij}^{kl}} \right)
\end{aligned}
\end{equation}
where the conditional moments can be evaluated using the Kalman filter equations, similar to \eqref{eq:moment_kf1}:


\begin{equation}\label{eq:moment_kf2}
\begin{aligned}
    \bm m_{a|Z_{ij}^{kl}} &= \bm m_{a} + \bm S_{aZ^{kl}} (\bm S_{Z^{kl}Z^{kl}} + \bm{\Lambda}^{kl})^{-1} (\bm Z_{ij}^{kl} - \bm{m}_{Z^{kl}}) \\
    \bm S_{a b|Z_{ij}^{kl}} 
    &= \bm S_{a b} - \bm S_{a Z^{kl}} 
    (\bm S_{Z^{kl}Z^{kl}} + \bm{\Lambda}^{kl})^{-1} \bm S_{Z^{kl}b} \\
\end{aligned}
\end{equation}
where $\bm a$ and $\bm b$ can be replaced by $\bm x$, $\bm z$, $\bm u$, or $\bm v$ to obtain the corresponding moments.





\subsubsection{\texorpdfstring{$\bm S_{hx}$ and $\bm S_{hu}$}{S_hx and S_hu}}

The derivations of the moments $\bm S_{hx}$ and $\bm S_{hu}$ are highly similar, so we only provide a detailed derivation for $\bm S_{hx}$ and then directly present the expression for $\bm S_{hu}$. According to \eqref{eq:moment}, the moment $\bm S_{hx}$ can be decomposed into two terms: $\bm S_{hx,1} = \mathbb{E}[\bm \mu_{\mathrm{gp}}(\bm{z}, \bm v)\bm x^T]$ and $\bm S_{hx,2} = \bm m_h \bm m_{x}^T$. For the $k$-th row of the first term, using \eqref{eq:ph_xu_1}, we have:


\begin{equation}\label{eq:Shx_1}
\begin{aligned}
    S_{h^kx, 1} 
    &= \mathbb{E} [\bm K_{h^k u^k} \bm{v}^k \bm{x}^T] \\
    &= \mathbb{E} 
    \left[\sum_{j=1}^{n_{u^k}} \bm K_{h^k u^k_j} \bm{v}_j^k \bm{x}^T\right] \\
    &= \sum_{j=1}^{n_{u^k}} 
    \mathbb{E} [\bm K_{h^k u^k_j} \bm{v}_j^k \bm{x}^T] \\
\end{aligned}
\end{equation}
where the evaluation of the expectation term is similar to that in \eqref{eq:phi_v}:

\begin{equation}\label{eq:Shx_2}
\begin{aligned}
    \mathbb{E} [\bm K_{h^k u^k_j} \bm{v}_j^k \bm{x}^T] 
    &= c^k \int \left[ \mathcal{N}({\bm z}_{j}^k | \bm z^k, \bm\Lambda^{k}) \cdot
    q(\bm z^k, \bm v, \bm{x}) 
    \, \bm v_j^k \bm{x}^T \right]
    \mathrm{d}\bm{z}^k \mathrm{d}\bm{v} \mathrm{d}\bm{x} \\
    &= b_j^{k} \int \left[  q(\bm z^k, \bm v, \bm{x} | \bm z_j^k)
    \, \bm{v}_j^k \bm{x}^T  \right]
    \, \mathrm{d}\bm{z}^k \mathrm{d}\bm v \mathrm{d}\bm{x} \\
    &= b_j^{k} \left( \bm{S}_{v_j^k x|z_j^k} 
    + \bm{m}_{v_j^k|z_j^k} \bm{m}_{x|z_j^k}^{T} \right)
\end{aligned}
\end{equation}
where the moments $\bm{S}_{v_j^k x|z_j^k}$, $\bm{m}_{v_j^k|z_j^k}$ and $\bm{m}_{x|z_j^k}$ can be evaluated using \eqref{eq:moment_kf1}.
Combining \eqref{eq:Shx_1} and \eqref{eq:Shx_2}, the expression for $\bm S_{hx}$ is:

\begin{equation}\label{eq:Shx}
\begin{aligned}
    \bm S_{h^kx} = \sum_{j=1}^{n_{u^k}} b_j^{k} \left( \bm{S}_{v_j ^k x|z_j^k} + \bm{m}_{v_j^k|z_j^k} \bm{m}_{x|z_j^k}^{T} \right) - \bm m_{h^k} \bm m_{x}^T \\
\end{aligned}
\end{equation}

The expression for $\bm S_{hu}$ is similar and can be given by:

\begin{equation}\label{eq:Shu}
\begin{aligned}
    \bm S_{h^ku} = \sum_{j=1}^{n_{u^k}} b_j^{k} \left( \bm{S}_{v_j^k u|z_j^k} + \bm{m}_{v_j^k|z_j^k} \bm{m}_{u|z_j^k}^T \right)- \bm m_{h^k} \bm m_{u}^T \\
\end{aligned}
\end{equation}
where the moments $\bm{S}_{v_j^k u|z_j^k}$, $\bm{m}_{v_j^k|z_j^k}$, $\bm{m}_{x|z_j^k}$ and $\bm{m}_{u|z_j^k}$ can be evaluated using \eqref{eq:moment_kf1}.

\subsection{Cholesky Version of the RGPSSM-H Algorithm}\label{subsec:chol}

Numerical instability may occur when updating the covariance matrix $\bm \Sigma_{t} = \mathrm{cov}[(\bm u, \bm x_t)]$, as is often observed in Kalman filter implementations. To improve stability, matrix factorizations such as the Cholesky decomposition are commonly used. The original RGPSSM \cite{zheng2024recursive} proposes using the factorization $\bm \Sigma_{t} = \bm L_t \bm L_t^T$, where the lower-triangular matrix $\bm L_t$ is the Cholesky factor. Based on this factorization, an update equation for the Cholesky factor is derived to replace the update equation for $\bm \Sigma_{t}$. This approach makes the computation compact and efficiently improves numerical stability.

Given the algorithm improvements in Section \ref{sec:heterogeneous}, 
there are three necessary modifications for the update equation of the Cholesky factor $\bm L_t$: the adding points operation, the discarding points operation, and moment propagation in the prediction step. Among these procedures, the adding points operation and the discarding points operation involve adding and deleting certain row and column blocks of the joint covariance matrix $\bm\Sigma_t = \mathrm{var}[(\bm u, \bm x_t)]$. 
The Cholesky update for these two operations can be efficiently performed using the algorithms in Appendix~\ref{subsec:chol_add} and Appendix~\ref{subsec:chol_delete}, with $\mathcal{O}(M^2)$ complexity. 
Here, we mainly focus on the Cholesky update equations for moment propagation, which are respectively derived for the three moment matching methods in Section \ref{sec:ukf_adf}.

\textbf{EKF-based moment propagation (Eq.\eqref{eq:pred_ekf}).} 
Firstly, we express the posterior covariance at time $t$, i.e., $\bar{\bm \Sigma}_{t} = \mathrm{cov}_{q(\bar{\bm u}, \bm x_t)}[(\bar{\bm u}, \bm x_t)]$, and the predicted covariance at time $t+1$, i.e., $\bar{\bm\Sigma}_{t+1}^- = \mathrm{cov}_{q^-(\bar{\bm u}, \bm x_{t+1})}[(\bar{\bm u}, \bm x_{t+1})]$, as follows:

\begin{equation}\label{eq:Sigma_chol}
\begin{aligned}
    \bar{\bm \Sigma}_{t} = \begin{bmatrix}
    \bm{S}_{\bar u\bar u} & \bm{S}_{\bar u x_t} \\ 
    \bm{S}_{\bar u x_t}^T & \bm{S}_{x_t x_t} \\ 
    \end{bmatrix} =
    \bar{\bm L}_t \bar{\bm L}_t^T, \quad
    \bar{\bm\Sigma}_{t+1}^- &= \begin{bmatrix}
        \bm S_{\bar u\bar u}^- &  \bm S_{\bar u, x_{t+1}}^- \\
        \bm S_{\bar u, x_{t+1}}^{-T} & \bm S_{x_{t+1}, x_{t+1}}^{-} \\
    \end{bmatrix} =
    \bar{\bm L}_{t+1}^- \bar{\bm L}_{t+1}^{-T}
\end{aligned}
\end{equation}
Then, express the two Cholesky factors as the following block structure:

\begin{equation}\label{eq:chol_factor}
\begin{aligned}
    \bar{\bm L}_t = \begin{bmatrix}
        \bm A & \bm 0 \\
        \bm B & \bm C \\
        \end{bmatrix}, \quad
    \bar{\bm L}_{t+1}^- = \begin{bmatrix}
        \bm A^- & \bm 0 \\
        \bm B^- & \bm C^- \\
        \end{bmatrix}
\end{aligned}
\end{equation}
Note that, the order of $\bm x$ and $\bar{\bm u}$ is reversed compared to that in \cite{zheng2024recursive}, which results in
$\bm S_{\bar u\bar u} = \bm A \bm A^T$. 
This adjustment is beneficial for reducing computational cost, since $\bm S^-_{\bar u\bar u} = \bm S_{\bar u\bar u}$ and thus $\bm A^- = \bm A$. 

Next, by combining \eqref{eq:pred_ekf} with the block expression in \eqref{eq:Sigma_chol} and \eqref{eq:chol_factor}, we obtain:

\begin{equation}
\begin{aligned}
    \bm S_{\bar u, x_{t+1}}^- &= \begin{bmatrix}
        \bm S_{\bar u \bar u} & \bm S_{\bar u x_t}  \\
        \end{bmatrix}
        \begin{bmatrix} \bm A_{\bar u}^T \\ \bm A_x^T \\ \end{bmatrix} \\
        &=  \bm A \begin{bmatrix}  \bm A^T & \bm B^T  \\ \end{bmatrix}
            \begin{bmatrix} \bm A_{\bar u}^T \\ \bm A_x^T \\ \end{bmatrix} \\
        &= \bm A^- \bm B^{-T}
\end{aligned}
\end{equation}
Therefore, $\bm B^- = \bm A_{\bar u} \bm A + \bm A_x\bm B$ which can be evaluated by a computational complexity $\mathcal{O}(M^2 d_x)$. In addition, since

\begin{equation}\label{eq:C_derivation}
\begin{aligned}
    \bm S_{x_{t+1}, x_{t+1}}^{-} 
    &= 
    \begin{bmatrix} \bm A_{\bar u} & \bm A_x \\ \end{bmatrix}
    \begin{bmatrix}
        \bm A & \bm 0 \\
        \bm B & \bm C \\
    \end{bmatrix}
    \begin{bmatrix}
        \bm A^T & \bm B^T \\
        \bm 0 & \bm C^T \\
    \end{bmatrix}
    \begin{bmatrix} \bm A_{\bar u}^T \\ \bm A_x^T \\ \end{bmatrix}
    + \bm\Sigma_p \\
    &= \begin{bmatrix} \bm B^{-} & \bm A_x \bm C \\ \end{bmatrix}
    \begin{bmatrix} \bm B^{-T} \\ \bm C^T \bm A_x^T \\ \end{bmatrix}
    + \bm\Sigma_p \\
    &= \bm B^- \bm B^{-T} + 
    \begin{bmatrix} \bm A_x \bm C & \bm\Sigma_p^{1/2}\\ \end{bmatrix}
    \begin{bmatrix} \bm C^T \bm A_x^T \\ \bm\Sigma_p^{1/2,T} \\ \end{bmatrix}
    \\
    &= \bm B^- \bm B^{-T} + \bm C^- \bm C^{-T} 
\end{aligned}
\end{equation}
where $\bm\Sigma_p^{1/2}$ is the Cholesky factor of the process noise covariance matrix $\bm\Sigma_p$. 
Therefore, the block $\bm C^-$ can be attained by the QR factorization:


\begin{equation}\label{eq:qr}
\begin{aligned}
    \begin{bmatrix}
        \bm{C}^T \bm A_x^T \\ \bm\Sigma_p^{1/2,T} \\
        \end{bmatrix} = \bm{O} \bm{R}
\end{aligned}
\end{equation}
where matrix $\bm O$ is an orthogonal matrix, and matrix $\bm R$ is an upper triangular matrix. Given \eqref{eq:C_derivation} and \eqref{eq:qr}, we can obtain $\bm C^- = \bm R^T$ by a computational complexity $\mathcal O(d_x^3)$. 

In summary, the Cholesky update equation for EKF-based moment propagation has been attained, which mainly involves computing the blocks $\bm B^-$ and $\bm C^-$, with a total computational complexity of $\mathcal{O}(M^2 d_x + d_x^3)$. In contrast, for the original RGPSSM, the complexity is $\mathcal{O}(M^3)$ (here $M$ uses the definition given in this paper).
Since $M \gg d_x$ in general, this paper achieves an improved computational efficiency by rearranging the order of $\bm x$ and $\bm u$ in the covariance matrix $\bm\Sigma$ as illustrated in \eqref{eq:Sigma_chol}.
For convenience in the following complexity analysis, we assume the low-dimensional system learning condition that $M \gg d_x, d_f, d_z$, and therefore we only retain the highest order of complexity with respect to $M$.



\textbf{UKF-based moment propagation (Algorithm \ref{alg:ukf}).} 
Given the block expressions in \eqref{eq:Sigma_chol} and \eqref{eq:chol_factor}, we have

\begin{equation}
\begin{aligned}
    \bm S_{\bar u, x_{t+1}}^- &= \bm A^- (\bm B^{-})^T \\
    \bm S_{x_{t+1}, x_{t+1}}^- &= \bm B (\bm B^{-})^T +  \bm C  (\bm C^{-})^T
\end{aligned}
\end{equation}
Therefore, we have:

\begin{equation}\label{eq:chol_ukf}
\begin{aligned}
\bm B^- &= \left[(\bm A^-)^{-1}\bm S_{\bar u, x_{t+1}}^- \right]^T \\
\bm C^- &= \mathrm{cholesky}\left(\bm S_{x_{t+1}, x_{t+1}}^- - \bm B^- (\bm B^{-})^T \right)
\end{aligned}
\end{equation}
where $\bm A^- = \bm A$ is the Cholesky factor for $\bm S_{\bar u\bar u}$, which is the same as in the EKF-based moment propagation. Considering Algorithm \ref{alg:ukf}, it can be found that:

\begin{equation}\label{eq:cond_ukf}
\begin{aligned}
\bm S^-_{\bar u, x_{t+1}} = 
\begin{bmatrix} \bm 0_{n_{\bar u} \times 1} 
    & \eta \bm A & \bm 0_{n_{\bar u} \times (d_x+d_f)} 
    & - \eta \bm A & \bm 0_{n_{\bar u} \times (d_x+d_f)} \end{bmatrix}
\mathrm{diag}(\bm W^{(c)}) \left[\bm{\mathcal X}_{x_{t+1}}^- - \bm m_{x_{t+1}}^- \right]^T 
\end{aligned}
\end{equation}
Therefore, by substituting \eqref{eq:cond_ukf} into \eqref{eq:chol_ukf}, we have:

\begin{equation}\label{eq:B_ukf}
\begin{aligned}
\bm B^- 
&= \left[\bm{\mathcal X}_{x_{t+1}}^- - \bm m_{x_{t+1}}^- \right] \mathrm{diag}(\bm W^{(c)})
\begin{bmatrix} \bm 0_{n_{\bar u} \times 1} 
    & \eta \bm I_{\bar u} & \bm 0_{n_{\bar u} \times (d_x+d_f)} 
    & - \eta \bm I_{\bar u} & \bm 0_{n_{\bar u} \times (d_x+d_f)} \end{bmatrix}^T \\
&= \eta W^{(c)}
    \left( \left[\bm{\mathcal X}_{x_{t+1}}^- - \bm m_{x_{t+1}}^- \right]_{1:1+n_{\bar u}}
   -\left[\bm{\mathcal X}_{x_{t+1}}^- - \bm m_{x_{t+1}}^- \right]_{a:a+n_{\bar u}} \right)
\end{aligned}
\end{equation}
where $a = 1 + n_{\bar u} + d_x + d_f$, and $W^{(c)}$ denotes the weight associated with the non-central sigma points (note that, the weight is the same for these points). The subscript $i:j$ indicates the matrix block corresponding to columns $i$ to $j$.
In addition, given Algorithm \ref{alg:ukf}, we have:

\begin{equation}\label{eq:C_ukf_0}
\begin{aligned}
\bm C^- &= \mathrm{cholesky}\left( 
    \bm{\mathcal X}^-_{x_{t+1}} 
    \mathrm{diag}(\bm W^{(c)}) 
    (\bm{\mathcal X}_{x_{t+1}}^{-})^T
    + \bm\Sigma_p - \bm B^- \bm B^{-,T}
\right)
\end{aligned}
\end{equation}
For the case when the first element of $\bm W^{(c)}$ is positive, namely, $W^{(c)}_0 > 0$, we can first perform QR factorization as follows:
\begin{equation}\label{eq:C_ukf_1}
\begin{aligned}
    \begin{bmatrix}
    \mathrm{diag}(\bm W^{(c)})^{1/2} (\bm{\mathcal X}_{x_{t+1}}^{-})^T \\
    \bm\Sigma_p^{1/2} \\
    \end{bmatrix}
    = \bm O \bm R
\end{aligned}
\end{equation}
and then conduct the Cholesky downdate algorithm:
\begin{equation}\label{eq:C_ukf_2}
\begin{aligned}
    \bm C^- = \mathrm{choldown}
    \left(\bm R^T, 
    \bm B^- \right)
\end{aligned}
\end{equation}
Here, $\mathrm{choldown}(\cdot, \cdot)$ denotes the Cholesky downdate algorithm (see Section 6.5.4-th of \cite{golubMatrixComputations2013}), an efficient method for evaluating the Cholesky factor of $(\bm R^T \bm R - \bm B^- (\bm B^{-})^T)$. For $W^{(c)}_0 < 0$, the evaluation process for $\bm C^-$ is similar, with the only difference being that we do not integrate $\bm{\mathcal X}_{0,x_{t+1}}^-$ in \eqref{eq:C_ukf_1} and handle this sample by the Cholesky downdate algorithm as in \eqref{eq:C_ukf_2}.

Therefore, through \eqref{eq:B_ukf}, \eqref{eq:C_ukf_1} and \eqref{eq:C_ukf_2}, we can attain the block of the Cholesky factor $\bar{\bm L}_{t+1}^-$. 
These computational procedures bypass the explicit computation of the full covariance matrices $\bm S_{\bar u, x_{t+1}}^-$ and $\bm S_{x_{t+1}, x_{t+1}}^-$, thus enhancing numerical stability. The underlying principle is that directly operating on the original matrix can easily introduce significant errors due to large differences in the magnitude of matrix elements. Therefore, by operating on the Cholesky factor (the square root of the original matrix), this problem can be alleviated.
At the same time, the evaluation process is efficient. Specifically, given the GP prediction equation \eqref{eq:qf} and the UKF algorithm, the complexity of the Cholesky version of the UKF-based moment propagation is $\mathcal{O}(M^2)$.


\textbf{ADF-based moment propagation (Eq.\eqref{eq:moment_adf}).} 
For simplicity, denote the joint $(\bm h, \bm x_t)$ as $\bm w$ and the Cholesky factor of its covariance as $\bm L_w$, which can be attained by directly Cholesky factorization:

\begin{equation}\label{eq:Lw}
\begin{aligned}
    \bm L_w = \mathrm{cholesky}(\bm S_{ww})
\end{aligned}
\end{equation}
This Cholesky factor can be used in the UKF to perform the state prediction task as shown in \eqref{eq:pred_x}. Based on this, we derive the evaluation formulas for the blocks $\bm B^-$ and $\bm C^-$ of the updated Cholesky factor $\bar{\bm L}_{t+1}$.
First, for the block $\bm B^-$, given the relationship in \eqref{eq:chol_ukf} and the evaluation formula for $\bm S_{\bar u x_{t+1}}^-$ in \eqref{eq:Sux}, we have
\begin{equation}\label{eq:chol_adf1}
\begin{aligned}
    \bm B^-
    = \left(\bm A^{-1} \bm S_{\bar u w} \bm S_{ww}^{-1} \bm S_{w x_{t+1}}^- \right)^T \\
\end{aligned}
\end{equation}
which can be further simplified by \eqref{eq:chol_adf2} and \eqref{eq:chol_adf3}. Specifically, 
by using the block structure as shown in \eqref{eq:Sigma_chol} and \eqref{eq:chol_factor}, we obtain:
\begin{equation}\label{eq:chol_adf2}
\begin{aligned}
    \bm A^{-1} \bm S_{\bar u w} &= \bm A^{-1} 
    \begin{bmatrix} \bm S_{\bar uh} & \bm A \bm B^T \\\end{bmatrix} \\
    &= \begin{bmatrix} \bm A^{-1} \bm S_{\bar uh} & \bm B^T \\\end{bmatrix}
\end{aligned}
\end{equation}
In addition, considering the UKF algorithm, we have:

\begin{equation}\label{eq:chol_adf3}
\begin{aligned}
    \bm S_{ww}^{-1} \bm S_{w x_{t+1}}^- &= \bm S_{ww}^{-1} 
    \begin{bmatrix} \bm 0_{d_w \times 1} 
        & \eta \bm L_w & - \eta \bm L_w & \end{bmatrix}
    \mathrm{diag}(\bm W^{(c)}) \left[\bm{\mathcal X}_{x_{t+1}}^- - \bm m_{x_{t+1}}^- \right]^T \\
    &= (\bm L_{w}^{-1})^T 
    \begin{bmatrix} \bm 0_{d_w \times 1} 
        & \eta I & - \eta \bm I & \end{bmatrix}
    \mathrm{diag}(\bm W^{(c)}) \left[\bm{\mathcal X}_{x_{t+1}}^- - \bm m_{x_{t+1}}^- \right]^T \\
    &= \eta W^{(c)} \cdot (\bm L_{w}^{-1})^T  
    \left(
    \left[\bm{\mathcal X}_{x_{t+1}}^- - \bm m_{x_{t+1}}^- \right]_{1:1+d_w}
    -\left[\bm{\mathcal X}_{x_{t+1}}^- - \bm m_{x_{t+1}}^- \right]_{1+d_w:1+2d_w} \right)^T
\end{aligned}
\end{equation}
which is similar to \eqref{eq:B_ukf}. 
Therefore, one can first evaluate \eqref{eq:chol_adf2} and \eqref{eq:chol_adf3}, and then use \eqref{eq:chol_adf1} to obtain the block $\bm B^-$.

In terms of the block $\bm C^-$, it can be obtained straightforwardly by applying the same method illustrated in (\ref{eq:C_ukf_0}) to (\ref{eq:C_ukf_2}).
Therefore, we have attained the Cholesky update equation for the ADF-based moment matching.
Combining the evaluation equations for the blocks $\bm B^-$ and $\bm C^-$, together with the function prediction equation summarized in \eqref{eq:moment_adf}, the total computational complexity is $\mathcal O(M^2)$.


In summary, this subsection mainly presents the Cholesky update equations for different moment propagation methods, including the EKF, the UKF, and the ADF. The Cholesky versions of these three methods improve numerical stability and all maintain quadratic complexity w.r.t. the maximum size of the inducing points, namely, $\mathcal O(M^2)$.

\subsection{Cholesky Factor Update for Block Augmentation of the Covariance Matrix}
\label{subsec:chol_add}

This subsection derives the update equations for the Cholesky factor when the covariance matrix is augmented by an additional block. Specifically, the original covariance matrix $\bm\Sigma$ and the augmented matrix $\bar{\bm \Sigma}$ have the following block structure:

\begin{equation}\label{eq:Sigma_add}
\begin{aligned}
    \bm\Sigma = \begin{bmatrix}
        \bm{\Sigma}_{11} & \bm{\Sigma}_{12} \\
        \bm{\Sigma}_{21} & \bm{\Sigma}_{22} \\
    \end{bmatrix}, \quad
    \bar{\bm\Sigma} = \begin{bmatrix}
        \bm{\Sigma}_{11} & \bm{\sigma}_1 & \bm{\Sigma}_{12} \\
        \bm{\sigma}_1^T & \bm{\sigma} & \bm{\sigma}_2^T \\
        \bm{\Sigma}_{21} & \bm{\sigma}_2  & \bm{\Sigma}_{22} \\
    \end{bmatrix}
\end{aligned}
\end{equation}
For facilitate of the derivation, we express the Cholesky factor of $\bm\Sigma$ and $\bar{\bm\Sigma}$ as follows:


\begin{equation}\label{eq:L_add}
\begin{aligned}
    \bm L = \begin{bmatrix}
        \bm A & \bm 0 \\
        \bm B & \bm C \\
    \end{bmatrix}, \quad
    \bar{\bm L} = \begin{bmatrix}
        \bar{\bm A} & \bm 0 & \bm 0\\
        \bar{\bm a}  & \bar{\bm b} & \bm 0 \\
        \bar{\bm B} & \bar{\bm c} & \bar{\bm C} \\
    \end{bmatrix}
\end{aligned}
\end{equation}
Therefore, using the definition of Cholesky factorization and the block expressions in \eqref{eq:Sigma_add} and \eqref{eq:L_add}, we have:
\begin{equation}\label{eq:expand_add}
\begin{aligned}
    \bar{\bm \Sigma} 
    &= \bar{\bm L} \bar{\bm L}^T
    = \begin{bmatrix}
        \bar{\bm A}\bar{\bm A}^T 
        & \bar{\bm A}\bar{\bm a}^T 
        & \bar{\bm A}\bar{\bm B}^T \\
        *
        & \bar{\bm a}\bar{\bm a}^T + \bar{\bm b}\bar{\bm b}^T 
        & \bar{\bm a}\bar{\bm B}^T + \bar{\bm b}\bar{\bm c}^T \\
        * 
        & * 
        & \bar{\bm B}\bar{\bm B}^T + \bar{\bm c}\bar{\bm c}^T + \bar{\bm C}\bar{\bm C}^T
    \end{bmatrix}
    = \begin{bmatrix}
        \bm A \bm A^T 
        & \bm\sigma_1
        & \bm A \bm B^T \\
        *
        & \bm\sigma 
        & \bm\sigma_2^T \\
        * 
        & * 
        & \bm B \bm B^T + \bm C \bm C^T
    \end{bmatrix}
\end{aligned}
\end{equation}
Given \eqref{eq:expand_add}, the blocks of $\bar{\bm L}$ can be easily evaluated from that of $\bm L$:

\begin{equation}\label{eq:chol_add}
\begin{aligned}
    \bar{\bm A} &= \bm A, \quad \bar{\bm B} = \bm B \\
    \bar{\bm a} &= \left( \bm A^{-1}\bm\sigma_1 \right)^T \\
    \bar{\bm b} &= \mathrm{cholesky}\!\left(\bm\sigma - \bar{\bm a} \bar{\bm a}^T \right), \\
    \bar{\bm c} &= \left[ \bar{\bm b}^{-1}\!\left(\bm\sigma_2^T - \bar{\bm a}\bm B^T\right) \right]^T \\
    \bar{\bm C} &= \mathrm{choldown}\!\left(\bm C,\, \bar{\bm c}\right).
\end{aligned}
\end{equation}
where $\mathrm{cholesky}(\cdot)$ denotes the Cholesky factorization and $\mathrm{choldown}(\cdot, \cdot)$ denotes the Cholesky downdate algorithm (see Section 6.5.4 of \cite{golubMatrixComputations2013}). 
When adding a block of width $b$, since $\bm A$ and $\bar{\bm b}$ are triangular, the computational complexity of \eqref{eq:chol_add} is $\mathcal O(bd^2)$, where $d$ is the dimension of $\bm\Sigma$.

\subsection{Cholesky Factor Update under Blockwise Reduction of the Covariance Matrix}\label{subsec:chol_delete}

This subsection derives the update equations for the Cholesky factor when the covariance matrix is reduced by removing an additional block. Here, we use the notation in \eqref{eq:Sigma_add} and \eqref{eq:L_add}, and consider $\bar{\bm\Sigma}$ as the original matrix and $\bm\Sigma$ as the reduced one.

To obtain the blocks in the reduced Cholesky factor $\bm L$, we use the expansion in \eqref{eq:expand_add} and the block expressions in \eqref{eq:Sigma_add} and \eqref{eq:L_add}, which leads to:




\begin{equation}
\begin{aligned}
\bm{\Sigma} = 
\begin{bmatrix}
\bm A \bm A^T & \bm A \bm B^T  \\
* & \bm B \bm B^T + \bm C \bm C^T
\end{bmatrix} = \begin{bmatrix}
\bar{\bm{A}} \bar{\bm{A}}^T & \bar{\bm{A}} \bar{\bm{B}}^T \\
* & \bar{\bm{B}} \bar{\bm{B}}^T + \bar{\bm{c}} \bar{\bm{c}}^T + \bar{\bm{C}} \bar{\bm{C}}^T \\
\end{bmatrix}
\end{aligned}
\end{equation}
Therefore, the blocks of $\bm L$ can be evaluated from that of $\bar{\bm L}$ by:

\begin{equation}\label{eq:chol_delete}
\begin{aligned}
\bm A &= \bar{\bm A}, \quad \bm B = \bar{\bm B} \\
\bm C &= \mathrm{cholupdate}(\bar{\bm{C}}, \bar{\bm{c}})
\end{aligned}
\end{equation}
where $\mathrm{cholupdate}(\cdot, \cdot)$ denotes the Cholesky update algorithm (see Section 6.5.4 of \cite{golubMatrixComputations2013}), an efficient method for evaluating the Cholesky factor of $(\bar{\bm{C}} \bar{\bm{C}}^T + \bar{\bm{c}} \bar{\bm{c}}^T)$. When deleting a block of width $b$, the computational complexity of \eqref{eq:chol_delete} is $\mathcal O(bd^2)$, where $d$ is the dimension of $\bm\Sigma$.



\bibliography{reference}

\end{document}